\documentclass{article}
\usepackage[T1]{fontenc}
\usepackage[utf8]{inputenc}
\usepackage{iclr2027_conference,times}

\usepackage{amsmath,amssymb,amsthm,mathtools}
\usepackage{graphicx}
\usepackage{booktabs,longtable,array}
\usepackage{placeins}
\usepackage{microtype}
\usepackage{flafter}
\usepackage{xcolor}
\usepackage{hyperref}
\usepackage{url}
\hypersetup{hidelinks}

\newtheorem{theorem}{Theorem}
\newtheorem{proposition}{Proposition}
\newtheorem{corollary}{Corollary}
\newtheorem{lemma}{Lemma}
\theoremstyle{definition}
\newtheorem{definition}{Definition}
\theoremstyle{remark}
\newtheorem{remark}{Remark}

\DeclareMathOperator{\TV}{TV}
\DeclareMathOperator{\KL}{KL}
\DeclareMathOperator{\Bern}{Bern}

\newcommand{\cH}{\mathcal{H}}
\newcommand{\cA}{\mathcal{A}}
\newcommand{\cF}{\mathcal{F}}

\newcommand{\E}{\mathbb{E}}
\newcommand{\1}{\mathbf{1}}

\newcommand{\defeq}{\mathrel{:=}}

\iclrfinalcopy
\title{The Price of Choice in Selective Prediction}
\author{%
Daniel Cie\'slak\thanks{Corresponding author: \texttt{daniel.cieslak@pg.edu.pl}}\\
{\normalfont\small Gda\'nsk University of Technology}\\
{\normalfont\small Gda\'nsk, Poland}
\And
Andrzej Czy\.zewski\\
{\normalfont\small Gda\'nsk University of Technology}\\
{\normalfont\small Gda\'nsk, Poland}
}

\begin{document}
\maketitle
\lhead{Preprint, October 2026}
\begin{abstract}
A vendor may claim that \emph{some} rule from a predeclared menu---a score threshold or a union of $K$ score intervals---lets its selective classifier keep a fraction $\kappa$ of inputs while erring on at most a fraction $\alpha$ of them. Refuting such an existential claim means ruling out every permitted rule at once. The intersection--union test (IUT) does this without a multiplicity correction, but its sample cost as the menu grows, and how close that cost is to the best valid audit, are not captured by the level argument. We characterize both. For a fixed interior risk target $\alpha$, menus of $K$ intervals and a risk gap $\gamma$, in the small-gap, large-menu regime of our theorem the minimax sample size at each fixed coverage floor $\kappa\in(0,1)$ is $K(1-\kappa)/(\kappa\gamma^2)$ up to logarithmic factors, and the exact IUT attains it: \emph{menu richness costs $K$ times the abstention odds}, even though validity costs nothing. The lower bound holds for audits that know the score distribution; the upper bound extends to any menu of VC dimension $d$. We then explain the linear cost. Every certificate that never rejects while some rule looks feasible on the audit data---the IUT, Bonferroni, DKW and uniform-convergence bounds---pays it even at laws where $\sqrt K/\gamma^2$ samples suffice, because the best-looking rule is optimistically biased. We then identify when the gap can and cannot be avoided: against constant-risk alternatives a reference-free audit reaches the testing scale, including the square-root rate on the sign family in its stated small-gap regime, whereas over a rich class with unknown heterogeneous risk the linear cost is unavoidable up to a logarithm. In simulation with $K=4$, the raw menu cost changes 7.7-fold between $\kappa=.5$ and $.95$; after dividing by the abstention odds and the theorem-motivated localized-VC logarithmic normalization used in Figure~\ref{fig:kappa}, the two endpoint values differ by about 5\%. Held-out model families show that the same feasible-mixture mechanism is not confined to synthetic constructions: across three contemporary VLM judges, all 60 split-specific confidence-gate families have positive exact-hiding windows and 28/30 contracts selected on audit-A remain hidden on untouched audit-B; on the cleaner multi-annotator Polaris target, hiding persists on all 20 splits but is much narrower and transfers in 3/10 resplits. Frozen image and multimodal classifiers separately show boundary behavior consistent with the exact IUT guarantee and strong power relative to the practical certificates we compare.
\end{abstract}

\section{Introduction}
Before a selective classifier is deployed---in procurement, certification or a regulatory feasibility check---its vendor often claims that it \emph{can be configured} to meet a standard: some permitted acceptance rule retains at least 90\% of inputs while erring on at most 10\% of the accepted ones. The menu of permitted rules is fixed in advance, for example thresholds on a confidence score or unions of a few score intervals. An auditor with fresh labeled data who wants to refute the claim cannot test one rule of its choice: showing that one rule fails leaves the vendor free to pick another. The null hypothesis is a \emph{union} over the menu, and a rejection must rule out every rule at once. This is the dual of risk control and Learn-then-Test, which certify one rule selected by calibration \citep{chow1970optimum,elyaniv2010foundations,geifman2017selective,bates2021riskcontrol,angelopoulos2024conformal,angelopoulos2025learntest}.

The intersection--union test (IUT) is the natural audit: reject when every rule fails its own valid test. Because a true claim has one fixed witness rule, no multiplicity correction is needed \citep{berger1982multiparameter,berger1996bioequivalence}. This invites a tempting conclusion---that the size of the menu is free. It is not. We characterize the statistical price of menu richness, compare the exact IUT with the best valid audit, and identify assumptions under which the gap can close.

\paragraph{The main result in one line.} For menus of at most $K$ score intervals, a fixed interior risk target $\alpha$, a fixed coverage floor $\kappa\in(0,1)$ and risk gap $\gamma$, the number of audit samples needed to refute every robustly false claim is
\begin{equation}
 N\;\asymp\;\underbrace{\frac{1-\kappa}{\kappa}}_{\text{abstention odds}}\cdot\frac{K}{\gamma^2}\qquad\text{in the small-gap, large-menu regime, up to logarithmic factors,}
 \label{eq:headline}
\end{equation}
and the exact IUT attains it (Theorem~\ref{cor:intervals}). This is the price of the menu, and it arises because the vendor's freedom lies in \emph{which} small part of the score line to abstain on (or, for $\kappa<1/2$, to accept): the auditor must rule out that the errors happen to be concentrated there, and the error mass of such a region fluctuates on the scale of that region, not of the whole sample. Richness is therefore cheap at high floors: at $K=4$ in our simulations the menu cost at $\kappa=.9$ is 4.4 times smaller than at $\kappa=.5$ (Figure~\ref{fig:kappa}).

\paragraph{Contributions.} Table~\ref{tab:regimes} summarizes the results; every lower bound holds for audits that are valid over the full, unrestricted null.
\begin{enumerate}
\item \emph{The abstention-odds law and optimality of the IUT} (Section~\ref{sec:main}). For any menu of VC dimension $d$ the IUT needs $\tilde O(d(1-\kappa+\gamma)/(\kappa\gamma^2))$ samples (Proposition~\ref{prop:bracket}); for unions of $K$ intervals, in the theorem's small-gap regime, no valid audit, even one that knows the score distribution, needs fewer than $\Omega(K(1-\kappa)/(\kappa\gamma^2\log K))$ at any fixed floor $\kappa$ (Theorem~\ref{cor:intervals}). The law covers the high floors used in practice ($\kappa=.9$ in our experiments), where richness is several times cheaper than at $\kappa=1/2$.
\item \emph{A plug-in barrier} (Section~\ref{sec:iut}). We call a certificate EFC if it never rejects while some rule looks feasible on the audit data; the IUT, Bonferroni, DKW and uniform-convergence bounds all are. On explicit families with $k$ effective groups, every valid EFC certificate needs $\Omega(k/\gamma^2)$ samples at a law where $\Theta(\sqrt k/\gamma^2)$ suffice (Theorems~\ref{thm:separation} and~\ref{thm:iut-rate}), and this carries over to interval menus (Remark~\ref{rem:interval-efc}). The obstruction is a winner's curse: the best empirical rule is optimistically biased by $\sqrt{k/n}$.
\item \emph{What a better audit must know} (Section~\ref{sec:knowledge}). An unknown score distribution does not force the linear cost: against constant-risk alternatives, a reference-free audit valid for the unrestricted null reaches the testing scale and, on the uniform sign family in the stated small-gap regime, the square-root rate (Theorem~\ref{thm:constant}). By contrast, uncertainty about heterogeneous conditional risk can force the linear cost up to a logarithm over the class of Theorem~\ref{thm:heterogeneity}.
\item \emph{Evidence} (Section~\ref{sec:experiments}). Simulations illustrate the coverage-floor dependence and the testing--estimation separation. Held-out families of frozen classifiers and VLM-judge confidence gates exhibit feasible-mixture hiding; on three VLM judges 28/30 audit-A midpoint contracts remain hidden on untouched audit-B, while a clean multi-annotator control retains positive but much smaller windows. Separate frozen-classifier audits show boundary behavior consistent with exact IUT level and strong finite-sample power against Bonferroni, DKW, total-loss and a valid non-EFC universal-inference audit.
\end{enumerate}
For practitioners the message is short. Use the exact IUT: it needs no correction, runs in $O(MKn)$ time over $M$ score cells, and is minimax up to logarithms in the regime of Theorem~\ref{cor:intervals}. Within that regime, budget for the menu with~\eqref{eq:headline}: richness is cheap at high coverage floors and the menu penalty grows as the floor decreases. The held-out classifier and VLM-judge studies show why a family-level audit matters in practice: multiple individually failing gates can still hide the reference through feasible mixtures, while the constant-risk square-root shortcut requires additional structure.

\begin{table}[t]
\caption{Minimax sample sizes of valid audits (up to constants; $\tilde\Omega,\tilde O$ hide logarithms). Row 1: robustly infeasible alternatives $\mathcal P_\gamma$ (Section~\ref{sec:main}); the interval bounds need large $K$, an interior risk target $\alpha$ bounded away from 0 and 1, and $\gamma\lesssim\min\{\kappa,1-\kappa\}/\log K$. Rows 2--4: the sign families of Section~\ref{sec:iut} ($\kappa=1/2$, $M=2B$ cells, $k=B$ in rows 3--4, large $B$, $\gamma\le\gamma_0$). ``Valid'' means level $\delta$ over the \emph{unrestricted} null.}
\label{tab:regimes}
\centering\footnotesize
\setlength{\tabcolsep}{4pt}
\begin{tabular}{@{}>{\raggedright\arraybackslash}p{2.1cm}>{\raggedright\arraybackslash}p{2.6cm}>{\raggedright\arraybackslash}p{4.5cm}>{\raggedright\arraybackslash}p{3.5cm}@{}}
\toprule
Known to auditor & Power required against & Minimax sample size & EFC certificates\\
\midrule
Menu only & $\mathcal P_\gamma$ (robust margin $\gamma$) & $K$ intervals, fixed $\kappa$: $\tilde\Theta\big(\frac{K(1-\kappa)}{\kappa\gamma^2}\big)$ (Thm.~\ref{cor:intervals}) & IUT: $\tilde O\big(\frac{d(1-\kappa+\gamma)}{\kappa\gamma^2}\big)$, VC dim.\ $d$\\
Reference $P_0$ & $P_0$ & $\Theta(\sqrt{k}/\gamma^2)$ (Thm.~\ref{thm:separation}) & $\Omega(k/\gamma^2)$; IUT: $\Theta(k/\gamma^2)$ (Thm.~\ref{thm:iut-rate})\\
Cells only & constant risk, any marginal & $\Theta(\sqrt{B}/\gamma^2)$ if $\gamma\le(2B)^{-1/4}$ (Thm.~\ref{thm:constant}) & $\Omega(B/\gamma^2)$ at $P_0$\\
Cells and marginal & heterogeneous risk & $\tilde\Theta(B/\gamma^2)$ (Thm.~\ref{thm:heterogeneity}) & IUT: $O(B/\gamma^2)$\\
\bottomrule
\end{tabular}
\end{table}

\paragraph{Positioning and contributions.} Classical analyses of the IUT concern a fixed number of normal means \citep{lehmann1952testing,berger1989uniformly}; tests of many moment inequalities and reality checks have an \emph{intersection} null and pay only $\log K$ \citep{white2000reality,chernozhukov2019inference}; property testing of function classes separates known-distribution from distribution-free testing \citep{kearns2000testing,blais2021vc}. Our null is a union of convex sets of laws with a ratio-type risk constraint, validity is required over all of it, and the question is how the sample size depends on the menu. The lower-bound tools---mixture second moments, identity and independence testing, moment matching---are classical \citep{ingster1987minimax,valiant2017identity,diakonikolas2016new,cai2011testing}. Our contributions are the EFC barrier, the abstention-odds law, and three constructions they require: a padded moment-matching prior whose null draws stay \emph{existentially} feasible while the alternative draws are robustly infeasible against every measurable set at any prescribed floor; a witness that controls the coverage screen and the risk simultaneously; and a localization of the IUT to the smaller side of the floor.

\section{The audit problem}
\label{sec:geometry}
Observe i.i.d. pairs $(S,L)$ of score and binary loss. Freeze a measurable family $\cA$ of acceptance rules (the \emph{menu}) and a contract $(\kappa,\alpha)$ independently of the audit sample. A law $Q$ is feasible if
\begin{equation}
 \exists A\in\cA:\quad c_Q(A)=Q(S\in A)\ge\kappa,\qquad
 r_Q(A)=Q(L=1\mid S\in A)\le\alpha.\label{eq:contract}
\end{equation}
Let $\cH(\cA)$ be this existential null. It is \emph{unrestricted}: score probabilities and conditional losses may both change. For a class $\mathcal P$ of infeasible laws, the minimax sample size is
\begin{equation}
 N(\mathcal P,\cH;\delta,\beta)=\inf\Big\{n\ge1:\exists\phi_n\in[0,1],\
 \sup_{Q\in\cH}\E_{Q^n}\phi_n\le\delta,\ \inf_{P\in\mathcal P}\E_{P^n}\phi_n\ge\beta\Big\}.
 \label{eq:sample-complexity}
\end{equation}
$N_{\rm IUT}$ is the same quantity with $\phi_n$ restricted to the exact IUT below. For a single law $P_0$ we write $N(P_0,\cH)$; the optimal test may then use $P_0$, so this is an information benchmark as in local minimax testing \citep{balakrishnan2019local}. Larger classes $\mathcal P$ encode what the auditor does not know.

\paragraph{The exact IUT.} For a rule $A$, let $T_A$ and $E_A$ count the audit observations it accepts and the accepted errors; $T_A\sim\mathrm{Bin}(n,c(A))$ and $E_A\mid T_A=t\sim\mathrm{Bin}(t,r(A))$. Let $U_A$ be the one-sided Clopper--Pearson upper bound for coverage at error $\zeta_c$ and $L_A$ the lower bound for risk at error $\zeta_r$ \citep{clopper1934use}, and define the \emph{certified frontier}
\begin{equation}
 L_{\rm IUT}(\cA,\kappa)=\min_{A:\,U_A\ge\kappa}L_A,
 \qquad\min\varnothing=+\infty.\label{eq:iut}
\end{equation}
It reads: with confidence $1-\delta$, no permitted rule of coverage at least $\kappa$ has risk below $L_{\rm IUT}$.
\begin{proposition}[Validity without correction]
\label{prop:iut}
Rejecting feasibility when $L_{\rm IUT}>\alpha$ has level at most $\zeta_c+\zeta_r\le\delta$ over the unrestricted null.
\end{proposition}
Under the null a fixed population witness $A^*$ exists, and rejection requires failure of one of its two bounds. For unions of at most $K$ intervals over $M$ frozen score cells a dynamic program computes~\eqref{eq:iut} exactly in $O(MKn)$ time (Appendix~\ref{app:iut}). We use $\zeta_c=\zeta_r=\delta/2$.

\paragraph{Plug-in (EFC) certificates.} Call $A$ \emph{empirically feasible} if $T_A\ge\kappa n$ and $E_A\le\alpha T_A$.
\begin{definition}[EFC audits]
An audit $\phi_n$ is \emph{empirical-feasibility consistent} (EFC) if $\phi_n=0$ whenever some $A\in\cA$ is empirically feasible.
\end{definition}
EFC audits never reject a claim that the vendor can apparently meet on the audit data itself, which makes their decisions easy to explain. Every certificate that screens with a coverage upper bound no smaller than the empirical coverage and rejects via risk lower bounds no larger than the empirical risk is EFC. This includes the IUT (for $\zeta_c,\zeta_r\le1/2$), its Bonferroni variant, DKW and Rademacher uniform-convergence certificates, and the total-loss certificate (Appendix~\ref{app:efc}). EFC plays the role of monotonicity in classical IUT theory \citep{perlman1999emperor}.

\section{The abstention-odds law}
\label{sec:main}
For an arbitrary menu we place the power requirement on alternatives that are \emph{robustly} infeasible:
\[
 \mathcal P_\gamma(\cA)=\{P:\ r_P(A)\ge\alpha+\gamma\ \text{for every }A\in\cA\text{ with }c_P(A)\ge\kappa-\gamma\}.
\]
The class allows arbitrary score marginals and arbitrary, including informative, risk functions. The coverage slack $\gamma$ excludes rules that sit just below the floor with very low risk, which no finite audit can resolve.
\begin{proposition}[The IUT for every menu]
\label{prop:bracket}
Let $\cA$ have VC dimension $d$, $0<\delta<\beta<1$, $\kappa\in(0,1]$, $0<\gamma\le\kappa/2$ and $\alpha+\gamma\le1$. For absolute constants $C$:
(i) the exact IUT has power at least $\beta$ uniformly over $\mathcal P_\gamma(\cA)$ once $n\ge C\{d+\log(2/\delta)+\log(4/(1-\beta))\}/(\kappa\gamma)^2$;
(ii) if some $A_0\in\cA$ has coverage at least $\kappa$ under some score law, every valid audit needs $n\ge2(\beta-\delta)^2\alpha(1-\alpha)/\gamma^2$;
(iii) \emph{localization:} the exact IUT has power at least $\beta$ over $\mathcal P_\gamma(\cA)$ whenever
\[
 n\ge C\,\frac{(1-\kappa+\gamma)\,d\log(2en/d)+\log\{8/(\delta(1-\beta))\}}{\kappa\gamma^2}.
\]
\end{proposition}
Part (i) combines VC uniform deviations with Clopper--Pearson inversion; part (ii) is a two-point bound (Appendix~\ref{app:bracket}). Part (iii) is the refinement that matters at high floors. A rule that passes the coverage screen differs from the whole score line only on a region of probability at most $1-\kappa+\gamma$, and it accepts at least a fraction $\kappa/2$ of the data. Writing its error mass as the overall error mass minus the error mass of the abstained region, only the latter varies over the menu. Relative (Bernstein-type) VC deviations bound it by $\sqrt{(1-\kappa+\gamma)d/n}$ up to logarithms, which for $\kappa\ge1/2$ is also the order of the risk error. When $\kappa<1/2$ the accepted region is the small side, its error mass fluctuates by $\sqrt{\kappa d/n}$, and dividing by $\kappa$ gives $\sqrt{d/(\kappa n)}$; both cases are $\sqrt{(1-\kappa+\gamma)d/(\kappa n)}$ up to a factor two. The Clopper--Pearson bounds adapt to both scales automatically, so the IUT needs no tuning.

For interval menus, the next result matches this dependence up to logarithmic factors at any fixed coverage floor.
\begin{theorem}[Interval menus at any fixed coverage floor]
\label{cor:intervals}
Let $\cA_K$ contain all unions of at most $K$ intervals of the score line, $\kappa\in(0,1)$, $a\in(0,1/8]$, $\alpha\in[2a,1-2a]$ and $0<\delta<\beta<1$. There are $c,c',C,K_0>0$ depending only on $(a,\delta,\beta)$ such that for $K\ge K_0$ and $\gamma\le c'\min\{\kappa,1-\kappa\}/\log K$,
\[
 \frac{c\,K(1-\kappa)}{\kappa\gamma^2\log K}\;\le\;N(\mathcal P_\gamma(\cA_K),\cH)\;\le\;N_{\rm IUT}(\mathcal P_\gamma(\cA_K))\;\le\;\frac{C\{K(1-\kappa)\log(e/(\kappa\gamma))+1\}}{\kappa\gamma^2}.
\]
The lower bound holds even for audits that know the score distribution.
\end{theorem}
Thus, for every fixed $\kappa\in(0,1)$ in the stated small-gap regime, the IUT is minimax optimal up to logarithmic factors. Moreover, Proposition~\ref{prop:bracket}(i) gives $O(K/\gamma^2)$ at fixed $\kappa$, leaving only the $\log K$ gap to the lower bound. \emph{Proof idea.} The lower bound embeds $K-1$ consecutive score groups, each split into two halves whose losses are $m\pm d_b$ with a random sign and a random magnitude $d_b$. Under the null and alternative priors the magnitudes match their first $\log K$ even moments but differ in their mean by the best polynomial approximation error of $\sqrt u$, so the two mixtures are indistinguishable until $n\gtrsim K/(\gamma^2\log K)$ (Theorem~\ref{thm:heterogeneity} below). To reach an arbitrary floor, a \emph{padding} region carries the rest of the score mass: for $\kappa\ge1/2$ it has risk $\alpha$ and is accepted by every good rule, so only a fraction $2(1-\kappa)$ of the data is informative and the risk gap shrinks by $(1-\kappa)/\kappa$; for $\kappa<1/2$ it has risk one and is never accepted. The delicate step is robust infeasibility: under the alternative prior every measurable set of coverage at least $\kappa-\gamma$ must have risk at least $\alpha+\gamma$, not only the menu's rules. Because the conditional risk is piecewise constant, the least-risk sets fill the score line in increasing order of risk, and trimming them changes the risk by at most $2\gamma D/\kappa$ (Appendix~\ref{app:intervals}).

\paragraph{Reading the law.} The menu cost $K(1-\kappa)/(\kappa\gamma^2)$ is a cost of \emph{power}, not of validity: the level proof pays nothing for $|\cA|$, yet a richer menu gives the vendor more ways to place the abstained region where the audit sample happens to contain few errors. Part (i) holds for any uniform-convergence certificate with radius of order $\sqrt{d/n}$; the endpoint DKW certificate (radius of order $K/\sqrt n$) and policywise Bonferroni (of order $\sqrt{K\log M/n}$ over $M$ cells) have weaker guarantees, and part (iii) uses the Clopper--Pearson bounds' adaptation to small proportions. The logarithmic gaps and constants remain open; at $\kappa=1/2$, with the constants of our proof, the new lower bound exceeds the elementary $\Omega(\sqrt K/\gamma^2)$ obstruction from Proposition~\ref{prop:weight-profile} only for small $\gamma$ and large $K$. The next two sections explain \emph{why} the cost is linear in $K$ and when an audit can avoid it.

\section{A testing--estimation gap for existential audits}
\label{sec:iut}
\label{sec:overlap}
\paragraph{Sign families.} Fix $p\in(0,1)$, $0<\gamma<\min\{p,1-p\}$ and $(\kappa,\alpha)=(1/2,p-\gamma)$. The reference $P_0$ has $S\sim\mathrm{Unif}[0,1]$ and $L\mid S\sim\Bern(p)$; it is infeasible. For positive weights $w$ in the simplex $\Delta_{B-1}$, partition $[0,1]$ into groups $G_\ell$ of mass $w_\ell$, each split into equal halves. Each $\theta\in\{-1,+1\}^B$ accepts one half per group, giving $2^B$ rules $A_\theta$. A feasible component keeps the score law and sets
\begin{equation}
 Q_\theta(L=1\mid S=x)=p-\gamma s_\theta(x),\qquad
 s_\theta(x)=1\ \text{on }A_\theta,\ -1\ \text{off }A_\theta.
 \label{eq:qtheta}
\end{equation}
Every component has coverage $1/2$, accepted risk $p-\gamma$, and the same distance from $P_0$; mixing them uniformly gives $P_0$ exactly. All positive weight profiles share the same single-rule testing problems, incidence structure, VC dimension $B$ and growth function (Propositions~\ref{prop:singleton}--\ref{prop:combinatorial}); the families differ only in how their rules overlap. For the plateau profile with $k$ heavy groups, $w_\ell=(1-B^{-2})/k$ for $\ell\le k$ and the remaining mass $B^{-2}$ spread over the other groups, the effective dimension is $d_{\rm eff}=\|w\|_2^{-2}=k\{1+o(1)\}$.

\begin{theorem}[Testing scale]
\label{thm:separation}
Fix $p\in(0,1)$ and $0<\delta<\beta<1$. If $\gamma=\gamma_B<\min\{p,1-p\}$ and $B^{-2}\le\gamma_B/16$, then as $B\to\infty$, uniformly in $k\le B$, the full-null sample size at the reference is $N(P_0,\cH_{B,k};\delta,\beta)=\Theta(\sqrt{k}/\gamma_B^2)$.
\end{theorem}
The lower bound is Ingster's second-moment method. For a prior $\Pi$ over feasible components with $\E_\Pi Q=P_0$, $1+\chi^2(\E_\Pi Q^n\Vert P_0^n)=\E_{Q,Q'\sim\Pi}[1+\Xi(Q,Q')]^n$ with $\Xi=\int(dQ/dP_0)(dQ'/dP_0)dP_0-1$ (Lemma~\ref{thm:overlap}); here $\Xi(\theta,\theta')=\lambda\sum_\ell w_\ell\theta_\ell\theta'_\ell$ with $\lambda=\gamma^2/[p(1-p)]$, which gives:
\begin{proposition}[Weight-profile obstruction]
\label{prop:weight-profile}
Every level-$\delta$ audit of a sign family has $\E_{P_0^n}\phi\le\delta+\tfrac12\{\exp(n^2\lambda^2\|w\|_2^2/2)-1\}^{1/2}$, hence $N_w=\Omega(\sqrt{d_{\rm eff}}/\gamma^2)$.
\end{proposition}
For the upper bound, every law of the full null---whether it changes scores, labels or both---is $\Omega(\gamma)$-far from $P_0$ after pushing forward to the $(4k+2)$ half/label cells of the heavy groups, so identity testing needs $O(\sqrt{k}/\gamma^2)$ samples \citep{valiant2017identity} (Appendix~\ref{app:separation}).

\begin{theorem}[The estimation cost of EFC audits]
\label{thm:iut-rate}
Fix $p\in(0,1)$ and $0<\delta<\beta<1$. There are $\gamma_0,c>0$ depending only on $(p,\delta,\beta)$, and for each split below a $C>0$, such that on the plateaus of Theorem~\ref{thm:separation}, whenever $\gamma\le\gamma_0$ and $B^{-2}\le\gamma/16$:
(i) every level-$\delta$ EFC audit has power below $\beta$ at $P_0$ for all $n\le ck/\gamma^2$;
(ii) the exact IUT with any fixed split $\zeta_c+\zeta_r=\delta$, $\zeta_c,\zeta_r\in(0,1/2]$, has power at least $\beta$ for all $n\ge Ck/\gamma^2$.
Hence $N_{\rm IUT}=\Theta(k/\gamma^2)$ and $N_{\rm IUT}/N(P_0,\cH_{B,k})=\Theta(\sqrt{k})$.
\end{theorem}
\emph{Mechanism.} In each heavy group choose the half containing more \emph{correct} audit observations. Conditional on the correct observations and the error count, this rule accepts each error independently with probability $1/2$, while its accepted correct count exceeds half of all correct observations by $G$ with $\E G\gtrsim\min\{n,\sqrt{nk}\}$. For $n\ll k/\gamma^2$ this gain overwhelms the population gap $n\gamma$ and keeps the empirical coverage at least $1/2$: a sample-dependent rule is empirically feasible, so no EFC audit rejects (Appendix~\ref{app:iut-rate}). Choosing the half with fewer errors would not control coverage. The optimal audit instead aggregates quadratic evidence whose fluctuation is $\sqrt{k}/n$ against a signal $\gamma^2$. The EFC bound does not rely on the reference coverage sitting exactly at the floor: with $q$-ary groups and reference coverage $m/q>\kappa$ it persists (Proposition~\ref{prop:efc-slack}, Appendix~\ref{app:efc-slack}).
\begin{remark}[The barrier on interval menus]
\label{rem:interval-efc}
Order the $2K$ halves of the uniform sign family ($k=B=K$) along the score line, each group's halves adjacent. Unions of at most $K$ intervals of these cells contain every sign rule, so Theorem~\ref{thm:iut-rate}(i) applies: for $\gamma\le\gamma_0$, every valid EFC audit of this interval menu needs $\Omega(K/\gamma^2)$ samples at $P_0$, without a logarithm, and the IUT needs $O(K/\gamma^2)$. Theorem~\ref{thm:constant} below gives a valid non-EFC audit with $O(\sqrt K/\gamma^2)$ samples at the same law for $\gamma\le(2K)^{-1/4}$.
\end{remark}

\section{When can the gap be escaped?}
\label{sec:knowledge}
The optimal audit of Theorem~\ref{thm:separation} knows $P_0$; the IUT knows nothing about it. Which missing knowledge forces the linear cost? Validity is always required over the \emph{unrestricted} null; only the power requirement is placed on a class $\mathcal P$. Let $C_1,\ldots,C_M$ be score cells frozen independently of the audit sample and $\cA$ any family of unions of cells.

\paragraph{Unknown score distribution, constant risk: a reference-free testing-scale audit.} Let $\mathcal P_{\rm c}(\alpha,\gamma)$ contain every law under which $L$ is independent of $S$ and $P(L=1)\ge\alpha+\gamma$, with an arbitrary score marginal.
\begin{theorem}[Reference-free audit at the testing scale]
\label{thm:constant}
Fix $\kappa\in(0,1]$ and $0<\delta<\beta<1$. An audit that uses only cell labels and losses, and knows neither the score marginal nor $P(L=1)$, has level $\delta$ over $\cH(\cA)$ and power at least $\beta$ uniformly over $\mathcal P_{\rm c}(\alpha,\gamma)$ for every
$n\ge C_{\delta,\beta}\{\sqrt M/(\kappa\gamma)^2+M^{2/3}/(\kappa\gamma)^{4/3}\}$.
On the uniform sign family ($M=2k$, $\kappa=1/2$) with $\gamma\le(2k)^{-1/4}$ this is $\Theta(\sqrt k/\gamma^2)$, while every valid EFC audit needs $\Omega(k/\gamma^2)$ at $P_0\in\mathcal P_{\rm c}$.
\end{theorem}
The audit rejects when the empirical loss rate exceeds $\alpha+\gamma/2$ \emph{and} an independence tester for cell and loss \citep{diakonikolas2016new} finds no dependence. The key fact is that \emph{feasibility forces dependence}: if $Q$ has a witness $A$ and $e=Q(L=1)>\alpha$, every product law $R$ on cells and labels has $\TV(Q,R)\ge(e-\alpha)\kappa/3$, because a product law assigns the witness the overall loss rate (Lemma~\ref{lem:dependence}). The audit is valid even for the family of all $2^M$ unions of cells, so it matches the testing scale only where that scale is $\sqrt M/\gamma^2$; for small menus the IUT is better. It abstains whenever its independence test detects score--loss dependence. Section~\ref{sec:experiments} separately reports strong score--loss dependence and held-out family-level hiding on frozen model outputs. The next result shows that, over a rich heterogeneous class, no valid audit can uniformly retain the square-root rate.

\paragraph{Unknown heterogeneous risk: the linear cost can be necessary.} Consider the uniform sign family with $B$ groups, $\kappa=1/2$ and the uniform score marginal, \emph{known} to the auditor. For $\eta\in[0,1]^{2B}$ let $P_\eta$ have loss probability $\eta_{b,h}$ on half $h$ of group $b$; the best risk is $r^*(\eta)=B^{-1}\sum_b\min\{\eta_{b,+},\eta_{b,-}\}$. Let $\mathcal P_{\rm het}(\alpha,\gamma)=\{P_\eta:\eta\in[a,1-a]^{2B},\ r^*(\eta)\ge\alpha+\gamma\}$.
\begin{theorem}[The price of unknown heterogeneity]
\label{thm:heterogeneity}
Fix $a\in(0,1/8]$, $\alpha\in[2a,1-2a]$ and $0<\delta<\beta<1$. There are $c_1,c_2,C,B_0>0$ depending only on $(a,\delta,\beta)$ such that for $B\ge B_0$ and $\gamma\le c_1/\log B$,
$c_2B/(\gamma^2\log B)\le N(\mathcal P_{\rm het},\cH)\le N_{\rm IUT}(\mathcal P_{\rm het})\le CB/\gamma^2$, whereas $N(P,\cH)=O(\sqrt B/\gamma^2)$ for every single $P\in\mathcal P_{\rm het}$.
\end{theorem}
The null and alternative priors set $\eta_{b,\pm}=m\pm s_bd_b$ with a uniform sign $s_b$ and a magnitude $d_b\in[0,D]$ from $\nu_0$ or $\nu_1$. These match $\E d^{2j}$ for $j\le L\asymp\log B$ but differ in $\E d$, and hence in $r^*$, by twice the best degree-$L$ approximation error of $\sqrt u$ on $[0,D^2]$, of order $D/L$ \citep{bernstein1914meilleure,lepski1999estimation,cai2011testing}. Under Poissonization the random signs make each group's $\chi^2$ series depend only on the even moments, so all terms up to order $L$ cancel (Appendix~\ref{app:heterogeneity}). Unlike $\ell_1$-functional estimation \citep{jiao2018minimax}, the priors must keep the null draws \emph{existentially} feasible. Together, Theorems~\ref{thm:constant} and~\ref{thm:heterogeneity} show that validity alone does not force the linear cost, while uncertainty about \emph{where} the classifier errs can force it on a rich heterogeneous class.

\section{Experiments}
\label{sec:experiments}
\subsection{The abstention-odds law}
\label{sec:kappa-law}
We audit menus of $K\in\{1,2,4,8,16\}$ intervals over $M=64$ equal-mass score cells at floors $\kappa\in\{.5,.7,.8,.9,.95\}$ (and $.3$ in Appendix~\ref{app:kappa-law}). The alternative has constant risk $\alpha+\gamma$ ($\alpha=.10$, $\gamma=.04$), so every rule has the same risk and the law lies in $\mathcal P_\gamma$ at every floor; any increase of the sample size with $K$ is therefore a pure cost of menu richness. For each $(K,\kappa)$ the exact IUT audits 250 samples at each of 15 log-spaced sizes between 150 and 15,846, and we record the 80\%-power crossing $n_{80}$ and the \emph{menu cost} $n_{80}(K)-n_{80}(1)$.

\begin{figure}[t]
 \centering\includegraphics[width=\linewidth]{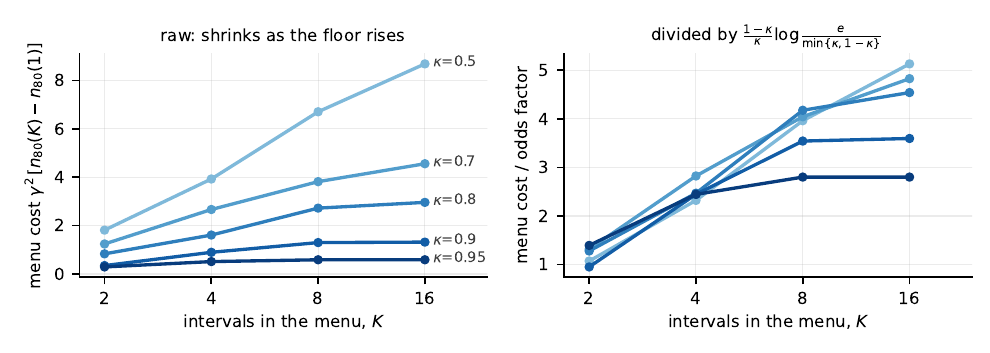}
 \caption{\textbf{Menu richness costs $K$ times the abstention odds.} Left: menu cost of the exact IUT, scaled by $\gamma^2$, for five coverage floors. At $K=4$ it falls 7.7-fold from $\kappa=.5$ to $\kappa=.95$. Right: the same costs divided by $\frac{1-\kappa}{\kappa}\log\frac{e}{\min\{\kappa,1-\kappa\}}$ lie within $2.32$--$2.82$ at $K=4$ and within $0.95$--$1.39$ at $K=2$. At $\kappa\ge.9$ the curves flatten for $K\ge8$ because the abstained region spans only $M(1-\kappa)\le6.4$ cells, so extra intervals do not enlarge the realized admissible rule set in this finite benchmark. Monte Carlo crossings from 250 audits per point.}
 \label{fig:kappa}
\end{figure}
Figure~\ref{fig:kappa} illustrates the coverage-floor dependence in Proposition~\ref{prop:bracket}(iii) and Theorem~\ref{cor:intervals}. The raw menu cost varies by almost an order of magnitude across floors. The right panel additionally divides by the theorem-motivated localized-VC logarithmic factor discussed in Appendix~\ref{app:bracket}; that factor is not used in the proofs, and the resulting variation is substantially reduced until the menu saturates. The flattening itself is the localization mechanism at work: only rules that differ on the small abstained region compete. The IUT's level is not at issue here (it is exact); boundary checks are reported below.

\subsection{Sign families: the IUT against the information envelope}
We fix $B=256$ ($2^{256}$ rules), $p=.25$, $\kappa=.5$, $k\in\{1,4,16,64,256\}$ and $\gamma\in\{.04,.02\}$. The exact dynamic program audits 1,500 samples per grid point. The comparator is the Neyman--Pearson curve for $P_0^n$ against the uniform feasible mixture; its population version bounds the power of every valid audit, so it is an information envelope, not an audit.
\begin{figure}[t]
 \centering\includegraphics[width=\linewidth]{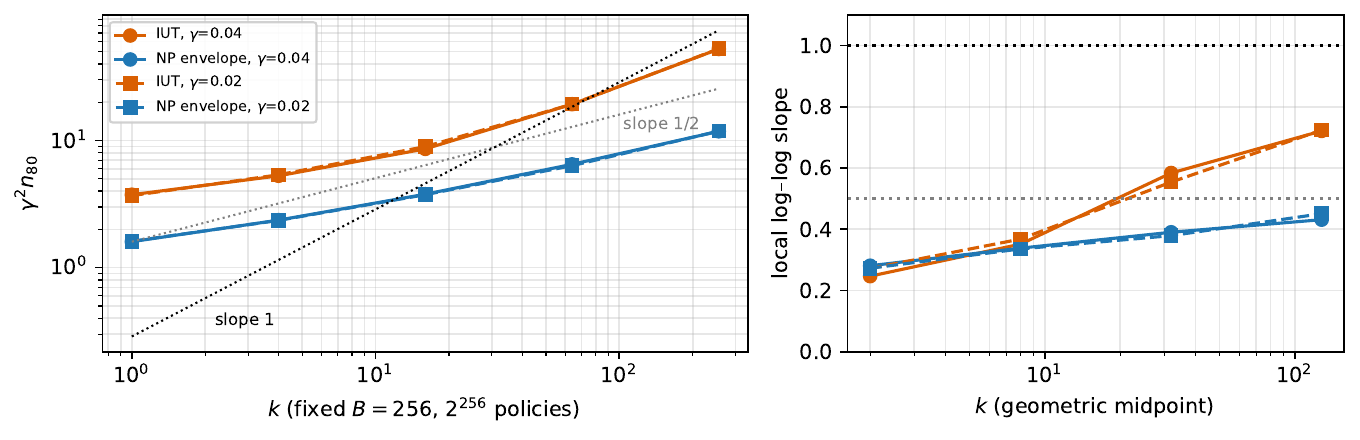}
 \caption{\textbf{The IUT and the information envelope have different scales} ($B=256$). Left: $\gamma^2n_{80}$ at two gaps; the curves nearly coincide, as $\gamma^{-2}$ scaling predicts. Right: local log--log slopes in $k$. The IUT's slope rises from 0.25 to 0.72 towards the proven exponent 1, the envelope's from 0.28 to 0.43 towards $1/2$; the range is pre-asymptotic. Crossings are in Table~\ref{tab:scaling} (Appendix~\ref{app:large-simulation}).}
 \label{fig:scaling}
\end{figure}
At $\gamma=.04$ the IUT's crossing rises from 2,342 to 32,796 samples as $k$ grows from 1 to 256, and the envelope's from 1,001 to 7,384; halving the gap multiplies all crossings by 3.91--4.17. Fitting $\gamma^2n_{80}\approx a_0+a_1\sqrt k+a_2k$ gives $a_2\approx0.13$ for the IUT at both gaps and $a_2\approx0$ for the envelope, as Theorems~\ref{thm:separation}--\ref{thm:iut-rate} predict. Ten boundary-null checks, one per $(k,\gamma)$ with 20,000 audits each, give rejection rates of 0--4.7\%. Across the simulated $k$ values the gap to the envelope is a factor of 2.2--4.4. In the separate fixed-degree ($L=2$) heterogeneity-envelope experiment, the heterogeneous envelope overtakes the known-reference one only at $B\approx2.6\times10^5$ (Appendix~\ref{app:het-envelope}).

\subsection{Auditing the auditors: VLM-judge reliability}
Automated VLM judges are increasingly used to automate evaluation, and a reliability claim is naturally existential: \emph{some} confidence gate should retain enough evaluations while keeping serious judge errors below a target. We use the public score-token log-probabilities and human scores released by \citet{kumar2026vlmjudge}. For each item let $L=\1\{|\hat y-y_{\rm human}|>1\}$, matching the source study's $\pm1$ scoring criterion; a non-1--5 parse is counted as a serious error, with a drop-invalid sensitivity analysis in the Supplementary Note. From the five score-token probabilities we predeclare three label-free confidence channels---maximum probability, negative entropy and top-two margin---and cross each with target coverages $.65$ and $.90$, giving six gates at contract floor $\kappa=.60$.

For each judge we use ten deterministic three-way resplits. Thresholds are frozen from calibration scores only; audit-A selects the midpoint of its exact fixed-marginal hiding window $I=[\alpha_H,\alpha_{\rm ind})$, and untouched audit-B tests transfer. Thus $W_{\rm hide}>0$ means no individual gate is feasible at some target even though feasible alternatives can convexify to the empirical reference. Table~\ref{tab:vlm-judge} reports the result.
\begin{table}[t]
\caption{\textbf{Auditing the auditors.} Held-out exact-hiding windows for six predeclared confidence gates at $\kappa=.60$ and serious judge error $L=\1\{|\hat y-y_{\rm human}|>1\}$. Policies are frozen on a disjoint calibration third. ``pos. A/B'' counts positive windows over audit-A and audit-B; ``A$\to$B'' asks whether the midpoint contract selected on A remains hidden on untouched B. Widths are median audit-B windows in percentage points.}
\label{tab:vlm-judge}
\centering\footnotesize
\setlength{\tabcolsep}{3.7pt}
\begin{tabular}{@{}llrrr@{}}
\toprule
Target & judge & pos. A/B & A$\to$B & med. $W_B$ \\
\midrule
MLLM-Judge & LLaVA-Critic-7B & 20/20 & 10/10 & 4.97 \\
MLLM-Judge & Gemini 2.5 Flash & 20/20 & 10/10 & 5.91 \\
MLLM-Judge & Phi-4 reasoning vision & 20/20 & 8/10 & 4.83 \\
Polaris (multi-annotator) & LLaVA-Critic-7B & 20/20 & 3/10 & 0.33 \\
\bottomrule
\end{tabular}
\end{table}

On MLLM-Judge, every one of the 60 split-specific windows is positive and 28/30 midpoint contracts transfer. To test whether this is merely noisy single-annotator supervision, we repeat the same LLaVA-Critic protocol on Polaris, where the released target averages at least three annotators and the same judge is much more accurate \citep{kumar2026vlmjudge}. All 20 windows remain positive, but the median held-out width falls from 4.97 to 0.33 percentage points and midpoint transfer falls from 10/10 to 3/10. On a representative Polaris audit-B split, singleton families have zero window, 60\% of two-gate and 90\% of three-gate subfamilies hide, and every subfamily of size at least four hides (Supplementary Note). The mechanism therefore persists under cleaner supervision, while its magnitude and split stability are population dependent. These are finite-target mechanism diagnostics, not an instance-specific minimax verification of Theorem~\ref{thm:heterogeneity}.

\subsection{Frozen classifier certificates and family geometry}
We also treat frozen classifier outputs as populations: CIFAR-10/100 ResNet-18, ImageNetV2 ViT-B/16 and CMU-MOSEI MLP outputs \citep{krizhevsky2009cifar,recht2019imagenet,zadeh2018multimodal}. Calibration fixes score cells and the audit half supplies labels. At $\kappa=.9$, the exact IUT has no observed boundary rejection above 1.8\% and is the most powerful certificate compared except for one $n=500$ CIFAR-100 point within Monte Carlo error; Bonferroni needs several times more data, structured DKW is vacuous here, and a score-agnostic total-loss certificate fails when total error is small relative to $1-\kappa$ (Table~\ref{tab:semisyn-full}, Appendix~\ref{app:structured-comparison}).

The constant-risk premise of Theorem~\ref{thm:constant} is strongly contradicted: a cell--loss independence diagnostic rejects in every one of 200 audits of size 500 on each population. More directly, a separate six-policy, three-way held-out geometry experiment freezes gates from calibration scores only. All 60 audit-A and all 60 audit-B windows are positive; 47/60 midpoint contracts selected on A transfer to untouched B. Singleton subfamilies have zero hiding, 96.7\% of pairs hide, and every subfamily of size at least three hides (Supplementary Note, Section S5). Thus the same feasible-mixture mechanism appears on standard classifier families, without claiming that these populations instantiate Theorem~\ref{thm:heterogeneity}'s asymptotic hard class.

Finally, as a valid non-EFC reference-free competitor we implement cross-fit universal inference \citep{wasserman2020universal}. It has no observed boundary rejections, but at $\Delta=.05$ its first 80\%-power grid point is $n=10{,}000$--$50{,}000$, versus 500--2,500 for the IUT; at $\Delta=.02$ it does not reach 80\% power by $50{,}000$ (Table~\ref{tab:nonefc}, Appendix~\ref{app:nonefc}). This evaluates one practical non-EFC construction, not all non-EFC audits.

\section{Related work and limitations}
\label{sec:related}
Risk control, Learn-then-Test and Pareto testing certify selected rules \citep{bates2021riskcontrol,angelopoulos2024conformal,angelopoulos2025learntest,laufer2023efficiently}; model falsification, subgroup auditing and PAC verification address adjacent goals \citep{muller2025models,cherian2024fairness,mutreja2023pac}. Intersection--union validity is classical \citep{berger1982multiparameter,berger1996bioequivalence,lehmann2005testing}; for a fixed number of normal means the IUT is the likelihood-ratio test and non-monotone tests can be uniformly more powerful \citep{lehmann1952testing,berger1989uniformly}, at the price of non-monotone rejection regions that can yield counterintuitive inferences \citep{perlman1999emperor}. Moment-inequality and intersection-bound tests \citep{andrews2010inference,chernozhukov2013intersection,romano2014practical,bugni2015specification} ask whether some parameter satisfies a set of inequalities; with many inequalities, as in reality checks \citep{white2000reality,hansen2005test}, the cost is logarithmic in their number \citep{chernozhukov2019inference} because that null is an intersection, whereas ours is a union. Inference on winners corrects the selection bias of an empirical optimum \citep{andrews2024inference}, the bias that EFC certificates do not correct; sequential tests of whether some arm lies below a threshold address a related existential question \citep{kaufmann2018sequential}; confidence sequences give anytime versions of the IUT \citep{howard2021time} (Appendix~\ref{app:iut}). Universal inference gives reference-free composite tests \citep{wasserman2020universal}; recent work on testable hypotheses and safe anytime-valid inference provides broader characterizations and tools for composite testing \citep{larsson2026testable,ramdas2023savi}. In property testing, unions of $d$ intervals are testable from $O(\sqrt d)$ samples when the input distribution is known but need $\Omega(d/\log d)$ distribution-free \citep{kearns2000testing,balcan2012active,blais2021vc}; related interval and flat structure also appears in identity testing of structured distributions \citep{diakonikolas2015testing}, and optimal risk can be estimated faster than learned \citep{kong2018estimating}. Minimax $L_1$-distance estimation and tolerant distribution testing exhibit related linear-over-log sample-complexity phenomena \citep{jiao2018minimax,canonne2022price}. Recent work studies calibrated uncertainty for VLM judges and releases the score-token features used in our auditing application \citep{kumar2026vlmjudge}. Our null is a union of convex sets of laws with ratio-type constraints, losses are noisy, and an unknown input distribution alone does not force the linear scale (Theorem~\ref{thm:constant}), whereas unknown conditional risk can force it on the heterogeneous class of Theorem~\ref{thm:heterogeneity}. The Supplementary Note relates the one-observation geometry to convex-set testing \citep{goldenshluger2015testing}.

\paragraph{Limitations.} Losses are binary, the contract has one coverage floor, and data are i.i.d.; distribution shift is outside our scope. For each fixed coverage floor, the interval-menu bounds are worst-case and match up to logarithms in the stated small-gap, large-$K$ regime; the lower bound is asymptotic in $K$; the sharp testing--estimation spectrum and the constant-risk result are proved on sign families; an $\Omega(d)$ bound for every VC-$d$ family is open. Theorem~\ref{thm:constant} is an existence result with an untuned tester. The natural-family and VLM-judge studies are finite-target mechanism diagnostics, not instance-specific minimax lower bounds for Theorem~\ref{thm:heterogeneity}; the six VLM gates are illustrative and use released benchmark outputs rather than production traffic. The non-EFC experiment evaluates one universal-inference construction rather than all possible non-EFC audits. Nonrejection certifies no rule.

\FloatBarrier
\phantomsection\label{main-end}
\subsection*{AI use statement}
Generative AI was used to improve the clarity and presentation of manuscript claims, assist with software implementation and debugging, and provide critical feedback on claims and arguments during manuscript development. AI-generated suggestions were treated as editorial and technical assistance rather than as evidence or independent validation. The authors are fully responsible for the final claims, proofs, code, citations, analyses, and submission.
\subsection*{Ethics statement}
The empirical studies reuse frozen benchmark model outputs and released human annotations and collect no new human-subject data. The work concerns statistical auditability and does not introduce a clinical decision system or an automated evaluator for deployment. Population interpretations require a sampling model appropriate to the intended setting.
\subsection*{Reproducibility statement}
The supplement contains this manuscript's figures, analysis scripts, exact rational certificates, frozen internal inputs, and derived VLM-audit summaries; the public third-party VLM feature files are referenced by source and are not redistributed. The analyses for Section~\ref{sec:knowledge} and Figure~\ref{fig:heterogeneity} are provided with their tests. The release README maps claims to code and distinguishes recomputed results from retained outputs. Appendix~\ref{app:experiments} and the Supplementary Note give protocols and limitations.

\clearpage
\appendix

\section{Roadmap and notation}
\label{app:notation}
\paragraph{Local notation.} A few symbols are reused locally with meanings fixed by context: $L$ is the loss, but the polynomial degree in Theorem~\ref{thm:heterogeneity}; $M$ is the number of cells, but the error count in the proofs of Theorem~\ref{thm:iut-rate} and Proposition~\ref{prop:efc-slack}; $m$ is a prior midpoint in Theorem~\ref{thm:heterogeneity} and the number of selected parts in the $q$-ary construction; $K$ counts intervals, and $K(Q,Q')$ is the overlap kernel of Lemma~\ref{thm:overlap}; $\lambda$ is $\gamma^2/[p(1-p)]$ in Section~\ref{sec:overlap} and a Poisson mean in Appendix~\ref{app:heterogeneity}.

\paragraph{Where the main results are proved.}
This appendix contains everything the main theorems use. EFC certificates, Theorem~\ref{thm:constant}, Proposition~\ref{prop:bracket} (including the localization in part (iii)), Theorem~\ref{cor:intervals} (Appendix~\ref{app:intervals}) and Theorem~\ref{thm:heterogeneity}: Appendix~\ref{app:knowledge}. Theorem~\ref{thm:iut-rate}, the Rademacher formula and Proposition~\ref{thm:adaptive}: Appendix~\ref{app:efficiency}. Theorem~\ref{thm:separation} and Propositions~\ref{prop:singleton}, \ref{prop:weight-profile} and~\ref{prop:combinatorial}: Appendix~\ref{app:separation}. Lemma~\ref{thm:overlap}: Appendix~\ref{app:overlap}. Proposition~\ref{prop:iut} and the dynamic programs: Appendix~\ref{app:iut}. The witness power ceiling (Corollary~\ref{cor:powerceiling}) and the collision bound (Theorem~\ref{thm:collision}): Appendix~\ref{app:aux}. Experimental protocols and all tables, including the abstention-odds simulation and the universal-inference audit: Appendix~\ref{app:experiments}. A separate Supplementary Note in the supplementary material collects one-observation geometry (exact projections, Hall--TV formulas, the eight-atom pairwise example and natural hiding windows) that the main theorems do not use.

\begin{center}
\scriptsize
\setlength{\tabcolsep}{2pt}
\begin{tabular}{@{}p{1.0in}p{3.45in}p{.65in}@{}}
\toprule
Symbol & Meaning & Sample-computable?\\
\midrule
$P_0$ & fixed infeasible reference / audit alternative & reference-dependent\\
$Q$ & feasible population law under the existential null & no\\
$\cA$ & predeclared family of acceptance policies $A$ & yes\\
$\cH(\cA)$ & laws for which at least one $A\in\cA$ meets the contract & implicit\\
$\kappa,\alpha$ & coverage floor and conditional-risk target & yes\\
$R_n=\E_{Q\sim\Pi}[Q^n]$ & product mixture induced by a prior over feasible laws & only with witness\\
$\Xi(Q,Q')$ & likelihood-ratio overlap kernel & only with components\\
$d_{\rm eff}$ & participation-ratio summary inside the matched sign construction & diagnostic proxy\\
$N(P_0,\cH;\delta,\beta)$ & samples needed for level $\delta$ and power $\beta$ & no\\
$L_{\mathrm{IUT}}$ & reference-free lower risk frontier for an existential contract & yes\\
EFC & audit that never rejects while some policy is empirically feasible & property\\
$\mathcal P_{\rm c},\mathcal P_{\rm het}$ & constant-risk and heterogeneous-risk alternative classes & power classes\\
\bottomrule
\end{tabular}
\end{center}

\section{EFC certificates, reference-free testing, and unknown heterogeneity}
\label{app:knowledge}

\subsection{Which certificates are EFC}
\label{app:efc}
\begin{lemma}[Clopper--Pearson bounds bracket the empirical proportion]
\label{lem:cp-median}
Let $X\sim\mathrm{Bin}(t,q)$ with observed value $x$. For $\zeta\le1/2$ the one-sided Clopper--Pearson upper bound at error $\zeta$ is at least $x/t$ and the lower bound is at most $x/t$.
\end{lemma}
\begin{proof}
When the mean $tq$ is an integer, it is a median of $\mathrm{Bin}(t,q)$ \citep{kaas1980mean}. At $q=x/t$ we therefore have $P_q(X\le x)\ge1/2\ge\zeta$ and $P_q(X\ge x)\ge1/2\ge\zeta$. The upper bound is the largest $q$ with $P_q(X\le x)\ge\zeta$, so it is at least $x/t$; the lower bound is symmetric.
\end{proof}
Let $A$ be empirically feasible: $T_A\ge\kappa n$ and $E_A\le\alpha T_A$.
\begin{itemize}
\item \emph{Exact IUT and policywise Bonferroni.} By Lemma~\ref{lem:cp-median}, $U_A\ge T_A/n\ge\kappa$, so $A$ passes the coverage screen, and $L_A\le E_A/T_A\le\alpha$. Hence $L_{\rm IUT}\le\alpha$ and the audit does not reject. Bonferroni replaces $\zeta$ by $\zeta/|\cA|\le1/2$ and is covered by the same argument.
\item \emph{DKW and Rademacher certificates} (Appendices~\ref{app:dkw-baseline}--\ref{app:rademacher-baseline}). Their screen $c_n(A)\ge\kappa-\epsilon$ is passed by $A$, and their lower bound $[m_n(A)-\epsilon]_+/\min\{1,c_n(A)+\epsilon\}$ is at most $m_n(A)/c_n(A)=E_A/T_A\le\alpha$.
\item \emph{Total-loss certificate.} Writing $\hat e$ for the empirical loss rate, $\hat e\le (E_A+n-T_A)/n\le1-(1-\alpha)T_A/n\le1-\kappa(1-\alpha)$. Hence $(\hat e+\kappa-1)/\kappa\le\alpha$; its lower confidence version is smaller by Lemma~\ref{lem:cp-median}, so it does not reject.
\end{itemize}
Part (i) of Theorem~\ref{thm:iut-rate} therefore applies to every certificate compared in Section~\ref{sec:experiments}.

\subsection{Proof of Theorem~\ref{thm:constant}}
\label{app:constant}
Throughout, laws are identified with their pushforwards to $[M]\times\{0,1\}$ (cell index and loss). This loses nothing for the null: every $A\in\cA$ is a union of cells, so the contract depends on $Q$ only through this pushforward, and the pushforward of a feasible law is feasible for the induced family. Products on the original space push forward to products.

\begin{lemma}[Feasibility forces dependence]
\label{lem:dependence}
Let $Q$ be feasible with witness $A$, and let $e=Q(L=1)>\alpha$. Then $\TV(Q,R)\ge(e-\alpha)Q(A)/3$ for every product law $R=\nu\otimes\Bern(q)$ on $[M]\times\{0,1\}$.
\end{lemma}
\begin{proof}
Let $t=\TV(Q,R)$. If $t\ge e$ or $t\ge Q(A)$ the claim holds because $(e-\alpha)Q(A)/3<\min\{e,Q(A)\}$. Otherwise, comparing the events $\{L=1\}$, $\{S\in A\}$ and $\{S\in A,L=1\}$ gives $q\ge e-t$, $\nu(A)\ge Q(A)-t$ and $q\nu(A)\le Q(A,L=1)+t\le\alpha Q(A)+t$. Thus
$\alpha Q(A)+t\ge(e-t)(Q(A)-t)\ge eQ(A)-t(e+Q(A))\ge eQ(A)-2t$, so $3t\ge(e-\alpha)Q(A)$.
\end{proof}

\paragraph{Independence tester.} For distributions on $[M]\times[2]$, \citet{diakonikolas2016new} give a tester that, from $O(\sqrt{M}/\epsilon^2+M^{2/3}/\epsilon^{4/3})$ samples, accepts every product law and rejects every law $\epsilon$-far in total variation from all product laws, each with probability at least $2/3$ (their $\ell_1$ parameter is $2\epsilon$). Taking the majority of $O(\log(1/\epsilon'))$ independent runs reduces both error probabilities to $\epsilon'$. Denote the amplified output by $\Psi\in\{0,1\}$, with $\Psi=1$ meaning ``dependent''.

\paragraph{The audit.} Let $\epsilon'=\min\{\delta,(1-\beta)/2\}$ and $\epsilon=\kappa\gamma/12$. Reject feasibility if and only if $\hat e\ge\alpha+\gamma/2$ and $\Psi=0$, where $\hat e$ is the empirical loss rate and $\Psi$ uses the independence tester at distance $\epsilon$ and error $\epsilon'$. Neither the score marginal nor $P(L=1)$ is an input. Let $n_0$ be the larger of $8\log(1/\epsilon')/\gamma^2$ and the amplified tester's sample size; for $n<n_0$ the audit does not reject, and for $n\ge n_0$ it applies the rule above to the first $n_0$ observations for the tester.

\paragraph{Level over the unrestricted null.} Let $Q\in\cH(\cA)$ and $e=Q(L=1)$. If $e\le\alpha+\gamma/4$, Hoeffding's inequality gives $Q(\hat e\ge\alpha+\gamma/2)\le\exp(-n\gamma^2/8)\le\epsilon'$ once $n\ge8\log(1/\epsilon')/\gamma^2$. If $e>\alpha+\gamma/4$, Lemma~\ref{lem:dependence} and $Q(A)\ge\kappa$ make $Q$ at least $\kappa\gamma/12$-far from every product law, so $Q(\Psi=0)\le\epsilon'$. In both cases the rejection probability is at most $\epsilon'\le\delta$. No assumption beyond i.i.d. sampling and frozen cells is used.

\paragraph{Power.} Let $P\in\mathcal P_{\rm c}(\alpha,\gamma)$. Its pushforward is a product law, so $P(\Psi=1)\le\epsilon'$, and $P(\hat e<\alpha+\gamma/2)\le\exp(-n\gamma^2/2)\le\epsilon'$. The power is at least $1-2\epsilon'\ge\beta$. Collecting sample sizes, and using $M\ge2$ and $\kappa\le1$ to absorb the $\log(1/\epsilon')/\gamma^2$ term, gives the displayed bound with a constant depending on $(\delta,\beta)$.

\paragraph{Rates on the sign family.} With $M=2k$ and $\kappa=1/2$, $M^{2/3}\gamma^{-4/3}\le M^{1/2}\gamma^{-2}$ if and only if $\gamma\le M^{-1/4}$, so $N(\mathcal P_{\rm c},\cH)=O(\sqrt k/\gamma^2)$ in that regime. The uniform reference $P_0$ with $p=\alpha+\gamma$ lies in $\mathcal P_{\rm c}$, so $N(\mathcal P_{\rm c},\cH)\ge N(P_0,\cH)=\Omega(\sqrt k/\gamma^2)$ by Proposition~\ref{prop:weight-profile} with uniform weights, and Theorem~\ref{thm:iut-rate}(i) gives the EFC lower bound at the same law. Conversely, the IUT upper bound of Appendix~\ref{app:iut-rate} extends to every $P\in\mathcal P_{\rm c}$ on the sign family: all policies have risk $P(L=1)\ge\alpha+\gamma$, and on the uniform-concentration event over the $2^k$ policies every policy passing the coverage screen has $T_A\ge\kappa n/2$ and $E_A-\alpha T_A\ge\gamma T_A-C\sqrt{nk}$, so $N_{\rm IUT}(\mathcal P_{\rm c})=O(k/(\kappa\gamma)^2)$; this justifies the entry in the third row of Table~\ref{tab:regimes}. Whether the $M^{2/3}\gamma^{-4/3}$ term is necessary for $\gamma>M^{-1/4}$ is left open.

\subsection{Proof of Proposition~\ref{prop:bracket}}
\label{app:bracket}
\paragraph{Two inversions.} For $X\sim\mathrm{Bin}(t,q)$ observed at $x$, Hoeffding's inequality inverted at error $\zeta$ gives that the one-sided Clopper--Pearson upper bound is at most $x/t+\sqrt{\log(1/\zeta)/(2t)}$ and the lower bound is at least $x/t-\sqrt{\log(1/\zeta)/(2t)}$.

\paragraph{Uniform deviations.} The classes $\cF_1=\{(s,l)\mapsto\1_A(s):A\in\cA\}$ and $\cF_2=\{(s,l)\mapsto l\,\1_A(s):A\in\cA\}$ have VC dimension at most $d$; a set shattered by $\cF_2$ consists of points with $l=1$ whose scores are shattered by $\cA$. The VC bound for Boolean classes \citep[Chapter~8]{vershynin2018high} gives $\E\sup_{f\in\cF_i}|P_nf-Pf|\le C_0\sqrt{d/n}$, and the bounded-differences inequality gives, with probability at least $1-(1-\beta)/2$ for both classes simultaneously,
\begin{align*}
 &\sup_{A\in\cA}|c_n(A)-c(A)|\le\epsilon_n,\qquad \sup_{A\in\cA}|m_n(A)-m(A)|\le\epsilon_n,\\
 &\epsilon_n=C_0\sqrt{d/n}+\sqrt{\log(8/(1-\beta))/(2n)},
\end{align*}
where $c_n(A)=T_A/n$, $m(A)=P(S\in A,L=1)$ and $m_n(A)=E_A/n$.

\paragraph{Part (i).} Let $P\in\mathcal P_\gamma(\cA)$ and work on the event above. Suppose $A$ passes the coverage screen, $U_A\ge\kappa$. By the first inversion $c_n(A)\ge\kappa-\sqrt{\log(2/\delta)/(2n)}$, so $c(A)\ge\kappa-\sqrt{\log(2/\delta)/(2n)}-\epsilon_n\ge\kappa-\gamma$ provided $\sqrt{\log(2/\delta)/(2n)}+\epsilon_n\le\gamma$. The definition of $\mathcal P_\gamma$ then gives $m(A)\ge(\alpha+\gamma)c(A)$, and since $c(A)\ge\kappa-\gamma\ge\kappa/2$,
\[
 \frac{E_A}{T_A}\ge\frac{m(A)-\epsilon_n}{c(A)+\epsilon_n}
 \ge\alpha+\gamma-\frac{2\epsilon_n}{c(A)}\ge\alpha+\gamma-\frac{4\epsilon_n}{\kappa}.
\]
If also $\epsilon_n\le\kappa/4$, then $T_A\ge n\kappa/4$, and the second inversion gives
$L_A\ge\alpha+\gamma-4\epsilon_n/\kappa-\sqrt{2\log(2/\delta)/(n\kappa)}$.
All three conditions, and $L_A>\alpha$, hold once $n\ge C\{d+\log(2/\delta)+\log(4/(1-\beta))\}/(\kappa\gamma)^2$ for an absolute $C$. Then every screened policy has $L_A>\alpha$, and if no policy is screened $L_{\rm IUT}=+\infty$; either way the IUT rejects. Its power is at least $1-(1-\beta)/2\ge\beta$.

\paragraph{Part (ii).} Choose a score law under which $c(A_0)\ge\kappa$ and let $P$ have constant risk $\alpha+\gamma$. Every policy has risk $\alpha+\gamma$, so $P\in\mathcal P_\gamma(\cA)$. Let $Q$ equal $P$ except that the risk inside $A_0$ is $\alpha$; then $A_0$ witnesses feasibility of $Q$. Since $\KL(P\Vert Q)=c(A_0)\,\mathrm{kl}(\alpha+\gamma\Vert\alpha)\le\gamma^2/\{\alpha(1-\alpha)\}$, Pinsker's inequality gives $\TV(P^n,Q^n)\le\{n\gamma^2/(2\alpha(1-\alpha))\}^{1/2}$. A level-$\delta$ audit with power $\beta$ at $P$ needs $\TV(P^n,Q^n)\ge\beta-\delta$, hence $n\ge2(\beta-\delta)^2\alpha(1-\alpha)/\gamma^2$.

\paragraph{Part (iii): localization to the smaller side of the floor.}
We use Vapnik's relative-deviation inequalities: for a class $\cF$ of $\{0,1\}$-valued functions with growth function $S_\cF$, with probability at least $1-\eta$,
$\sup_{f}(Pf-P_nf)/\sqrt{Pf}\le\epsilon$ and $\sup_f(P_nf-Pf)/\sqrt{P_nf}\le\epsilon$, where $\epsilon=2\{(\log S_\cF(2n)+\log(8/\eta))/n\}^{1/2}$ \citep[Theorem~5.1]{boucheron2005theory}. Solving the second inequality for $P_nf$ gives, for every $f\in\cF$,
\begin{equation}
 |P_nf-Pf|\le\epsilon\sqrt{Pf}+\epsilon^2.
 \label{eq:relative-deviation}
\end{equation}
Apply this to $\cF=\{\1_A\}\cup\{\1_{A^c}\}\cup\{l\1_A\}\cup\{l\1_{A^c}\}$ ($A\in\cA$). Each of the four classes has VC dimension at most $d$, so $\log S_\cF(2n)\le\log4+d\log(2en/d)$ for $n\ge d$. Take $\eta=(1-\beta)/4$, and let Hoeffding's inequality give $|\hat e-e|\le\{\log(8/(1-\beta))/(2n)\}^{1/2}$ for the overall loss rate with probability at least $1-(1-\beta)/4$. On the intersection $\Omega$ of these events, which has probability at least $1-(1-\beta)/2$, every $A\in\cA$ satisfies, with $c=c(A)$, $m=m(A)$ and $c_n,m_n$ their empirical versions,
\begin{align}
 &|c_n-c|\le\epsilon\sqrt{\min\{c,1-c\}}+\epsilon^2,\tag{D1}\label{eq:D1}\\
 &|m_n-m|\le\min\big\{\epsilon\sqrt c+\epsilon^2,\ |\hat e-e|+\epsilon\sqrt{1-c}+\epsilon^2\big\},\tag{D2}\label{eq:D2}
\end{align}
using $c_n-c=-(c_n(A^c)-c(A^c))$, $m_n-m=(\hat e-e)-(m_n(A^c)-m(A^c))$ and $m(A^c)\le1-c$.

\emph{Clopper--Pearson widths.} Let $X\sim\mathrm{Bin}(t,q)$ be observed at $x$, $\hat q=x/t$. Since $\partial_q\mathrm{kl}(\hat q,q)=(q-\hat q)/\{q(1-q)\}$, we have $\mathrm{kl}(\hat q,q)\ge(q-\hat q)^2/\{2\max_{s\in[\hat q,q]}s(1-s)\}\ge(q-\hat q)^2/(2\min\{q,1-\hat q\})$ for $q\ge\hat q$. The Chernoff bound $P_q(X\le x)\le\exp\{-t\,\mathrm{kl}(\hat q,q)\}$ therefore gives, for the upper bound $U$ at error $\zeta$,
\begin{equation}
 U-\hat q\le\{2\min\{U,1-\hat q\}\log(1/\zeta)/t\}^{1/2},
 \label{eq:cp-width}
\end{equation}
and the lower bound satisfies $\hat q-L\le\{\log(1/\zeta)/(2t)\}^{1/2}$ as before.

Write $\ell=\log(2/\delta)$, $\sigma=\min\{\kappa,1-\kappa+\gamma\}$ and $u=\epsilon^2+\ell/n$. Suppose
\begin{equation}
 u\le c_0\gamma^2\kappa/(1-\kappa+\gamma),\qquad |\hat e-e|\le\gamma/48,\qquad n\ge8\ell/(\kappa\gamma^2),
 \qquad c_0=96^{-2}.
 \label{eq:local-conditions}
\end{equation}
Because $\kappa^2\le\kappa\le1-\kappa+\gamma$ when $\sigma=\kappa$, the first condition implies $u\le c_0\gamma^2/\sigma$, and since $\sigma\ge\gamma$ (as $\gamma\le\kappa/2$) also $u\le c_0\gamma$.

\emph{Step 1: the screen admits only policies of coverage at least $\kappa-\gamma$.} Suppose $U_A\ge\kappa$ but $t\defeq\kappa-c(A)>\gamma$. By \eqref{eq:cp-width} applied to the $n$ coverage trials, $\kappa-c_n\le\{2v\ell/n\}^{1/2}$ with $v=\min\{U_A,1-c_n\}$, and by \eqref{eq:D1}, $c_n-c\le\epsilon\sqrt{\min\{c,1-c\}}+\epsilon^2$. Now $\min\{c,1-c\}\le\min\{\kappa,1-\kappa+t\}\le(t/\gamma)\sigma$ because $t\ge\gamma$. Using $\epsilon\sqrt x\le(x+\epsilon^2)/2$ in \eqref{eq:D1} and $\sqrt{2U\ell/n}\le U/2+\ell/n$ in \eqref{eq:cp-width} gives $1-c_n\le2(1-\kappa+t)+2u$ and $U_A\le4\kappa+4u$, so $v\le4(t/\gamma)\sigma+4u$. Adding the two displays,
\[
 t\le\sqrt{8(t\sigma/\gamma)u}+\sqrt{8}\,u+\sqrt{(t\sigma/\gamma)u}+u
 \le(\sqrt8+1)\{\sqrt{c_0}+c_0\}\,t<t,
\]
using $(t\sigma/\gamma)u\le c_0t\gamma\le c_0t^2$ and $u\le c_0\gamma\le c_0t$. This contradiction proves the claim.

\emph{Step 2: every screened policy is certified infeasible.} Let $U_A\ge\kappa$, so $c=c(A)\ge\kappa-\gamma\ge\kappa/2$, and $P\in\mathcal P_\gamma(\cA)$ gives $m\ge(\alpha+\gamma)c$. With $\varrho=\alpha+\gamma/2$ we have $m-\varrho c\ge\gamma c/2$, and by \eqref{eq:D1}--\eqref{eq:D2}
\[
 |(m_n-\varrho c_n)-(m-\varrho c)|\le\Delta\defeq\min\big\{2\epsilon\sqrt c+2\epsilon^2,\ |\hat e-e|+2\epsilon\sqrt{1-c}+2\epsilon^2\big\}.
\]
If $\kappa\le1/2$, then $\kappa/(1-\kappa+\gamma)\le2\kappa$, so $\epsilon^2\le2c_0\gamma^2\kappa\le\gamma^2c/256$ and the first term gives $\Delta\le\gamma c/8+\gamma c/8$. If $\kappa>1/2$, then $c\ge1/4$, $1-c\le1-\kappa+\gamma$ and $\epsilon^2\le c_0\gamma^2/(1-\kappa+\gamma)$, so the second term gives $\Delta\le\gamma/48+\gamma/48+\gamma/48\le\gamma c/4$. In both cases $m_n-\varrho c_n\ge\gamma c/4>0$; also $c_n\ge c/2\ge\kappa/4$ by \eqref{eq:D1}. Hence $E_A/T_A\ge\alpha+\gamma/2$ with $T_A\ge n\kappa/4$, and the Clopper--Pearson lower bound satisfies $L_A\ge\alpha+\gamma/2-\{2\ell/(n\kappa)\}^{1/2}>\alpha$ by the last condition in \eqref{eq:local-conditions}.

\emph{Sample size.} On $\Omega$, every screened policy has $L_A>\alpha$, so the IUT rejects; its power is at least $1-(1-\beta)/2\ge\beta$. The conditions \eqref{eq:local-conditions} hold whenever
$n\ge C\{(1-\kappa+\gamma)\,[d\log(2en/d)+\log(8/(1-\beta))]+\log(2/\delta)+\log(8/(1-\beta))\}/(\kappa\gamma^2)$
for an absolute constant $C$, using $(1-\kappa+\gamma)\le1$ and $\kappa\le1$ for the terms without $d$. The left side minus the right side is increasing in $n$, which proves part (iii).

\emph{Remark on the logarithm.} The factor $\log(2en/d)$ comes from the growth function. Localized VC bounds for sets of small probability \citep{gine2006concentration} suggest that it can be replaced by $\log(e/\min\{\kappa,1-\kappa\})$; the experiments of Section~\ref{sec:kappa-law} are consistent with this form, but we do not use it.

\subsection{Proof of Theorem~\ref{cor:intervals}}
\label{app:intervals}
\paragraph{Upper bound.} Unions of at most $K$ intervals have VC dimension $d=2K$. Part (iii) of Proposition~\ref{prop:bracket} with $1-\kappa+\gamma\le2(1-\kappa)$ (as $\gamma\le1-\kappa$) gives power at least $\beta$ once $n\ge C\{K(1-\kappa)\log(2en/K)+1\}/(\kappa\gamma^2)$, with $C$ depending on $(\delta,\beta)$. This condition holds for $n\ge C'\{K(1-\kappa)\log(e/(\kappa\gamma))+1\}/(\kappa\gamma^2)$ because the logarithm of such an $n/K$ is at most a constant times $\log(e/(\kappa\gamma))$.

\paragraph{Lower bound: a padded construction.} Fix $B=K-1$ and let $\gamma_4$, $D$, $m$, $L$, $\nu_0,\nu_1$ be as in Steps 1--2 of the proof of Theorem~\ref{thm:heterogeneity} with this $B$; the conditions imposed below imply those of that theorem. Scores are uniform on $[0,1]$. A \emph{construction region} $R$ of mass $\pi$ is split into $B$ consecutive groups of mass $\pi/B$, each split into a left and a right half, with loss probability $\eta_{b,\pm}=m\pm s_bd_b$ on the two halves as in Step~2. Its complement, the \emph{padding}, has a fixed conditional loss probability $\eta_{\rm pad}$ that is the same under both priors. The inner best risk is $r^*_{\rm in}(\eta)=m-B^{-1}\sum_bd_b$, and the lower halves (those with risk $m-d_b$) have total mass $\pi/2$. We distinguish two cases.

\emph{Case $\kappa\ge1/2$.} Take $\pi=2(1-\kappa)$, padding $F=[0,1-\pi)$ with $\eta_{\rm pad}=\alpha$, and $R=[1-\pi,1]$. The set $W$ formed by $F$ and the lower halves has coverage $1-\pi+\pi/2=\kappa$, is a union of at most $B+1=K$ intervals, and has risk
\[
 r(W)=\alpha+\tfrac{\pi}{2\kappa}\{r^*_{\rm in}(\eta)-\alpha\}=\alpha+\tfrac{1-\kappa}{\kappa}\{r^*_{\rm in}(\eta)-\alpha\}.
\]
Under the null prior, on the event $r^*_{\rm in}\le\alpha$ of Step~2, $W$ witnesses feasibility for $\cA_K$. Under the alternative prior, on the event $r^*_{\rm in}\ge\alpha+\gamma_4$, consider any measurable set. The conditional risk is constant on each half and on $F$. Every upper half has risk $m+d_b\ge m$, and every point of $W$ has risk at most $m$, because $m-\alpha=\E_{\nu_0}d-g/4\ge3g/4>0$. Hence, as in the proof for $\kappa=1/2$, the least risk at coverage exactly $\kappa$ is $r^*\defeq r(W)\ge\alpha+(1-\kappa)\gamma_4/\kappa$, larger coverage cannot lower it, and at coverage $\kappa-s$ with $s\le\kappa/2$ the optimal set removes mass $s$ of risk at most $m$ from $W$, so its risk is at least $r^*-s(m-r^*)/(\kappa-s)\ge r^*-2sD/\kappa$, because $m-r^*\le m-\alpha\le D$. Put $\gamma=(1-\kappa)\gamma_4/(2\kappa)$. Taking $s=\gamma$ and using $D\le1/16\le\kappa/2$, every measurable set of coverage at least $\kappa-\gamma$ has risk at least $\alpha+2\gamma-\gamma=\alpha+\gamma$: the alternative draws lie in $\mathcal P_\gamma(\cA_K)$.

\emph{Case $\kappa<1/2$.} Take $\pi=2\kappa$, $R=[0,\pi)$ and padding $H=[\pi,1]$ with $\eta_{\rm pad}=1$. The lower halves form a set $W$ of coverage $\kappa$ that is a union of at most $B$ intervals, with risk $r^*_{\rm in}$; under the null prior it witnesses feasibility. Under the alternative, the points of least risk are the lower halves, then the upper halves (risk at least $m$), then $H$ (risk $1>m+D$). As before, larger coverage cannot lower the risk and coverage $\kappa-s$ lowers it by at most $2sD/\kappa$. Put $\gamma=\gamma_4/2$ and require $D\le\kappa/2$; then every set of coverage at least $\kappa-\gamma$ has risk at least $\alpha+\gamma_4-\gamma=\alpha+\gamma$.

\paragraph{Indistinguishability.} Poissonize as in Step~3 of the proof of Theorem~\ref{thm:heterogeneity}, with mean $n'$. Counts in the padding are independent of counts in $R$ and have the same law under both priors, so the total variation between the two mixtures equals that between the mixtures of the $R$-counts. These are independent Poisson counts with means $\lambda\eta_{b,h}$ and $\lambda(1-\eta_{b,h})$ in each half, now with $\lambda=\pi n'/(2B)$, so Steps~4--6 apply verbatim with $\rho=\pi n'/\{Bm(1-m)\}$. The condition $\tau=\rho D^2\le(2L+2)/e^3$ holds whenever $\pi n'\le c\,B/(\gamma_4^2\log B)$. In the case $\kappa\ge1/2$ this reads $n'\le c\,B(1-\kappa)/(8\kappa^2\gamma^2\log B)$, and in the case $\kappa<1/2$ it reads $n'\le c\,B/(8\kappa\gamma^2\log B)$. Both are at least $c''K(1-\kappa)/(\kappa\gamma^2\log K)$ for $K\ge3$. Hence no valid audit with fewer samples has power $\beta$ over $\mathcal P_\gamma(\cA_K)$.

\paragraph{Conditions.} Theorem~\ref{thm:heterogeneity} needs $\gamma_4\le c_1/\log B$, the case $\kappa<1/2$ needs $D\le\kappa/2$, and $D=\gamma_4/E_L\le\gamma_4L/c_*$ with $L=\lceil\log B\rceil$. In the case $\kappa\ge1/2$, $\gamma_4=2\kappa\gamma/(1-\kappa)\le2\gamma/(1-\kappa)$; in the case $\kappa<1/2$, $\gamma_4=2\gamma$. All conditions therefore hold when $\gamma\le c'\min\{\kappa,1-\kappa\}/\log K$ for a suitable $c'$ depending only on $a$, and $K\ge K_0$. The auditor may know the uniform score marginal, the cells and the padding: only the location and size of the risk heterogeneity inside $R$ are random. At $\kappa=1/2$ the padding is empty and the construction is the one of Theorem~\ref{thm:heterogeneity} with $B=K-1$ consecutive groups. \qed

\paragraph{Numerical checks.} The artifact test \texttt{test\_v61.py} verifies the trimming inequality of both cases on 20,000 random padded instances with $\kappa\in(.05,.95)$, the width bound~\eqref{eq:cp-width} for every outcome with $t\le10^4$ trials and $\zeta\in\{.025,.05,.25\}$, and Lemma~\ref{lem:profile} against constrained numerical optimization.

\paragraph{Consequences.} The ratio of the bounds in parts (i) and (ii) of Proposition~\ref{prop:bracket} is $O(d/(\kappa^2\alpha(1-\alpha)))$ up to $(\delta,\beta)$ terms. On the uniform sign family, the alternative prior draws of Theorem~\ref{thm:heterogeneity} with $r^*(\eta)\ge\alpha+\gamma$ lie in $\mathcal P_{\gamma/2}$ by the trimming argument of Theorem~\ref{cor:intervals} (case $\kappa=1/2$), which uses $d_b\le D\le1/16$; the whole class $\mathcal P_{\rm het}$ need not, since trimming a high-risk lower half can lower the risk by more. Hence Theorem~\ref{thm:heterogeneity}'s construction also shows that the upper end is attained up to $\log d$ there.

\subsection{Proof of Theorem~\ref{thm:heterogeneity}}
\label{app:heterogeneity}
Constants $c,C$ below depend only on $(a,\delta,\beta)$ unless stated otherwise.

\paragraph{Upper bounds.} Let $P_\eta\in\mathcal P_{\rm het}$. Every policy has coverage exactly $1/2$ and risk $r_\eta(A_\theta)\ge\alpha+\gamma$. Hoeffding's inequality and a union bound over the $2^B$ policies give, with probability at least $1-\eta_0$, $T_A\ge n/2-C\sqrt{n(B+\log(1/\eta_0))}$ and $E_A-\alpha T_A\ge n\gamma/2-C\sqrt{n(B+\log(1/\eta_0))}$ for every $A$. As in the upper-bound part of Appendix~\ref{app:iut-rate}, for $n\ge CB/\gamma^2$ every Clopper--Pearson risk lower bound exceeds $\alpha$, so the IUT rejects with probability at least $\beta$ uniformly over the class. For a single known $P\in\mathcal P_{\rm het}$ and a feasible $Q$ with witness $A$, the function $f=\1_A(L-\alpha)$ has oscillation one, $\E_Qf\le0$ and $\E_Pf=\{r_\eta(A)-\alpha\}/2\ge\gamma/2$. So every feasible law is $\gamma/2$-far after the $4B$-cell pushforward, and identity testing gives $N(P,\cH)=O(\sqrt B/\gamma^2)$.

\paragraph{Step 1: moment-matched magnitudes.} For $L\ge1$ let $E_L=\inf_g\sup_{u\in[0,1]}|\sqrt u-g(u)|$ over polynomials $g$ of degree at most $L$. Substituting $u=x^2$, $E_L$ equals the error of best approximation of $|x|$ on $[-1,1]$ by polynomials of degree $2L$, and Bernstein's theorem gives $2LE_L\to\beta_*\approx0.2802$ \citep{bernstein1914meilleure}. Since each $E_L>0$, $c_*=\inf_{L\ge1}LE_L>0$; numerically $LE_L\ge1/8$ for $L\le12$ (Appendix~\ref{app:heterogeneity-check}). By duality \citep[Lemma~1]{cai2011testing} there are probability measures $\pi_0,\pi_1$ on $[0,1]$ with $\int u^jd\pi_0=\int u^jd\pi_1$ for $j=1,\ldots,L$ and $\int\sqrt u\,d\pi_0-\int\sqrt u\,d\pi_1=2E_L$. For $D>0$ let $\nu_i$ be the law of $D\sqrt U$ with $U\sim\pi_i$. Then $\nu_0,\nu_1$ are supported on $[0,D]$, match $\E d^{2j}$ for $j\le L$, and $g\defeq\E_{\nu_0}d-\E_{\nu_1}d=2DE_L$.

\paragraph{Step 2: priors and their contracts.} Set $L=\lceil\log B\rceil$, $D=\gamma/E_L$ and $m=\alpha+\E_{\nu_0}d-g/4$. Under prior $i$, draw independently for each group a sign $s_b$ uniform on $\{\pm1\}$ and $d_b\sim\nu_i$, and set $\eta_{b,\pm}=m\pm s_bd_b$. Since $D\le\gamma L/c_*$, the condition $\gamma\le c_1/\log B$ with $c_1$ small gives $D\le a/2$; together with $m\in[\alpha-D/4,\alpha+D]$ and $\alpha\in[2a,1-2a]$ this keeps every $\eta_{b,h}$ in $[a,1-a]$. Because $r^*(\eta)=m-B^{-1}\sum_bd_b$, Hoeffding's inequality gives
\begin{align*}
 \Pr_0\{r^*(\eta)>\alpha\}+\Pr_1\{r^*(\eta)<\alpha+\gamma\}
 &\le2\exp\{-Bg^2/(8D^2)\}\\
 &=2\exp(-BE_L^2/2)\le2\exp\{-Bc_*^2/(2L^2)\},
\end{align*}
using $\E_0r^*=\alpha-g/4$, $\E_1r^*=\alpha+3g/4=\alpha+\gamma+g/4$ and $\gamma=g/2$. Under prior $0$ the policy choosing the lower half in each group has coverage $1/2$, so on the event $r^*\le\alpha$ the law $P_\eta$ is feasible; under prior $1$ and $r^*\ge\alpha+\gamma$ it lies in $\mathcal P_{\rm het}$.

\paragraph{Step 3: Poissonization.} If a test with $n$ samples has level $\delta$ and power $\beta$, so does one with any $n''\ge n$ samples; take $n''=\max\{n,n_1\}$ with $n_1=\lceil4\log(4/(\beta-\delta))\rceil$. Draw $N\sim\mathrm{Poi}(2n'')$ observations and apply the test to the first $n''$ if $N\ge n''$, otherwise accept. The level is unchanged and the power drops by at most $\Pr(N<n'')\le e^{-n''/4}\le(\beta-\delta)/4$. Under Poissonization with mean $n'=2n''$, the numbers of errors and correct observations in the $2B$ halves are independent Poisson variables with means $\lambda\eta_{b,h}$ and $\lambda(1-\eta_{b,h})$, $\lambda=n'/(2B)$; groups are independent under both priors.

\paragraph{Step 4: one group.} Let $F_0$ be the law of one group's four counts at $\eta\equiv m$, and let $\mu_i$ be its law under prior $i$. For Poisson likelihood ratios, $\E_{F_0}[(dP_x/dF_0)(dP_{x'}/dF_0)]=\exp\{\lambda xx'/m+\lambda xx'/(1-m)\}$ for error-rate shifts $x,x'$ of one half. The two halves have shifts $\pm sd$, so the group-level kernel is $\exp(\rho ss'dd')$ with $\rho=2\lambda/\{m(1-m)\}=n'/\{Bm(1-m)\}$. Averaging over independent signs gives $\cosh(\rho dd')$ and
\[
 \int\frac{(d\mu_1-d\mu_0)^2}{dF_0}
 =\sum_{j\ge0}\frac{\rho^{2j}}{(2j)!}\Big(\E_{\nu_1}d^{2j}-\E_{\nu_0}d^{2j}\Big)^2
 \le\sum_{j>L}\frac{(\rho D^2)^{2j}}{(2j)!}.
\]
With $\tau=\rho D^2$, $\sum_{j>L}\tau^{2j}/(2j)!\le e^\tau\{e\tau/(2L+2)\}^{2L+2}$. If $\tau\le(2L+2)/e^3$ this is at most $\exp\{-(2-e^{-3})(2L+2)\}\le e^{-3.9(L+1)}$. By Cauchy--Schwarz, $\TV(\mu_0,\mu_1)\le\frac12e^{-1.95(L+1)}$.

\paragraph{Step 5: all groups.} With $H^2(P,Q)=1-\int\sqrt{dP\,dQ}$ we have $H^2\le\TV$, $H^2(\mu_0^{\otimes B},\mu_1^{\otimes B})\le BH^2(\mu_0,\mu_1)$ and $\TV\le\sqrt{2H^2}$. Hence $\TV(\mu_0^{\otimes B},\mu_1^{\otimes B})\le\sqrt{B}\,e^{-0.975(L+1)}\le B^{-0.47}$ for $L=\lceil\log B\rceil$.

\paragraph{Step 6: conclusion.} Conditioning the priors on the events of Step~2 changes the mixtures by at most their probabilities. A test of level $\delta$ over $\cH$ and power $\beta$ over $\mathcal P_{\rm het}$ therefore satisfies
\[
 \beta-\delta-\tfrac{\beta-\delta}{4}\le B^{-0.47}+2\exp\{-Bc_*^2/(2L^2)\},
\]
which fails for $B\ge B_0$ whenever the condition $\tau\le(2L+2)/e^3$ holds. That condition reads $n'\le(2L+2)Bm(1-m)E_L^2/(e^3\gamma^2)$; with $m(1-m)\ge a/2$, $E_L\ge c_*/L$ and $L\le2\log B$ it holds whenever $n''\le c_2B/(\gamma^2\log B)$ for a suitable $c_2$. Since $n''=\max\{n,n_1\}$ and $n_1\le c_2B/(\gamma^2\log B)$ for $B\ge B_0$, no valid audit with $n\le c_2B/(\gamma^2\log B)$ samples has power $\beta$. \qed

\paragraph{Remarks.} The auditor in this lower bound knows the family, the cells and the exact uniform score marginal; only the location and size of the risk heterogeneity are random. The unknown quantity is a sum of $B$ nonsmooth terms $\min\{\eta_{b,+},\eta_{b,-}\}$, which is why the argument parallels minimax estimation of $\ell_1$-type functionals \citep{lepski1999estimation,cai2011testing,wu2016minimax}.

\subsection{Numerical checks for Theorem~\ref{thm:heterogeneity}}
\label{app:heterogeneity-check}
The script \texttt{heterogeneity\_checks.py} (i) computes $E_L$ by a linear program on a grid; (ii) solves the dual linear program for $\pi_0,\pi_1$ and confirms that the optimal moment-matched gap equals $2E_L$; and (iii) compares the $\chi^2$ series of Step~4 with direct summation over Poisson counts for one group (errors only, which has the same structure with $\rho=2\lambda/m$).
\begin{table}[ht]
\caption{Numerical checks behind Theorem~\ref{thm:heterogeneity}. Left: best-approximation error of $\sqrt u$ on $[0,1]$ (the monomial LP is reliable in this range; values increase towards Bernstein's constant $\beta_*\approx0.2802$). Middle: optimal moment-matched gap from the dual LP versus $2E_L$. Right: largest relative error between the closed-form $\chi^2$ series and direct Poisson summation over two parameter settings.}
\label{tab:hetcheck}
\centering\small
\begin{tabular}{rrr@{\hspace{2em}}rrr@{\hspace{2em}}rr}
\toprule
$L$ & $E_L$ & $2LE_L$ & $L$ & gap & $2E_L$ & $L$ & max rel.\ error\\
\midrule
1 & 0.12500 & 0.2500 & 2 & 0.13524 & 0.13524 & 2 & 1.4e-13\\
2 & 0.06762 & 0.2705 & 3 & 0.09186 & 0.09186 & 3 & 1.8e-11\\
3 & 0.04593 & 0.2756 & 4 & 0.06938 & 0.06938 & 4 & 1.6e-09\\
4 & 0.03469 & 0.2775 & & & & &\\
6 & 0.02325 & 0.2790 & & & & &\\
8 & 0.01747 & 0.2795 & & & & &\\
10 & 0.01399 & 0.2797 & & & & &\\
\bottomrule
\end{tabular}
\end{table}

\subsection{Protocol for the envelope comparison}
\label{app:het-envelope}
\emph{Setting.} Uniform sign family with $B\in\{64,256,1024,4096,16384,65536,262144\}$, uniform score marginal, $\kappa=1/2$, $\alpha=.2$ and Poissonized sample size $n$ (independent Poisson counts of errors and correct observations in each half).

\emph{Heterogeneity envelope.} The LP of Appendix~\ref{app:heterogeneity-check} with $L=2$ gives $\nu_0$ with atoms $\{0.0569,0.2\}$ and weights $\{0.904,0.096\}$, and $\nu_1$ with atoms $\{0,0.1554\}$ and weights $\{0.719,0.281\}$ ($D=.2$). Then $g=2DE_2$, $\gamma=g/2=0.01352$ and $m=\alpha+\E_{\nu_0}d-g/4=0.2639$. The statistic is the exact log-likelihood ratio between the two product mixtures, obtained by summing over atoms and signs in each group. For a valid audit $\phi$, $\E_{\rm mix_0}\phi\le\delta+\Pr_0(G_0^c)=\delta'$ and $\E_{{\rm mix}_1\mid G_1}\phi\le{\rm NP}_{\delta'}/\Pr_1(G_1)$. We estimate the conditioning probabilities from $2\times10^5$ multinomial draws, calibrate the NP threshold at level $\delta'$ on independent draws from ${\rm mix}_0$, and evaluate power on independent draws from ${\rm mix}_1$.

\emph{Known-reference envelope.} The reference has constant risk $p=\alpha+\gamma$. The null mixture averages the hiding components of Equation~\eqref{eq:qtheta} with the same $\gamma$, so every component is feasible and no conditioning is needed. The NP threshold is calibrated at level $\delta=.05$.

\begin{figure}[ht]
 \centering\includegraphics[width=\linewidth]{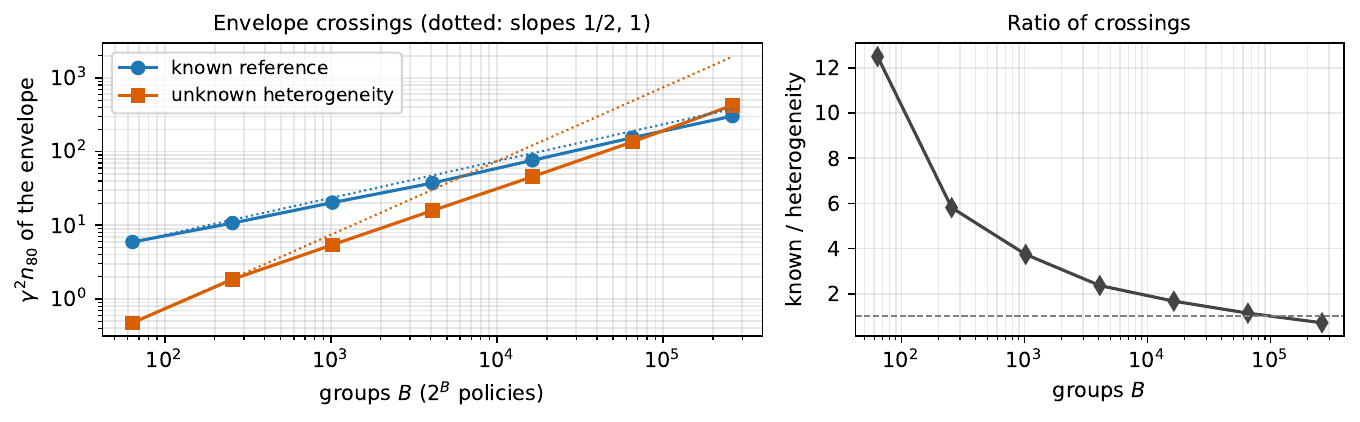}
 \caption{\textbf{Unknown heterogeneity changes the exponent and eventually dominates.} Left: 80\%-power crossings of the two envelopes, scaled by $\gamma^2$. The known-reference crossing grows with fitted exponent $0.48$ (theory: $1/2$); the heterogeneity crossing grows with exponent $0.80$, approaching $1-1/(2L+2)=0.83$ for fixed $L=2$ (the theorem lets $L\asymp\log B$). Right: their ratio falls from 12.5 at $B=64$ to 0.71 at $B=262{,}144$. Because the constant-risk reference belongs to $\mathcal P_{\rm het}$, the larger crossing is the operative lower bound on $N(\mathcal P_{\rm het},\cH)$. Envelopes bound valid audits; they are not audits.}
 \label{fig:heterogeneity}
\end{figure}

\emph{Replicates and crossings.} We use 2,000 replicates for calibration and 2,000 for power per grid point when $B\le4096$, and 1,000 of each when $B\ge16384$, with sample-size grids around the crossing. The $B=262{,}144$ runs used a second machine with different NumPy/SciPy versions; rerunning $B=256$ there changed the crossings by at most 1.2\%, within Monte Carlo error. The $n_{80}$ values are log-linear interpolations of Monte Carlo power estimates; the two envelopes use independent seeds. The script \texttt{heterogeneity\_envelope.py} produces all runs and \texttt{merge\_het\_envelope.py} merges them and draws Figure~\ref{fig:heterogeneity}.

\begin{table}[ht]
\caption{Envelope crossings behind Figure~\ref{fig:heterogeneity}. Because the constant-risk reference belongs to $\mathcal P_{\rm het}$, the larger of the two crossings is the operative lower bound on $N(\mathcal P_{\rm het},\cH)$. The last two columns report how often a prior draw leaves its hypothesis; this probability is charged in the heterogeneity envelope.}
\label{tab:hetenv}
\centering\small
\begin{tabular}{rrrrrr}
\toprule
$B$ & known $n_{80}$ & heterogeneity $n_{80}$ & ratio & $\Pr_0(G_0^c)$ & $\Pr_1(G_1^c)$\\
\midrule
64 & 32,484 & 2,600 & 12.49 & 0.123 & 0.238\\
256 & 58,713 & 10,108 & 5.81 & 0.003 & 0.055\\
1,024 & 111,210 & 29,743 & 3.74 & 0.000 & 0.001\\
4,096 & 205,041 & 86,773 & 2.36 & 0.000 & 0.000\\
16,384 & 417,386 & 249,855 & 1.67 & 0.000 & 0.000\\
65,536 & 846,246 & 742,665 & 1.14 & 0.000 & 0.000\\
262,144 & 1,660,543 & 2,327,210 & 0.71 & 0.000 & 0.000\\
\bottomrule
\end{tabular}
\end{table}

\section{Estimation complexity, IUT inefficiency, and reference adaptation}
\label{app:efficiency}

\subsection{Exact distribution-dependent Rademacher complexity}
\label{app:rademacher-exact}
The profiles of Theorem~\ref{thm:separation} are distinguished by the population Rademacher complexity, $\mathfrak R_n\asymp\sum_b\min\{w_b,\sqrt{w_b/n}\}$, of order $\sqrt{k/n}$ on a plateau; but it does so at the \emph{estimation} scale $k/\gamma^2$, whereas the optimal audit works at the \emph{testing} scale $\sqrt k/\gamma^2$, as in instance-optimal identity testing \citep{valiant2017identity,balakrishnan2019local}.
For $\cF_w=\{\1_{A_\theta}:\theta\in\{-1,1\}^B\}$ define
$\mathfrak R_n(\cF_w)=\E_{S,\sigma}\sup_\theta n^{-1}\sum_{i=1}^n\sigma_i\1_{A_\theta}(S_i)$, with independent Rademacher signs $\sigma_i$. Let $h(s)$ denote the $+$/$-$ half within its group. Since $\1_{A_\theta}(s)=(1+\theta_bh(s))/2$ on $G_b$,
\begin{equation}
 \mathfrak R_n(\cF_w)=\frac1{2n}\sum_{b=1}^B\E|Z_b|,
 \qquad Z_b=\sum_{i=1}^n\sigma_i h(S_i)\1\{S_i\in G_b\}.
 \label{eq:rad-exact}
\end{equation}
The constant term has expectation zero. A summand in $Z_b$ is $\pm1$ with probability $w_b/2$ each, and zero otherwise. Thus, putting $v=nw_b$,
\[
 \E Z_b^2=v,\qquad \E Z_b^4\le v+3v^2.
\]
H\"older's inequality gives $\E|Z_b|\ge(\E Z_b^2)^{3/2}/(\E Z_b^4)^{1/2}\ge v/\sqrt{1+3v}$, while $\E|Z_b|\le\min\{v,\sqrt v\}$. Consequently
\begin{equation}
 \frac14\sum_b\min\{w_b,\sqrt{w_b/n}\}
 \le\mathfrak R_n(\cF_w)
 \le\frac12\sum_b\min\{w_b,\sqrt{w_b/n}\}.
 \label{eq:rad-bounds}
\end{equation}
If $nw_{\min}\ge1$, this is $\Theta(\sum_b\sqrt{w_b}/\sqrt n)$. For a plateau and $n\ge k$, the heavy groups contribute $\Theta(\sqrt{k/n})$ and the residual contributes at most $\varepsilon_B/2$. Therefore, around $n\asymp k/\gamma^2$ with $\varepsilon_B\le\gamma/16$, the complexity is of order $\gamma$. Distribution-dependent complexity does distinguish our profiles; it is not one of the quantities matched in Theorem~\ref{thm:separation}.

For indicator policies the squared $L^2(P_0^S)$ distance is
\[
 \|\1_{A_\theta}-\1_{A_{\theta'}}\|_2^2
 =\sum_b w_b\1\{\theta_b\ne\theta_b'\}.
\]
At any fixed sufficiently small positive resolution, a plateau has logarithmic covering and packing numbers $\Theta(k)$: ignoring residual coordinates gives a $2^k$ cover once the squared resolution exceeds $\varepsilon_B$, and a constant-relative-distance binary code on the heavy coordinates gives a packing of size $\exp(c k)$. Thus the usual metric entropy sees the heavy coordinates. The point is a difference between estimation and testing rates, not failure of every distribution-dependent complexity measure.

\subsection{Proof of the IUT rate separation}
\label{app:iut-rate}
\begin{table}[ht]
\caption{Two statistical scales on a $k$-plateau (constants and the residual suppressed).}
\label{tab:mechanisms}
\centering\small
\begin{tabular}{lccc}
\toprule
Evidence & Signal & Reference fluctuation & Sample scale\\
\midrule
Best (minimum) empirical policy risk & $\gamma$ & $\sqrt{k/n}$ & $k/\gamma^2$\\
Quadratic family evidence & $\gamma^2$ & $\sqrt{k}/n$ & $\sqrt{k}/\gamma^2$\\
\bottomrule
\end{tabular}
\end{table}
Part (ii) of Theorem~\ref{thm:iut-rate} is proved for the exact Clopper--Pearson IUT with any fixed positive allocation $\zeta_c+\zeta_r=\delta$. Part (i) is proved for every level-$\delta$ EFC audit: on the main range the argument below only uses that the audit does not reject while the constructed policy $\widehat A$ is empirically feasible, and level $\delta$ is used only in the bounded ranges at the end. Constants below may depend on $p,\delta,\beta$ but not on $B,k,\gamma$.

\paragraph{Upper bound.}
Let $\mathcal R$ be the residual groups, of reference mass $\varepsilon\le\gamma/16$, with $\varepsilon=0$ for $k=B$. For a heavy-coordinate sign pattern $\theta_H$, define
\[
 T_H=\sum_i\1\{S_i\in A_{\theta_H}\setminus\mathcal R\},\qquad
 W_H=\sum_i(L_i-\alpha)\1\{S_i\in A_{\theta_H}\setminus\mathcal R\}.
\]
Their expectations are $n(1-\varepsilon)/2$ and $n(1-\varepsilon)\gamma/2$. A union bound over only $2^k$ heavy patterns and Hoeffding's inequality give, with probability at least $1-\eta$, simultaneously
\[
 T_H\ge n(1-\varepsilon)/2-C\sqrt{n(k+\log(1/\eta))},\qquad
 W_H\ge n(1-\varepsilon)\gamma/2-C\sqrt{n(k+\log(1/\eta))}.
\]
Also $N_{\mathcal R}\le n\varepsilon+C\sqrt{n\log(1/\eta)}$ on an event of the same order of probability. Any full policy has $T_A\ge T_H$ and
$E_A-\alpha T_A\ge W_H-\alpha N_{\mathcal R}$. Since $\varepsilon\le\gamma/16$, at $n\ge C'(k+\log(1/\eta))/\gamma^2$ these inequalities imply $T_A\ge n/4$ and $E_A/T_A\ge\alpha+c\gamma$ for \emph{every} policy. A one-sided Clopper--Pearson lower bound is at least the empirical proportion minus $\sqrt{\log(1/\zeta_r)/(2T_A)}$ whenever this difference is positive: this follows by inverting the Hoeffding bound for a binomial upper tail. Increasing $C'$ makes every risk lower bound exceed $\alpha$. Taking $\eta<1-\beta$ proves $N_{\rm IUT}\le C k/\gamma^2$. This argument does not pay for the number of residual sign patterns.

\paragraph{A constructive empirical witness for the lower bound.}
Condition on the total number of errors $M$, and write $m=n-M$ for the number of correct observations. In each heavy group choose the half with \emph{more correct observations}; fix the $+$ half in every residual group. This defines a policy $\widehat A$ using correct observations alone. Conditional on these observations and $M$, the error observations have independent uniform scores, and any selected sign policy has score probability $1/2$. Hence
\[
 E_{\widehat A}=M/2+Z,\qquad\E[Z\mid M,\text{correct data}]=0,
 \quad\E[Z^2\mid M,\text{correct data}]=M/4.
\]
If $C_{b,+},C_{b,-}$ are the heavy correct counts and $C_{\mathcal R,+}$ is the correct count in the fixed residual halves, then
\[
 C_{\widehat A}=m/2+G,\qquad
 G=\tfrac12\sum_{b\le k}|C_{b,+}-C_{b,-}|
   +C_{\mathcal R,+}-C_{\mathcal R}/2.
\]
The last term has conditional mean zero. Each heavy difference has variance $v=m(1-\varepsilon)/k$ and fourth moment at most $v+3v^2$, so
\[
 \E[G\mid M]\ge\frac{m(1-\varepsilon)}{2\sqrt{1+3m(1-\varepsilon)/k}}
 \ge c_0\min\{m,\sqrt{mk}\}.
\]
The second inequality holds uniformly for $\varepsilon\le1/2$. Replacing one correct observation changes the maximum selected correct count by at most one. Efron--Stein therefore gives $\operatorname{Var}(G\mid M)\le m/2$ (the weaker bound $2m$ also suffices). Chebyshev, binomial concentration for $M$, and the conditional variance of $Z$ imply, except with probability at most $C_0(1/n+1/k)$,
\[
 G\ge c_1\min\{n,\sqrt{nk}\},\quad
 |M-np|+2|Z|\le c_2\min\{n,\sqrt{nk}\},
\]
where $c_2>0$ can be chosen sufficiently small relative to $p c_1$ by changing $C_0$. Here the first bound also uses $m\ge(1-p)n/2$; its failure has probability $O(1/n)$.

For $\gamma\le\gamma_0(p)$ sufficiently small and $n\le c k/\gamma^2$ with $c$ sufficiently small,
$n\gamma\le c_3\min\{n,\sqrt{nk}\}$. Since $\alpha=p-\gamma\ge p/2$, the same event yields
\[
 T_{\widehat A}=n/2+G+Z\ge n/2,
 \qquad
 E_{\widehat A}-\alpha T_{\widehat A}
   =\tfrac12(M-\alpha n)+(1-\alpha)Z-\alpha G\le0.
\]
Thus $\widehat A$ is empirically feasible, so no EFC audit rejects; for the exact IUT this is Lemma~\ref{lem:cp-median}. For large enough $n,k$, its rejection probability is below $\beta$ throughout $n\le c k/\gamma^2$.

For completeness, the bounded-$k$ cases follow from the information-theoretic lower bound $N\ge c'\sqrt{k}/\gamma^2$, which is at least a constant times $k/\gamma^2$ on any fixed bounded range. For bounded $n$ and sufficiently large $k$, the mixture ceiling in Proposition~\ref{prop:weight-profile} is below $\beta$, uniformly over bounded $\gamma$: $\|w\|_2^2=O(1/k)$ makes its exponent $O(n^2\gamma^4/k)$. Combining these ranges proves the claimed uniform lower rate. This proof controls the coverage screen as well as risk; minimizing error counts alone would not suffice.

\subsection{Proof of Proposition~\ref{prop:efc-slack}}
\label{app:efc-slack}
Every rule of the sign family has reference coverage exactly $\kappa$. The estimation cost does not depend on this boundary case.
\begin{proposition}[Estimation cost with coverage slack]
\label{prop:efc-slack}
Split each group of a $k$-plateau into $q$ equal parts and let each policy accept $m$ parts per group (Proposition~\ref{prop:rational-kappa}). With reference coverage $m/q$, a fixed floor $\kappa<m/q$ and residual mass at most $m\gamma/(16q)$, there are $\gamma_0,c_1,C_1>0$ depending only on $(p,q,m,\kappa,\delta,\beta)$ such that, for $\gamma\le\gamma_0$, every level-$\delta$ EFC audit has power below $\beta$ at $P_0$ for all $n\le c_1k/\gamma^2$, while the exact IUT has power at least $\beta$ for $n\ge C_1k/\gamma^2$.
\end{proposition}
The witness chooses, in each heavy group, the $m$ parts with the most correct observations; with slack the coverage screen is passed automatically, and the gain in correct counts again scales as $\min\{n,\sqrt{nk}\}$.
Use the construction of Proposition~\ref{prop:rational-kappa}: $P_0$ has uniform scores and independent $L\sim\Bern(p)$, each group $G_\ell$ is split into $q$ equal parts, a policy selects $m$ parts per group, $c=m/q>\kappa$, $\alpha=p-\gamma$, and the residual groups have total mass $\varepsilon\le c\gamma/16$.

\paragraph{Upper bound.} Repeat the upper-bound argument of Appendix~\ref{app:iut-rate} with the $\binom qm^k$ heavy patterns in place of $2^k$: the union bound costs $k\log\binom qm$, every policy has $T_A\ge nc/2$ and $E_A-\alpha T_A\ge nc\gamma/4$ once $n\ge Ck/\gamma^2$, and the Clopper--Pearson inversion makes every risk bound exceed $\alpha$.

\paragraph{A top-$m$ inequality.} If $x_1,\ldots,x_q$ sum to zero, the sum of the $m$ largest is at least $\frac{\min\{m,q-m\}}{2q}\sum_a|x_a|$. Indeed, let $P=\sum_a(x_a)_+=\frac12\sum_a|x_a|$ and let $r$ be the number of positive entries. If $m\le r$, the $m$ largest positive entries sum to at least $(m/r)P\ge(m/q)P$. Otherwise the top $m$ contain all positive entries and the $m-r$ negative entries of smallest magnitude, whose total magnitude is at most $\frac{m-r}{q-r}P$, so the sum is at least $\frac{q-m}{q-r}P\ge\frac{q-m}{q}P$.

\paragraph{Witness.} Condition on the error count $M$ and on all correct observations; let $m_0=n-M$. In each heavy group choose the $m$ parts with the most correct observations, and fix $m$ parts in every residual group. Conditionally, the error observations have independent uniform scores and every policy has score probability $c$, so the accepted errors are $E=cM+Z$ with $\E Z=0$ and $\operatorname{Var}Z=Mc(1-c)$. The accepted correct count is $cm_0+G$, where
\[
 G=\sum_{b\le k}\Big(\sum_{a\in\widehat C_b}N_{b,a}-cN_b\Big)+R,
\]
$N_{b,a}$ is the correct count in part $a$ of group $b$, $N_b=\sum_aN_{b,a}$, and $R$ is the centred residual contribution. By the top-$m$ inequality applied to $x_a=N_{b,a}-N_b/q$, each heavy term is at least $\frac{\min\{m,q-m\}}{2q}\sum_a|N_{b,a}-N_b/q|$. Each $N_{b,a}-N_b/q$ is a sum of $m_0$ independent centred terms bounded by one, with variance $v=m_0w_b(q-1)/q^2$ and fourth moment at most $v+3v^2$. H\"older's inequality gives $\E|N_{b,a}-N_b/q|\ge v/\sqrt{1+3v}$, and therefore $\E[G\mid M]\ge c_0\min\{m_0,\sqrt{m_0k}\}$ with $c_0>0$ depending on $(q,m)$. Moving one correct observation changes the top-$m$ sums by at most one and $c\sum_bN_b$ by at most $2c$, so Efron--Stein gives $\operatorname{Var}(G\mid M)\le9m_0/2$.

\paragraph{Empirical feasibility.} As in Appendix~\ref{app:iut-rate}, Chebyshev's inequality and binomial concentration give, except with probability $C_0(1/n+1/k)$,
\[
 G\ge c_1'\min\{n,\sqrt{nk}\},\qquad |M-np|+2|Z|\le c_2'\min\{n,\sqrt{nk}\},
\]
with $c_2'$ small relative to $\alpha c_1'$ and to $c-\kappa$. The accepted count is $T=cn+G+Z\ge cn-|Z|\ge\kappa n$, since $|Z|\le(c-\kappa)n$ on this event. Moreover
\[
 E-\alpha T=c(M-pn)+c\gamma n+(1-\alpha)Z-\alpha G\le0
\]
whenever $c\gamma n\le c_3\min\{n,\sqrt{nk}\}$, which holds for $n\le c_1k/\gamma^2$ and $\gamma\le\gamma_0$. The selected policy is then empirically feasible, and no EFC audit rejects. The bounded-$k$ and bounded-$n$ ranges are handled exactly as in Appendix~\ref{app:iut-rate}, using the lower bound $\Omega(\sqrt k/\gamma^2)$ and the product-mixture ceiling of Proposition~\ref{prop:rational-kappa}. \qed

\subsection{Explicit constants when only the loss rate is unknown}
\label{app:adaptive}
Theorem~\ref{thm:constant} relies on a generic independence tester. When the score marginal is known and uniform, a pilot-and-collision audit gives explicit constants.
\begin{proposition}[Adaptation to an unknown constant risk]
\label{thm:adaptive}
On the uniform sign family $B=k$, let $P_p$ have uniform scores over the $2k$ halves and independent $L\sim\Bern(p)$, and let $\mathcal P_{a,\alpha,\gamma}=\{P_p:p\in[a,1-a],\ p\ge\alpha+\gamma\}$. Fix $0<a<1/3$, $\alpha\in[a,1-2a]$ and $0<\delta<\beta<1$. For sufficiently small $0<\gamma\le a$, the audit below has level $\delta$ over the unrestricted null, does not take $p$ as input, and has power $\beta$ over $\mathcal P_{a,\alpha,\gamma}$ with $O_{a,\delta,\beta}(\sqrt k/\gamma^2)$ samples, which is optimal.
\end{proposition}
We prove Proposition~\ref{thm:adaptive}. The score-space partition is known, but the loss probability $p$ is not an input to the audit. For power, the reference class has uniform scores on the $2k$ halves and independent $L\sim\Bern(p)$ with $p\in[a,1-a]$ and $p\ge\alpha+\gamma$. Type-I validity continues to cover \emph{every} law in the unrestricted existential null, without uniform-score or score--loss independence assumptions.

Set $\epsilon=\min\{\delta,(1-\beta)/3\}$. Use $m$ pilot observations to estimate $p$ by their mean loss projected onto $[a,1-a]$, denoted $\widehat p$. If $\widehat p<\alpha+\gamma/2$, do not reject. Otherwise apply the collision test from Theorem~\ref{thm:collision} to an independent batch of $n$ observations with the candidate reference $P_{\widehat p}$. For each candidate,
\[
 \Delta_{\widehat p}=\frac{(\widehat p-\alpha)^2}{2\widehat p(1-\widehat p)}
 \ge\gamma^2/2,\qquad p_{\min}\ge a/(2k),\qquad d=4k.
\]
All categorical cells are reference-positive. A common sufficient size for the second batch is
\begin{equation}
 n\ge\max\!\left\{2,\left\lceil\frac{32}{\epsilon}
 \max\left\{\frac{4\sqrt{k}}{\gamma^2},
 \frac2{\gamma^2}+\frac{2\sqrt{k}}{\gamma\sqrt a}\right\}\right\rceil\right\}.
 \label{eq:adaptive-n}
\end{equation}
Indeed, $\sqrt{d-1}/\Delta\le4\sqrt{k}/\gamma^2$ and
$u_0/\Delta=1/\Delta+1/\sqrt{\Delta p_{\min}}$ is bounded by the second expression. Conditional on \emph{any} pilot, the collision test has level at most $\epsilon$ for every feasible law; a failed pilot screen never rejects. Therefore the entire procedure has level at most $\delta$ without assuming the power model.

Under $P_p$, projection cannot increase squared estimation error, so $\E(\widehat p-p)^2\le1/(4m)$. For candidate Bernoulli laws,
\[
 \KL(P_p\Vert P_{\widehat p})=\mathrm{kl}(p\Vert\widehat p)
 \le\frac{(p-\widehat p)^2}{a(1-a)}.
\]
Pinsker and Jensen yield
\[
 \E\TV(P_p^n,P_{\widehat p}^n)
 \le\sqrt{\frac{n}{8ma(1-a)}}.
\]
Also $P_p(\widehat p<\alpha+\gamma/2)\le e^{-m\gamma^2/2}$. Taking
\begin{equation}
 m\ge\max\left\{\frac{n}{8a(1-a)\epsilon^2},
                 \frac{2\log(1/\epsilon)}{\gamma^2}\right\}
 \label{eq:adaptive-m}
\end{equation}
bounds both terms by $\epsilon$. On a passed screen the test has power at least $1-\epsilon$ under its candidate $P_{\widehat p}$; transferring to $P_p$ and averaging gives power at least $1-3\epsilon\ge\beta$. The total $m+n$ is $O_{a,\delta,\beta}(\sqrt{k}/\gamma^2)$. The lower bound follows by fixing $p=\alpha+\gamma$ in the reference class and applying the uniform-profile mixture bound, with constants uniform over $[a,1-a]$.

The statistic can be computed from cell counts in $O(n+k)$ time using $U_n=[\sum_j N_j(N_j-1)/p_j]/[n(n-1)]-1$. Its conservative constants are not claimed to be practically competitive. The result establishes the rate and refutes a universal interpretation of $k/\gamma^2$ as the unavoidable cost of not knowing $p$. It does not solve adaptation to an arbitrary unknown score marginal or conditional risk function.

\section{Proof of the matched same-cardinality complexity spectrum}
\label{app:separation}

\subsection{Exact singleton sufficiency and audit equivalence}
Every component has exactly
\begin{align}
 \TV(P_0,Q_\theta)&=\gamma, \label{eq:component-tv}\\
 \KL(P_0\Vert Q_\theta)&=\tfrac12\,\mathrm{kl}(p\Vert p-\gamma)+\tfrac12\,\mathrm{kl}(p\Vert p+\gamma),\\
 \KL(Q_\theta\Vert P_0)&=\tfrac12\,\mathrm{kl}(p-\gamma\Vert p)+\tfrac12\,\mathrm{kl}(p+\gamma\Vert p).
 \label{eq:matchedinfo}
\end{align}

For reference coverage $c$, accepted risk $r$, $0<\kappa\le c$, and $\alpha=r-\gamma$ with $0<\gamma<r$, the exact projections are as follows.
Moreover, putting $d=\mathrm{kl}(r-\gamma\Vert r)$ and
$t^*=\max\{\kappa,ce^{-d}/(1-c+ce^{-d})\}$,
\begin{align}
 \inf_{Q\in\cH(\{A\})}\TV(P_0,Q)&=c\gamma,\label{eq:singleton-tv}\\
 \inf_{Q\in\cH(\{A\})}\KL(P_0\Vert Q)&=c\,\mathrm{kl}(r\Vert r-\gamma),\\
 \inf_{Q\in\cH(\{A\})}\KL(Q\Vert P_0)&=\mathrm{kl}(t^*\Vert c)+t^*d.
 \label{eq:singleton-projections}
\end{align}
The four-cell sufficient-statistic reduction is proved below. The constant-risk case used below has $r=s=p$ and gives these formulas as a special case. Note that the canonical hiding component $Q_\theta$ in Equation~\eqref{eq:qtheta} is not the closest feasible singleton law: for $c=1/2$, Equation~\eqref{eq:component-tv} gives $\TV(P_0,Q_\theta)=\gamma$, whereas the singleton projection in Equation~\eqref{eq:singleton-tv} is $\gamma/2$.

\label{app:singleton-proof}
\begin{proposition}[Singleton experiments are matched]
\label{prop:singleton}
For a policy $A$ whose four $(\1\{S\in A\},L)$ reference cells are positive, the unrestricted singleton testing problem at every sample size is determined by this four-cell law. Hence every member of every sign family has the same singleton minimax complexity.
\end{proposition}
We prove Proposition~\ref{prop:singleton} without assuming that $L$ is independent of the score. Let $M=\1\{S\in A\}$ and $T(S,L)=(M,L)$. The singleton contract depends on a law only through the four probabilities of $T$. Fix once and for all a Markov kernel $K(ds\mid m,l)$ supported on $A$ when $m=1$ and on $A^c$ when $m=0$, taking $K(B\mid m,l)=P_0(S\in B,M=m,L=l)/P_0(M=m,L=l)$, the normalized restriction of $P_0$ to that finite event. Proposition~\ref{prop:singleton} assumes all four cells are positive, so no general disintegration theorem is needed.

Every feasible four-cell law $q$ has a canonical lift $Q_q$: draw $T\sim q$ and then $S\sim K(\cdot\mid T)$. Conversely, $T_\#Q$ is feasible whenever $Q\in\cH(\{A\})$. Hence data processing gives one direction for every divergence and every test.

The testing reduction is exact for every sample size. Given any original-sample test $\phi_n$, define
\[
 \psi_n(t^n)=\int \phi_n(s^n,l^n)\prod_{i=1}^n K(ds_i\mid t_i).
\]
For every canonical lift $Q_q$, $\E_{Q_q^n}\phi_n=\E_{q^n}\psi_n$, and the same identity holds under $P_0^n$. Therefore every valid original test induces a four-cell test with the same power, while the converse follows by composition with $T^n$. The unrestricted singleton minimax problem thus depends on $(P_0,A)$ only through $T_\#P_0$. In particular, two policies with identical reference four-cell laws have identical singleton audit complexity even when $\eta(s)=P_0(L=1\mid S=s)$ is nonconstant.

For the explicit projections, write
\[
 c=P_0(M=1),\quad r=P_0(L=1\mid M=1),\quad s=P_0(L=1\mid M=0),\quad \alpha=r-\gamma.
\]
For TV, $f(M,L)=M(L-\alpha)$ has oscillation one. Under $P_0$, $\E f=c\gamma$, whereas every feasible $Q$ has $\E_Qf\le0$, so $\TV(P_0,Q)\ge c\gamma$. Equality is attained in the four-cell problem by preserving $P(M=1)=c$ and the outside conditional risk $s$ while changing the accepted risk from $r$ to $r-\gamma$; the canonical lift preserves TV exactly.

For the KL projections, write $t=Q(M=1)$, $u=Q(L=1\mid M=1)$, and $v=Q(L=1\mid M=0)$. Data processing and the Bernoulli chain rule give
\begin{align*}
 \KL(P_0\Vert Q)
 &\ge \mathrm{kl}(c\Vert t)+c\,\mathrm{kl}(r\Vert u)+(1-c)\,\mathrm{kl}(s\Vert v),\\
 \KL(Q\Vert P_0)
 &\ge \mathrm{kl}(t\Vert c)+t\,\mathrm{kl}(u\Vert r)+(1-t)\,\mathrm{kl}(v\Vert s).
\end{align*}
The feasible constraints are $t\ge\kappa$ and $u\le r-\gamma$. The forward direction is minimized at $(t,u,v)=(c,r-\gamma,s)$, giving $c\,\mathrm{kl}(r\Vert r-\gamma)$.

For the reverse direction first set $(u,v)=(r-\gamma,s)$, optimal for every fixed $t$. With $d=\mathrm{kl}(r-\gamma\Vert r)$ the remaining objective is
\[
 h(t)=\mathrm{kl}(t\Vert c)+td,\qquad t\ge\kappa.
\]
Its unconstrained minimizer is $t_0=ce^{-d}/(1-c+ce^{-d})$, so the constrained optimizer is $t^*=\max\{\kappa,t_0\}$ and the minimum is $\mathrm{kl}(t^*\Vert c)+t^*d$. Canonical lifts attain the data-processing bounds because the likelihood ratio depends only on $T$. This proves Equation~\eqref{eq:singleton-projections}. The score-independent special case is obtained by $r=s=p$.

\subsection{Common component information}
Every policy $A_\theta$ accepts one half of each group, so $P_0(S\in A_\theta)=1/2$ and under \eqref{eq:qtheta},
$Q_\theta(L=1\mid S\in A_\theta)=p-\gamma=\alpha$. Half of the score mass shifts the Bernoulli parameter down by $\gamma$ and half shifts it up by $\gamma$, independently of group weights. Summing the two-point channel distances gives \eqref{eq:matchedinfo} and $\TV(P_0,Q_\theta)=\gamma$.

For independent uniform signs $\theta,\theta'$, the standardized Bernoulli calculation gives
\[
 \Xi(\theta,\theta')
 =\frac{\gamma^2}{p(1-p)}\sum_{\ell=1}^B w_\ell\theta_\ell\theta'_\ell
 =\lambda\sum_{\ell=1}^B w_\ell\varepsilon_\ell,
\]
where $\varepsilon_\ell$ are i.i.d. Rademacher variables. The nonzero covariance eigenvalues are proportional to $w_\ell$, giving $d_{\mathrm{eff}}=(\sum_\ell w_\ell^2)^{-1}$.

\begin{proposition}[Combinatorial invariance of the matched sign families]
\label{prop:combinatorial}
For any two positive profiles $w,w'\in\Delta_{B-1}$, the corresponding sign families have the same nonempty membership-signature incidence structure up to relabeling of the $2B$ half-group atoms. In particular, both have cardinality $2^B$, VC dimension exactly $B$, and growth function $\Pi_{\cA}(m)=2^{\min\{m,B\}}$. Thus changing the profile changes only the $P_0$-masses assigned to a common combinatorial set system, not its unweighted class complexity.
\end{proposition}

\begin{remark}[Pairwise matching inside the plateau spectrum]
\label{rem:pairwise-barrier}
If $\theta^{(\ell)}$ is obtained from $\theta$ by flipping only coordinate $\ell$, then $A_\theta\triangle A_{\theta^{(\ell)}}=G_\ell$ and hence $P_0(A_\theta\triangle A_{\theta^{(\ell)}})=w_\ell$. Thus adjacent pairwise membership laws reveal the plateau weights. A pairwise-matched analogue of Theorem~\ref{thm:separation} therefore requires a qualitatively different construction; the Supplementary Note gives such a construction for the weaker but already nontrivial question of one-observation family hiding.
\end{remark}

\subsection{Proof of Proposition~\ref{prop:combinatorial}: fixed incidence and VC complexity}
For every positive profile, quotient the score space by policy-membership signature. The nonempty signature atoms are exactly the $2B$ half-groups $G_\ell^+$ and $G_\ell^-$: policy $A_\theta$ contains precisely $G_\ell^{\theta_\ell}$ from group $\ell$. This $2B$-atom incidence matrix depends on $B$ and the sign labels but not on the mass profile $w$. Hence any two profiles have isomorphic quotient set systems; only the reference masses of the atoms differ.

For the VC dimension, choose one point from $G_\ell^+$ for each of $m\le B$ distinct groups. The corresponding $m$ membership bits can be set arbitrarily by the $m$ sign coordinates, so every such $m$ is shattered and $\mathrm{VCdim}(\cA)\ge B$. Conversely, among any $B+1$ points two lie in the same group. If they lie in the same half, every policy gives them identical membership; if they lie in opposite halves, every policy gives complementary membership. In either case the pair cannot realize all four binary labelings, so no $B+1$ points are shattered. Thus $\mathrm{VCdim}(\cA)=B$.

Finally, for $m\le B$ the shattered construction realizes $2^m$ traces. For $m>B$, the family contains only $2^B$ policies, while an $m$-point set containing one point from each $G_\ell^+$ plus additional distinct points in any already represented half-group still distinguishes all $2^B$ sign vectors. Therefore $\Pi_{\cA}(m)=2^{\min\{m,B\}}$.

\subsection{Proof of Proposition~\ref{prop:weight-profile}}
Let $w$ be any positive group-mass vector and let $\varepsilon_\ell=\theta_\ell\theta'_\ell$ for independent uniform sign vectors. Then
\[
 \Xi(\theta,\theta')=\lambda\sum_{\ell=1}^B w_\ell\varepsilon_\ell,
 \qquad \lambda=\frac{\gamma^2}{p(1-p)}.
\]
By Lemma~\ref{thm:overlap}, $1+\chi^2(R_n\Vert P_0^n)=\E(1+\Xi)^n$. Since $1+x\le e^x$ for $x>-1$ and the $\varepsilon_\ell$ are independent Rademacher variables,
\begin{align*}
 1+\chi^2(R_n\Vert P_0^n)
 &\le \E\exp\!\left(n\lambda\sum_\ell w_\ell\varepsilon_\ell\right)\\
 &=\prod_{\ell=1}^B\cosh(n\lambda w_\ell)
 \le \exp\!\left(\frac{n^2\lambda^2}{2}\sum_{\ell=1}^B w_\ell^2\right),
\end{align*}
where $\cosh t\le e^{t^2/2}$. Corollary~\ref{cor:powerceiling} now gives the ceiling of Proposition~\ref{prop:weight-profile}. Choosing the constant in $n\le c/(\lambda\|w\|_2)$ so that the right-hand side stays below $\beta$ yields $N_w=\Omega(\sqrt{d_{\rm eff}}/\gamma^2)$.

Two uniform bounds delimit the entire sign-family class. First, any one canonical component has directed KL $\Theta_p(\gamma^2)$ from $P_0$, so the standard two-point lower bound \citep{lecam2000asymptotics} gives $N_w=\Omega_p(1/\gamma^2)$ for every profile. Second, for any witness policy $A_\theta$, feasibility implies
\[
 Q(A_\theta,L=0)\ge \frac{1-\alpha}{2},
 \qquad
 P_0(A_\theta,L=0)=\frac{1-p}{2},
\]
so every law in the full existential null is at least $\gamma/2$ away after the $4B$ half-group/label pushforward. Generic known-distribution identity testing on $4B$ categories therefore gives $N_w=O(\sqrt B/\gamma^2)$, uniformly in $w$. Consequently the ratio between any two complexities in this matched sign-family class is at most $O(\sqrt B)$. Theorem~\ref{thm:separation} attains every $\sqrt{k}$ order along the plateau subclass and in particular attains this maximal endpoint spread.

\subsection{Proof of Theorem~\ref{thm:separation}: plateau spectrum}
For $k<B$, the plateau profile $w_\ell^{(k)}=(1-\varepsilon_B)/k$ ($\ell\le k$), $w_\ell^{(k)}=\varepsilon_B/(B-k)$ ($\ell>k$), $\varepsilon_B=B^{-2}$, satisfies
\[
 \|w^{(k)}\|_2^2
 =\frac{(1-\varepsilon_B)^2}{k}
  +\frac{\varepsilon_B^2}{B-k},
 \qquad \varepsilon_B=B^{-2}.
\]
Uniformly over $1\le k<B$, the second term is $o(1/k)$ and $(1-\varepsilon_B)^2=1+o(1)$; for $k=B$, $\|w^{(B)}\|_2^2=1/B$. Hence $d_{\rm eff}(w^{(k)})=k\{1+o(1)\}$ uniformly in $k$. Proposition~\ref{prop:weight-profile} therefore gives
\[
 N_{B,k}=\Omega\!\left(\frac{\sqrt{k}}{\gamma_B^2}\right).
\]

It remains to prove the matching upper bound against the \emph{full unrestricted} existential null. For $k<B$, let
\[
 R=\bigcup_{\ell=k+1}^B G_\ell,
 \qquad P_0(R)=\varepsilon_B,
\]
and retain separately the two halves of each heavy group $G_1,\ldots,G_k$. Push $(S,L)$ to the $(4k+2)$ categories consisting of heavy-group half/label pairs plus $R\times\{0,1\}$. Let $Q$ be any law in the full null and let $A_\theta$ be a witness. Write
$A_H=A_\theta\cap R^c$ for the accepted halves of the heavy groups.

If $|Q(R)-\varepsilon_B|\ge\gamma_B/16$, the pushed-forward laws already differ in TV by at least $\gamma_B/16$. Otherwise $Q(R)\le\varepsilon_B+\gamma_B/16$. Feasibility of $A_\theta$ gives
\[
 Q(A_\theta,L=0)\ge(1-\alpha)Q(A_\theta)
 \ge\frac{1-p+\gamma_B}{2},
\]
and hence
\begin{align*}
 Q(A_H,L=0)
 &\ge \frac{1-p+\gamma_B}{2}-Q(R),\\
 P_0(A_H,L=0)
 &=\frac{(1-p)(1-\varepsilon_B)}{2}.
\end{align*}
Therefore
\[
 Q(A_H,L=0)-P_0(A_H,L=0)
 \ge \frac{7\gamma_B}{16}
       -\frac{1+p}{2}\varepsilon_B
 \ge \frac{3\gamma_B}{8}
\]
whenever $\varepsilon_B\le\gamma_B/16$. Thus every law in the full null is at TV distance at least $\gamma_B/16$ from $P_0$ after a pushforward with $4k+2$ categories. Known-distribution identity testing on $m=4k+2$ categories uses $O(\sqrt m/\gamma_B^2)=O(\sqrt{k}/\gamma_B^2)$ samples. For $k=B$, the residual is empty and the same argument reduces to the standard $4B$ pushforward. Combining with the lower bound proves Theorem~\ref{thm:separation}, and $d_{\rm eff}(w^{(k)})=k\{1+o(1)\}$ as computed above.

The condition $\varepsilon_B\le\gamma_B/16$ also makes the vanishing-gap statement stable when $\varepsilon_B/\gamma_B$ tends to a nonzero constant below $1/16$: the displayed separation remains a fixed positive fraction of $\gamma_B$, so no hidden constant degenerates in the pushforward upper bound.

\subsection{Balanced q-ary spectrum with positive coverage slack}
\label{app:high-coverage}
The binary construction uses reference coverage $1/2$ only for notation. The following balanced construction simultaneously extends the matched spectrum to rational reference coverage and removes the boundary condition $P_0(A)=\kappa$.

\begin{proposition}[Matched spectrum with positive coverage slack]
\label{prop:rational-kappa}
Fix a rational reference coverage $c=m/q\in(0,1)$ with integers $1\le m<q$, a contract floor $0<\kappa\le c$, $p\in(0,1)$, and $0<\delta<\beta<1$. Let $\gamma=\gamma_B$ satisfy
\[
 \gamma_B<p,
 \qquad
 \frac{c}{1-c}\gamma_B<1-p,
 \qquad
 B^{-2}\le \frac{c\gamma_B}{16}.
\]
For each $k\in\{1,\ldots,B\}$ use the plateau profile $w^{(k)}$ of Section~\ref{sec:iut}. There is a family with
$J_B=\binom qm^B$ policies, common reference $P_0$, common raw support, common contract $(\kappa,p-\gamma_B)$, and exact one-observation hiding such that every policy has reference coverage exactly $c$ and the same unrestricted singleton testing problem, while, uniformly in $k$,
\[
 N^{(c,\kappa)}_{B,k}
 =\Theta_{p,c,\kappa,\delta,\beta}\!\left(\frac{\sqrt{k}}{\gamma_B^2}\right).
\]
Consequently, for every fixed $c>\kappa$ the construction has fixed positive coverage slack $c-\kappa$ for every member, and its endpoint complexity ratio is
$\Theta_c(\sqrt{\log J_B})$.
\end{proposition}

\paragraph{Construction and exact hiding.}
Partition $[0,1]$ into groups $G_1,\ldots,G_B$ of masses $w_1,\ldots,w_B$, and split every group into $q$ equal-mass consecutive subintervals $G_{\ell,1},\ldots,G_{\ell,q}$. A local policy chooses an $m$-subset $C_\ell\subset[q]$; a global policy $\theta=(C_1,\ldots,C_B)$ accepts
\[
 A_\theta=\bigcup_{\ell=1}^B\bigcup_{a\in C_\ell}G_{\ell,a}.
\]
Thus $P_0(A_\theta)=m/q=c$ for every member. In this one-dimensional realization each policy is a union of at most $mB$ intervals, so the construction does not rely on nonmeasurable or arbitrarily fragmented acceptance sets, although the interval budget grows with $B$.

Let
\[
 a_c=\frac{c}{1-c}\gamma_B,
 \qquad
 Q_\theta(L=1\mid S=x)=
 \begin{cases}
 p-\gamma_B,&x\in A_\theta,\\
 p+a_c,&x\notin A_\theta.
 \end{cases}
\]
The parameter conditions keep both Bernoulli probabilities in $(0,1)$. Each canonical component is feasible because it has coverage $c\ge\kappa$ and accepted risk $p-\gamma_B=\alpha$. Under the uniform prior on the $\binom qm^B$ policies, each score subinterval is selected with probability $m/q=c$, hence its average perturbation is
$-c\gamma_B+(1-c)a_c=0$. Therefore the mixture barycenter is exactly $P_0$.

All members have the same singleton problem by Proposition~\ref{prop:singleton}. The canonical components are also matched componentwise across profiles:
\begin{align*}
 \TV(P_0,Q_\theta)&=2c\gamma_B,\\
 \KL(P_0\Vert Q_\theta)
 &=c\,\mathrm{kl}(p\Vert p-\gamma_B)
 +(1-c)\,\mathrm{kl}(p\Vert p+a_c),\\
 \KL(Q_\theta\Vert P_0)
 &=c\,\mathrm{kl}(p-\gamma_B\Vert p)
 +(1-c)\,\mathrm{kl}(p+a_c\Vert p).
\end{align*}
These quantities depend on $(p,c,\gamma_B)$ but not on the group-mass profile.

\paragraph{Product-mixture lower bound.}
For a local accepted subset $C\subset[q]$ define
\[
 h_C(a)=\begin{cases}-1,&a\in C,\\ c/(1-c),&a\notin C,\end{cases}
 \qquad
 Z(C,C')=\frac1q\sum_{a=1}^q h_C(a)h_{C'}(a).
\]
For independent uniform $m$-subsets $C,C'$, $Z$ is centered and bounded by a constant depending only on $c$ (equivalently on the fixed pair $m,q$). Across groups the $Z_\ell$ are independent, and direct Bernoulli calculation gives
\[
 \Xi(\theta,\theta')
 =\frac{\gamma_B^2}{p(1-p)}\sum_{\ell=1}^B w_\ell Z_\ell.
\]
Applying Hoeffding's lemma inside Lemma~\ref{thm:overlap} yields
\[
 1+\chi^2(R_n\Vert P_0^n)
 \le \exp\!\left(C_{p,c}\,n^2\gamma_B^4\sum_{\ell=1}^B w_\ell^2\right).
\]
Corollary~\ref{cor:powerceiling} therefore implies
$N_{B,k}^{(c,\kappa)}=\Omega_{p,c,\delta,\beta}(\sqrt{k}/\gamma_B^2)$ because
$\sum_\ell (w_\ell^{(k)})^2=\Theta(1/k)$ uniformly over the plateau spectrum.

\paragraph{Matching upper bound against the full unrestricted null.}
For $k<B$, let $R=\cup_{\ell>k}G_\ell$, so $P_0(R)=\varepsilon_B=B^{-2}$, retain the $q$ subcells of every heavy group separately, and push $(S,L)$ to the $2(qk+1)$ resulting cell/label categories. Let $Q$ be any law in the full existential null and let $A_\theta$ be a witness. Write $D=A_\theta\cap R^c$ and $d_0=P_0(D)=c(1-\varepsilon_B)$. Put $a=c/16$.

If either $|Q(R)-\varepsilon_B|\ge a\gamma_B$ or $|Q(D)-d_0|\ge a\gamma_B$, the pushed-forward laws already differ in TV by at least $a\gamma_B$. Otherwise
$Q(R)\le\varepsilon_B+a\gamma_B$ and $Q(D)\ge d_0-a\gamma_B$. Feasibility gives
\begin{align*}
 Q(D,L=0)
 &\ge Q(A_\theta,L=0)-Q(R)\\
 &\ge (1-p+\gamma_B)Q(A_\theta)-Q(R)\\
 &\ge (1-p+\gamma_B)Q(D)-Q(R).
\end{align*}
Since $P_0(D,L=0)=(1-p)d_0$,
\begin{align*}
 Q(D,L=0)-P_0(D,L=0)
 &\ge \gamma_B d_0-a\gamma_B(2-p+\gamma_B)-\varepsilon_B\\
 &\ge \frac{11c\gamma_B}{16},
\end{align*}
where the last inequality uses $d_0\ge15c/16$, $2-p+\gamma_B<3$, and
$\varepsilon_B\le c\gamma_B/16$. Thus every full-null law is separated from $P_0$ by at least $c\gamma_B/16$ after a pushforward with $O_c(k)$ categories. Known-distribution identity testing gives
$O_{p,c}(\sqrt{k}/\gamma_B^2)$ samples. For $k=B$ the residual cell is absent and the same argument uses the $2qB$ subcell/label categories. Combining the two bounds proves the proposition.

Taking, for example, $c=9/10$ and any fixed $\kappa<9/10$ gives a high-coverage construction with nonzero contract slack. The theorem does not settle the different regime in which the number of allowed intervals $K$ is fixed while $B\to\infty$; that structured question is left open.

\subsection{Informative nonconstant-reference spectrum}
\label{app:informative-reference}
The score-independent reference in Equation~\eqref{eq:qtheta} is a convenient special case, not a requirement for the separation. The next construction makes the score genuinely informative about risk and makes selective acceptance strictly useful under the reference law.

\begin{proposition}[Matched spectrum under informative nonconstant risk]
\label{prop:informative-spectrum}
Fix constants $a,b,\rho>0$ with $a+b+\rho=1$, risks $p_C,p_V,p_H\in(0,1)$, and $h>0$ such that $p_V\pm h\in(0,1)$. Put
\[
 c=a+\rho/2,\qquad
 r_A=\frac{a p_C+(\rho/2)p_V}{c},\qquad
 \bar p=a p_C+\rho p_V+b p_H,
\]
and assume $r_A<\bar p$. Fix $0<\kappa\le c$ and $0<\delta<\beta<1$. For any sequence $\gamma_B\downarrow0$ such that
\[
 \tau_B:=\frac{2c\gamma_B}{\rho}
 <\min\{p_V-h,\,1-p_V-h\},
 \qquad
 \rho B^{-2}\le \frac{c\gamma_B}{8},
\]
there are $k$-plateau families with $J_B=2^B$ policies, a common reference law with nonconstant $\eta(s)=P_0(L=1\mid S=s)$, common contract $(\kappa,r_A-\gamma_B)$, exact one-observation hiding, and identical unrestricted singleton testing problems for every member, while uniformly in $k\in\{1,\ldots,B\}$,
\[
 N^{\mathrm{inf}}_{B,k}
 =\Theta_{a,b,\rho,p_C,p_V,p_H,h,\kappa,\delta,\beta}
 \!\left(\frac{\sqrt{k}}{\gamma_B^2}\right).
\]
Every policy has reference coverage $c$ and reference conditional risk $r_A<\bar p$, so abstention strictly improves risk under $P_0$.
\end{proposition}

\paragraph{Reference and policy family.}
The reference law must remain literally fixed as the plateau profile changes.  Let the variable region carry product coordinates $(X,Z)\in[0,1]\times\{-1,+1\}$, with conditional reference law $X\sim\mathrm{Unif}[0,1]$, $Z$ uniform and independent of $X$.  Give the disjoint regions $C,V,H$ masses $a,\rho,b$; set $\eta=p_C$ on $C$, $\eta=p_H$ on $H$, and
\[
 \eta(X,Z)=p_V+hZ\qquad\text{on }V.
\]
This defines one common $P_0$ for every profile.  For a mass vector $w$, partition only the $X$ coordinate into groups $G_1,\ldots,G_B$ of masses $w_1,\ldots,w_B$ and split each $G_\ell$ into equal $X$-halves $G_\ell^+,G_\ell^-$.  Each half is crossed with both values of $Z$, so it always contains equal masses at risks $p_V-h$ and $p_V+h$, independently of its width or of the profile.  (A standard nonatomic-space isomorphism can encode these product coordinates in a scalar score if desired; interval structure is not claimed for this informative extension.)

For $\theta\in\{-1,+1\}^B$, let $A_\theta$ accept all of $C$ and $(G_\ell^{\theta_\ell}\times\{-1,+1\})$ from every variable group. Then $P_0(A_\theta)=c$ and every member has reference accepted risk $r_A$; its rejected risk is also profile-independent. In particular the four-cell law of $(\1\{S\in A_\theta\},L)$ is the same for every $\theta$ and every plateau profile, so Proposition~\ref{prop:singleton} gives exact singleton equivalence without any score--loss independence assumption.

\paragraph{Canonical feasible components and exact hiding.}
Let $u_\theta(X,Z)=+1$ on the selected $X$-half of a variable group (for either $Z$) and $-1$ on the unselected half. Define $Q_\theta$ by preserving the score marginal and setting
\[
 Q_\theta(L=1\mid S=x)=
 \begin{cases}
  \eta(x),&x\in C\cup H,\\
  \eta(x)-\tau_B u_\theta(x),&x\in V.
 \end{cases}
\]
The parameter condition keeps all Bernoulli probabilities in $(0,1)$. The accepted error mass decreases by $(\rho/2)\tau_B=c\gamma_B$, hence
$Q_\theta(L=1\mid S\in A_\theta)=r_A-\gamma_B$ and the component is feasible. Under the uniform prior on signs, each variable point is selected and rejected with equal probability, so the perturbation cancels pointwise and $\E_\theta Q_\theta=P_0$ exactly. Canonical TV and both directed KL divergences are independent of the plateau profile because every group contains the same replicated two-level risk pattern.

\paragraph{Overlap lower bound.}
Define
\[
 H_V=\frac12\left\{\frac{1}{(p_V-h)(1-p_V+h)}+\frac{1}{(p_V+h)(1-p_V-h)}\right\}.
\]
By the finite-cell Fisher representation, for independent signs $\theta,\theta'$,
\[
 \Xi(\theta,\theta')
 =\Lambda_B\sum_{\ell=1}^B w_\ell\theta_\ell\theta'_\ell,
 \qquad
 \Lambda_B=\rho\tau_B^2H_V
 =\frac{4c^2H_V}{\rho}\gamma_B^2.
\]
The same Rademacher MGF calculation as Proposition~\ref{prop:weight-profile} therefore gives
$N^{\mathrm{inf}}_{B,k}=\Omega(\sqrt{k}/\gamma_B^2)$ uniformly over the plateau spectrum.

\paragraph{Matching upper bound against the full unrestricted null.}
For $k<B$, collapse the residual variable groups $R=\cup_{\ell>k}G_\ell$, whose reference mass is $P_0(R)=\rho\varepsilon_B$, while retaining $C,H$ and each heavy half separately; retain the label in every cell. This gives $4k+6$ categories. Let $Q$ be any full-null law with witness $A_\theta$, put $A_H=A_\theta\setminus R$, and define on the pushforward
$g=\1\{S\in A_H\}(L-\alpha)$ with $\alpha=r_A-\gamma_B$. Since $P_0(A_\theta)\{r_{P_0}(A_\theta)-\alpha\}=c\gamma_B$,
\[
 \E_{P_0}g\ge c\gamma_B-P_0(R).
\]
Feasibility gives $\E_Q[\1\{S\in A_\theta\}(L-\alpha)]\le0$, hence $\E_Qg\le Q(R)$. If $|Q(R)-P_0(R)|\ge c\gamma_B/4$, the residual-mass coordinate already separates the pushforwards by $c\gamma_B/4$. Otherwise
\[
 \E_{P_0}g-\E_Qg
 \ge \frac{3c\gamma_B}{4}-2P_0(R)
 \ge \frac{c\gamma_B}{2},
\]
using $\rho\varepsilon_B\le c\gamma_B/8$. Because $g$ has oscillation one, the pushed-forward TV distance is therefore at least $c\gamma_B/4$ in all cases. Known-distribution identity testing on $O(k)$ categories gives $O(\sqrt{k}/\gamma_B^2)$ samples; for $k=B$ there is no residual category. This matches the lower bound. All constants are uniform over $k$ and over every admissible sequence $\gamma_B$ satisfying the displayed conditions.

A concrete choice is $(a,\rho,b)=(.30,.50,.20)$ and $(p_C,p_V-h,p_V+h,p_H)=(.08,.16,.32,.60)$, for which $c=.55$, the accepted reference risk is about $.153$, the population risk is $.264$, and the score carries substantial risk information while the matched spectrum remains unchanged in order.

\section{Product-overlap identity and Fisher geometry}
\label{app:overlap}

\begin{lemma}[Product-mixture identity; \citealp{ingster1987minimax}]
\label{thm:overlap}
For dominated feasible components write $K(Q,Q')=\int(dQ/dP_0)(dQ'/dP_0)dP_0$ and $\Xi=K-1$. For a prior $\Pi$ with $\E_\Pi Q=P_0$ and $R_n=\E_\Pi Q^n$,
\begin{equation}
1+\chi^2(R_n\Vert P_0^n)=\E_{Q,Q'\sim\Pi}[1+\Xi(Q,Q')]^n\quad\text{(in the extended sense)}.
\label{eq:overlap}
\end{equation}
\end{lemma}
\subsection{Proof of Lemma~\ref{thm:overlap}}
Write $h_Q=dQ/dP_0$. Since $Q\ll P_0$ almost surely, the product-mixture likelihood ratio is
\[
 \frac{dR_n}{dP_0^n}(x_{1:n})
 =\E_Q\prod_{i=1}^n h_Q(x_i).
\]
All terms are nonnegative, so Tonelli applies without any per-component $L^2$ assumption. With independent $Q,Q'\sim\Pi$,
\begin{align*}
 1+\chi^2(R_n\Vert P_0^n)
 &=\E_{Q,Q'}\prod_{i=1}^n
 \int h_Q(x)h_{Q'}(x)\,dP_0(x)\\
 &=\E_{Q,Q'}K(Q,Q')^n.
\end{align*}
The barycenter condition implies $\E_Q h_Q(x)=1$ for $P_0$-almost every $x$, hence by Tonelli
$\E_{Q,Q'}K(Q,Q')=\int(\E_Q h_Q)^2dP_0=1$. Therefore $K<\infty$ for $\Pi\otimes\Pi$-almost every pair, $R_1=P_0$, and with $\Xi=K-1$,
\[
 \chi^2(R_2\Vert P_0^2)=\E K^2-1=\E(K-1)^2=\E\Xi^2
\]
in the extended sense. This is the standard product-mixture second-moment calculation; the paper-specific content is the selective-contract overlap kernel and its role in the matched separation.

\begin{corollary}[Approximate hiding under TV mismatch]
\label{cor:approx-hiding}
Let $\Pi$ be supported on $\cH(\cA)$, write $\bar Q=\E_{Q\sim\Pi}Q$, and suppose $\TV(P_0,\bar Q)\le\varepsilon$. Assume $Q\ll\bar Q$ almost surely and define $\bar K(Q,Q')=\int(dQ/d\bar Q)(dQ'/d\bar Q)\,d\bar Q$ and $\bar\Xi=\bar K-1$. Then every level-$\delta$ full-null-valid test satisfies
\begin{equation}
 \E_{P_0^n}\phi\le \delta+1-(1-\varepsilon)^n+\frac12\sqrt{\E[1+\bar\Xi(Q,Q')]^n-1}
 \le \delta+n\varepsilon+\frac12\sqrt{\E[1+\bar\Xi(Q,Q')]^n-1}.
 \label{eq:approx-hiding}
\end{equation}
\end{corollary}

\begin{corollary}[Approximate hiding under quadratic mismatch]
\label{cor:approx-hiding-chi2}
Under the same mixture assumptions, additionally suppose $P_0\ll\bar Q$ and $\chi^2(P_0\Vert\bar Q)\le\rho$. Then every level-$\delta$ full-null-valid test satisfies
\begin{equation}
 \E_{P_0^n}\phi\le\delta+\frac12\sqrt{(1+\rho)^n-1}+\frac12\sqrt{\E[1+\bar\Xi(Q,Q')]^n-1}.
 \label{eq:approx-hiding-chi2}
\end{equation}
\end{corollary}

\subsection{Proof of Corollary~\ref{cor:approx-hiding}}
Let $\bar Q=\E_{Q\sim\Pi}Q$ and $R_n=\E_{Q\sim\Pi}Q^n$. For any level-$\delta$ full-null-valid test,
\[
 \E_{P_0^n}\phi\le \E_{R_n}\phi+\TV(P_0^n,R_n)\le \delta+\TV(P_0^n,R_n).
\]
By the triangle inequality,
\[
 \TV(P_0^n,R_n)\le \TV(P_0^n,\bar Q^n)+\TV(\bar Q^n,R_n).
\]
A maximal coupling of one draw from $P_0$ and $\bar Q$ disagrees with probability $\TV(P_0,\bar Q)\le\varepsilon$; coupling the $n$ coordinates independently gives
\[
 \TV(P_0^n,\bar Q^n)\le 1-(1-\varepsilon)^n\le n\varepsilon.
\]
Finally, $\bar Q$ is the exact barycenter of $\Pi$, so Lemma~\ref{thm:overlap} applied with reference $\bar Q$ gives
\[
 \TV(\bar Q^n,R_n)
 \le \frac12\sqrt{\chi^2(R_n\Vert \bar Q^n)}
 =\frac12\sqrt{\E[1+\bar\Xi(Q,Q')]^n-1}.
\]
Combining the three displays proves \eqref{eq:approx-hiding}. The TV statement is intentionally assumption-light; its linear product penalty is useful only when $n\varepsilon$ is small.

\subsection{Proof of Corollary~\ref{cor:approx-hiding-chi2}}
With the same notation and triangle inequality,
\[
 \TV(P_0^n,R_n)
 \le \TV(P_0^n,\bar Q^n)+\TV(\bar Q^n,R_n).
\]
The second term is controlled exactly as above by Lemma~\ref{thm:overlap} with reference $\bar Q$. For the first term, product likelihood ratios give
\[
 1+\chi^2(P_0^n\Vert\bar Q^n)
 =\{1+\chi^2(P_0\Vert\bar Q)\}^n
 \le (1+\rho)^n.
\]
Therefore $\TV(P_0^n,\bar Q^n)\le\frac12\sqrt{(1+\rho)^n-1}$. Combining the two terms and test validity proves \eqref{eq:approx-hiding-chi2}. When $n\rho=o(1)$, $(1+\rho)^n-1=O(n\rho)$, giving an $O(\sqrt{n\rho})$ mismatch contribution.

When exact one-observation hiding is possible, minimizing the witness bound in Corollary~\ref{cor:powerceiling} over all hiding priors yields the family-level obstruction
\begin{equation}
 \Omega_n(P_0,\cH)=\inf_{\Pi:\,\Pi(\cH)=1,\;\E_{Q\sim\Pi}[Q]=P_0}
 \TV\!\left(P_0^n,\E_{Q\sim\Pi}Q^n\right),
 \label{eq:canonical-overlap-obstruction}
\end{equation}
so every level-$\delta$ full-null-valid audit has power at most $\delta+\Omega_n(P_0,\cH)$. Lemma~\ref{thm:overlap} evaluates any chosen hiding witness, while the infimum above removes dependence on a particular decomposition. We do not assume this infimum is attained without additional compactness; in the finite constructions used below, the displayed witness prior is explicit and order-tight.

\subsection{Finite-cell Fisher representation}
Assume component $Q_k$ preserves cell masses $\mu_j$ and has Bernoulli error rate $\theta_{k,j}$ in cell $j$, while $P_0$ has rate $\eta_j\in(0,1)$. Directly evaluating the two label states shows
\[
 \Xi(Q_k,Q_\ell)
 =\sum_j\mu_j\frac{(\theta_{k,j}-\eta_j)(\theta_{\ell,j}-\eta_j)}{\eta_j(1-\eta_j)}
 =\sum_j\mu_jg_k(j)g_\ell(j).
\]
Thus the overlap matrix is exactly a Gram matrix in the local Bernoulli Fisher metric.

\section{Auxiliary testing bounds}
\label{app:aux}
For a prior $\Pi$ over feasible laws with $\E_\Pi Q=P_0$ (a hiding prior) and $R_n=\E_\Pi Q^n$:
\begin{corollary}[Witness power ceiling]
\label{cor:powerceiling}
For every full-null-valid level-$\delta$ test and any hiding prior,
\begin{equation}
 \E_{P_0^n}\phi\le\delta+\TV(P_0^n,R_n)
 \le\delta+\tfrac12\sqrt{\E[1+\Xi(Q,Q')]^n-1}.
 \label{eq:powerceiling}
\end{equation}
\end{corollary}
\begin{proof}
Every component of the prior is feasible, so a level-$\delta$ test satisfies $\E_{R_n}\phi\le\delta$. Hence $\E_{P_0^n}\phi\le\delta+\TV(P_0^n,R_n)$; $\TV\le\frac12\sqrt{\chi^2}$ and Lemma~\ref{thm:overlap} give the second inequality.
\end{proof}

\subsection{A collision bound for supported nulls}
Let $p_i>0$ on $d\ge2$ reference-positive cells and define $\cH_+$ as the feasible categorical laws that put zero mass on all reference-zero cells. Assume $\cH_+$ is nonempty and
\[
 \Delta=\min_{q\in\cH_+}\sum_i(q_i-p_i)^2/p_i>0,\quad
 p_{\min}=\min_i p_i,\quad u_0=1+\sqrt{\Delta/p_{\min}}.
\]
The infimum is over members, not their one-observation convex hull.

\begin{theorem}[Supported collision bracket]
\label{thm:collision}
Set $\epsilon=\min\{\delta,1-\beta\}$ and
$c_{\delta,\beta}=\min\{1/2,\frac14\log[1/(1-(\beta-\delta)^2)]\}$. Then
\[
 \frac{c_{\delta,\beta}}{\Delta}\le N(p,\cH_+;\delta,\beta)
 \le \max\!\left\{2,\left\lceil\frac{32}{\epsilon}\frac{\max\{\sqrt{d-1},u_0\}}{\Delta}\right\rceil\right\}.
\]
The lower bound also holds for the full null. The upper bound is achieved by rejecting feasibility when $U_m<\Delta/2$, where
\[
 U_m={\binom m2}^{-1}\sum_{a<b}\left(\frac{\1\{X_a=X_b\}}{p_{X_a}}-1\right).
\]
\end{theorem}
\begin{proof}
Let $f(i,j)=\1\{i=j\}/p_i-1$. Under $p$, $\sum_jp_jf(i,j)=0$ and $\E_{p\otimes p}f^2=d-1$. Thus $\E_p U_m=0$ and $\operatorname{Var}_p(U_m)=(d-1)/\binom m2$. Under a component $q$, let $D=\chi^2(q\Vert p)\ge\Delta$, $r_i=q_i/p_i$ and $u=\max_i r_i$. Then $\E_qU_m=D$,
\[
 \zeta_1=\operatorname{Var}_q(r_i-1)\le uD,\qquad
 \zeta_2\le\E_{q\otimes q}f^2\le4(d-1)+4D/p_{\min}+1.
\]
For the last inequality split the sum $\sum_i r_i^2(1-p_i)^2$ at $r_i=2$, use $r_i^2\le4(r_i-1)^2$ on the second part, and bound the off-diagonal contribution by one. The U-statistic decomposition gives
\[
 \operatorname{Var}_q(U_m)\le4\zeta_1/m+2\zeta_2/[m(m-1)].
\]
Chebyshev therefore bounds the reference error by $16(d-1)/(m^2\Delta^2)$ and the component error by $16u/(mD)+16\zeta_2/(m^2D^2)$ for $m\ge2$. Since $D\ge p_{\min}(u-1)^2$, we have $u/D\le u_0/\Delta$. Also
\[
 \sqrt{\zeta_2}/D\le[2\sqrt{d-1}+2(u_0-1)+1]/\Delta
 \le5\max\{\sqrt{d-1},u_0\}/\Delta.
\]
The stated constant controls both errors by $\epsilon$. Validity is componentwise and therefore survives mixing. For the lower bound use a minimizing $q$ and $H^2(p,q)\le\chi^2(q\Vert p)=\Delta$. The two-point inequality gives
$N\ge(1-H^2)/(2H^2)\log[1/(1-(\beta-\delta)^2)]$.
If $\Delta\le1/2$ this is at least the claimed lower bound; if $\Delta>1/2$, use $N\ge1$ and $c_{\delta,\beta}\le1/2$.
\end{proof}

\paragraph{Reference-zero cells.}
The collision statistic is evaluated only on reference-positive cells. For a full-null extension let $Z$ be the reference-zero cells, $\rho(q)=q(Z)$, and
\[
 \Delta(\rho_0)=\inf_{q\in\cH:\rho(q)\le\rho_0}
 \chi^2(q(\cdot\mid Z^c)\Vert p).
\]
Combine the collision rejection with the requirement that no sample falls in $Z$. If $\rho(q)>\rho_0$, rejection probability is at most $(1-\rho(q))^m\le e^{-m\rho_0}$. Otherwise, conditional on avoiding $Z$, the samples are i.i.d. from $q(\cdot\mid Z^c)$ and the collision proof applies with $\Delta(\rho_0)$. Thus a valid full-null upper bound uses the maximum of the collision sample bound at this radius and $\log(1/\epsilon)/\rho_0$, whenever $\Delta(\rho_0)>0$. We do not substitute the supported $\Delta$ for this radius in our empirical claims.

\section{Reference-free intersection--union certificate}
\label{app:iut}

For a fixed policy $A$, let $c(A)=P(S\in A)$ and $r(A)=P(L=1\mid S\in A)$. Then $T_A\sim\mathrm{Bin}(n,c(A))$ and, conditional on $T_A=t$, $E_A\sim\mathrm{Bin}(t,r(A))$. Let $U_A$ be a one-sided $(1-\zeta_c)$ exact upper bound on $c(A)$ and $L_A$ a one-sided $(1-\zeta_r)$ exact lower bound on $r(A)$, with $\zeta_c+\zeta_r\le\delta$.

\subsection{Proof of Proposition~\ref{prop:iut}}
If the existential null is true, choose a fixed population witness $A^*$ satisfying $c(A^*)\ge\kappa$ and $r(A^*)\le\alpha$. This witness depends on the population, not on the sample. If the procedure rejects, then either $U_{A^*}<\kappa\le c(A^*)$, or $U_{A^*}\ge\kappa$, in which case $A^*$ participates in the minimization and rejection forces $L_{A^*}>\alpha\ge r(A^*)$. The two one-sided bounds fail with probabilities at most $\zeta_c$ and $\zeta_r$, respectively. A union bound over these two events gives at most $\zeta_c+\zeta_r\le\delta$. No event is required to hold simultaneously for all searched policies. The equal split used in our experiments is therefore a convenience rather than a requirement; any predeclared allocation is valid. More generally, a fixed-policy joint confidence region for $(c(A),r(A))$ can replace the rectangular pair whenever its simultaneous failure probability is at most $\delta$ and the corresponding coverage/risk projections are used in the same screening logic.

\begin{proposition}[Anytime-valid existential certificate]
\label{prop:anytime-iut}
For each fixed $A$, replace $(U_A,L_A)$ by one-sided confidence sequences $(U_{A,t},L_{A,t})$ whose anytime failure probabilities are at most $\zeta_c$ for coverage and $\zeta_r$ for risk. Reject at the first time $t$ for which $\min_{A:U_{A,t}\ge\kappa}L_{A,t}>\alpha$. Under the existential null, the probability of ever rejecting is at most $\zeta_c+\zeta_r\le\delta$, again without a multiplicity correction over $\cA$.
\end{proposition}

\subsection{Proof of Proposition~\ref{prop:anytime-iut}}
If the existential null is true, fix a population witness $A^*$. Let $\mathcal E_c$ be the event that the coverage upper confidence sequence ever falls below the true $c(A^*)$, and let $\mathcal E_r$ be the event that the risk lower confidence sequence ever rises above the true $r(A^*)$. By assumption, their probabilities are at most $\zeta_c$ and $\zeta_r$, respectively. If the anytime procedure ever rejects while neither event occurs, then $U_{A^*,t}\ge c(A^*)\ge\kappa$, so $A^*$ is in the screened family, while $L_{A^*,t}\le r(A^*)\le\alpha$, contradicting rejection. Hence ever-rejection is contained in $\mathcal E_c\cup\mathcal E_r$ and has probability at most $\zeta_c+\zeta_r\le\delta$. The argument requires simultaneous-in-time validity only for the one fixed population witness, not for all searched policies.

\paragraph{Why a one-policy split-sample comparator is not a valid existential-null test.}
Selecting one $A$ on one split and rejecting that policy on an independent split proves only that the selected member is infeasible. The null is a union over all $A\in\cA$, so another member may still be the population witness. A valid split-sample existential test would still need to reject every potential witness (or otherwise control the union-null logic), restoring the structural problem addressed by the IUT. A ``select one, then test one'' comparator therefore tests a different null hypothesis, and we do not report it.

\subsection{Operational power on theorem-admissible matched families}
\label{app:iut-overlap-power}
An additional finite-size control uses $B=16$, $p=.25$, $\gamma=.08$, $\alpha=.17$, $\kappa=.5$, and $k=1,4,16$. The residual mass is $B^{-2}=1/256$, which is at most $\gamma/16=.005$, as required by Theorem~\ref{thm:separation}. The $k=16$ profile is uniform. For every $(k,n)$, generate independent multinomial counts over the $2B$ group halves and then binomial errors with common probability $p$. The boundary-null control instead uses error probability $p-\gamma$ on one selected half of each group and $p+\gamma$ on the other; the selected policy is a population witness.

The sample grid is $n\in\{300,500,750,1000,1400,2000\}$ for alternatives and $n\in\{500,1400\}$ for boundary checks, with 2,000 replicates per point. The artifact records all deterministic seeds. Error bars are pointwise Wilson 95\% intervals for Monte Carlo rejection probabilities. The reported $n_{80}$ values linearly interpolate neighboring power estimates; no confidence interval or minimax interpretation is attached to a crossing. The endpoint ratio is $1318.11/572.29=2.303$. All grid values, including null controls, are retained below.
\begin{table}[ht]
\caption{Additional sign-family simulation: observed rejection percentages, 2,000 replicates per cell. Null rows refer to a feasible boundary component.}
\label{tab:sign-power}
\centering\small
\begin{tabular}{llrrr}
\toprule
Source & $n$ & $k=1$ & $k=4$ & $k=16$\\
\midrule
Alternative & 300 & 42.75 & 16.95 & 0.75\\
Alternative & 500 & 75.20 & 47.60 & 7.00\\
Alternative & 750 & 91.80 & 77.00 & 26.85\\
Alternative & 1000 & 97.45 & 92.95 & 54.75\\
Alternative & 1400 & 99.90 & 98.90 & 86.50\\
Alternative & 2000 & 99.95 & 99.95 & 98.35\\
Boundary null & 500 & 3.50 & 2.15 & 0.25\\
Boundary null & 1400 & 3.95 & 2.80 & 1.30\\
\bottomrule
\end{tabular}
\end{table}

\paragraph{Sign-family dynamic program.}
Write $n_{bh},e_{bh}$ for counts and errors in half $h\in\{0,1\}$ of group $b$. Initialize $D_0(0)=0$ and $D_0(t)=+\infty$ otherwise. The exact recurrence is
\[
 D_b(t)=\min_{h\in\{0,1\}}\{D_{b-1}(t-n_{bh})+e_{bh}\},
\]
with invalid indices assigned infinity. After all $B$ groups, $D_B(t)$ is the least errors among policies accepting exactly $t$ observations. The lower binomial bound is monotone in errors, so applying the coverage screen and risk bound to $D_B$ computes the IUT over all $2^B$ policies in $O(Bn)$ time and $O(n)$ memory. Exhaustive tests compare every state and final decision with direct policy enumeration on small instances. This recurrence changes computation only; Proposition~\ref{prop:iut} supplies validity.

\paragraph{Coverage-floor and audit-size sensitivity.}
On the held-out ImageNetV2 audit with $K=16$ and cells frozen on disjoint calibration data, we also evaluate $\kappa\in\{.50,.60,.75,.90\}$ and audit sizes $n\in\{500,1000,2500,5000\}$. For the full $n=5000$ audit the exact 95\% IUT lower frontiers are 4.06\%, 6.28\%, 10.89\%, and 17.71\%, respectively. In 40 deterministic held-out subsamples, the corresponding median frontiers at $n=500$ are 0.97\%, 2.62\%, 6.70\%, and 13.55\%; at $n=1000$ they are 2.41\%, 4.20\%, 8.79\%, and 15.64\%. This is a finite-benchmark sensitivity analysis, not a population learning curve, but it makes explicit how the certificate weakens at lower coverage floors and smaller audit sizes.

\subsection{Weighted interval dynamic program}
Let cell $m$ contain $n_m$ audit observations and $e_m$ errors. A policy using at most $K$ contiguous intervals selects a subset of cells that is a union of at most $K$ contiguous intervals. Define
\[
 e_K(t)=\min\left\{\sum_{m\in A}e_m:
 \sum_{m\in A}n_m=t,\; A\text{ uses at most }K\text{ intervals}\right\}.
\]
A standard dynamic program over cell index, selected count, number of opened intervals, and whether the current cell is inside an interval computes all $e_K(t)$ exactly. If zero-count cells are possible, an additional one-bit state distinguishes the structurally empty mask from a nonempty rule that happens to accept no audit observation; this leaves the $O(MKn)$ state and $O(Kn)$ memory bounds unchanged. Because the one-sided binomial lower bound is monotone in the observed error count, the IUT objective reduces to the one-dimensional minimization reported in Section~\ref{sec:iut}. In the reported experiments the coverage floors are large enough that a $t=0$ rule cannot pass the coverage screen, but the implementation check retains the distinction for completeness.

\subsection{Structured DKW simultaneous baseline}
\label{app:dkw-baseline}
To compare the IUT against a policy-count-free baseline, let $F(x)=P(S\le x)$ and
$G(x)=P(S\le x,L=1)$. The latter is the CDF on the extended line of the random variable equal to $S$ when $L=1$ and $+\infty$ otherwise. By the Dvoretzky--Kiefer--Wolfowitz inequality with Massart's constant \citep{massart1990dkw}, allocating failure probability $\delta/2$ to each CDF gives, simultaneously,
\[
 \|F_n-F\|_\infty\le\epsilon_n,
 \qquad
 \|G_n-G\|_\infty\le\epsilon_n,
 \qquad
 \epsilon_n=\sqrt{\frac{\log(4/\delta)}{2n}}.
\]
For a union of at most $K$ intervals, telescoping its $2K$ endpoints gives
$|c_n(A)-c(A)|\le\rho_K$ and $|m_n(A)-m(A)|\le\rho_K$ with
$\rho_K=2K\epsilon_n$, where $m(A)=P(S\in A,L=1)$. Hence every population policy with $c(A)\ge\kappa$ must satisfy $c_n(A)\ge\kappa-\rho_K$ and
\[
 r(A)=\frac{m(A)}{c(A)}
 \ge
 \frac{[m_n(A)-\rho_K]_+}{\min\{1,c_n(A)+\rho_K\}}.
\]
The same weighted interval DP minimizes this expression exactly over the empirical $(t,e_K(t))$ frontier. This comparator exploits the one-dimensional interval structure and contains no factor $|\cA_{M,K}|$; it is nevertheless conservative when $K\epsilon_n$ is not small.

\subsection{VC/Rademacher simultaneous baseline}
\label{app:rademacher-baseline}
The class of unions of at most $K$ intervals on a line has VC dimension $2K$. To compare against a standard uniform-convergence route rather than only endpoint DKW or policywise Bonferroni, define
$\mathcal F_K=\{(s,l)\mapsto\mathbf 1\{s\in A\}:A\in\cA_{M,K}\}$ and
$\mathcal G_K=\{(s,l)\mapsto l\mathbf 1\{s\in A\}:A\in\cA_{M,K}\}$.
For either $[0,1]$-valued class $\mathcal F$, the standard empirical-Rademacher inequality gives, with probability at least $1-\eta$,
\[
 \sup_{f\in\mathcal F}\{Pf-P_nf\}
 \le 2\widehat{\mathfrak R}_n(\mathcal F)
 +3\sqrt{\frac{\log(2/\eta)}{2n}},
\]
and the reverse deviation follows by applying the same bound to $-\mathcal F$ \citep{mohri2018foundations}. We allocate the data failure budget $0.049$ over the four one-sided events for $\mathcal F_K$ and $\mathcal G_K$. The two empirical Rademacher expectations are evaluated by 200,000 independent sign draws; for each draw the supremum over at most $K$ intervals is solved exactly by a vectorized interval DP. A Hoeffding upper confidence radius with failure probability $5\times10^{-4}$ is added to each Monte-Carlo mean, leaving total data-plus-numerical failure probability at most $0.05$.

Let the resulting two-sided radii be $\epsilon_c(K)$ for coverage and $\epsilon_m(K)$ for accepted-error mass. Then every population policy with $c(A)\ge\kappa$ obeys
\[
 c_n(A)\ge\kappa-\epsilon_c(K),\qquad
 r(A)\ge\frac{[m_n(A)-\epsilon_m(K)]_+}{\min\{1,c_n(A)+\epsilon_c(K)\}},
\]
and the same weighted interval DP minimizes this lower bound over the class. This is a generic VC/Rademacher comparator, not a claim that the bound is sharp for interval unions.

\section{Experimental details and complete empirical tables}
\label{app:experiments}

\subsection{Matched-family efficiency simulation}
\label{app:large-simulation}
\begin{table}[t]
\caption{Measured 80\%-power sample crossings for fixed $B=256$. NP denotes the simple-mixture envelope, not a full-null-valid test. Ratios compare IUT to that envelope; no minimax constant is inferred.}
\label{tab:scaling}
\centering\small
\begin{tabular}{rrrrrr}
\toprule
 & \multicolumn{3}{c}{$\gamma=.04$} & \multicolumn{2}{c}{$\gamma=.02$}\\
$k$ & IUT & NP & ratio & IUT & NP\\
\midrule
1 & 2,342 & 1,001 & 2.34 & 9,216 & 4,030\\
4 & 3,298 & 1,478 & 2.23 & 13,437 & 5,883\\
16 & 5,370 & 2,363 & 2.27 & 22,386 & 9,374\\
64 & 12,065 & 4,060 & 2.97 & 48,268 & 15,859\\
256 & 32,796 & 7,384 & 4.44 & 131,679 & 29,637\\
\bottomrule
\end{tabular}
\end{table}

We fix $B=256$, $p=.25$, $\kappa=.5$, $k\in\{1,4,16,64,256\}$ and $\gamma\in\{.04,.02\}$. The finite sample-size grid for IUT multiplies $(.12k+.9\sqrt{k}+1.5)/\gamma^2$ by $.4,.6,.8,1,1.25,1.6,2$, rounded to multiples of 50; every crossing is bracketed. These grids are computational design choices, not fitted rate claims. Each of the 70 alternative grid points uses 1,500 independent audits. Ten boundary checks, one per $(k,\gamma)$, use the grid point nearest observed 80\% power, a boundary-null law and independent randomness, with 20,000 audits each (\texttt{boundary\_all.py}). For $k=1,4,16,64,256$ the rejection rates are 4.73, 2.75, 1.44, 0.37 and 0.00\% at $\gamma=.04$ and 4.73, 2.59, 1.66, 0.29 and 0.00\% at $\gamma=.02$; the largest Wilson upper limit is 5.03\%. The alternatives use 105,000 IUT audits in total. Seeds are recorded in the scripts and depend on the configuration.

For the NP envelope, the grid multiplies $(.7\sqrt{k}+1)/\gamma^2$ by $.5,.7,.9,1.1,1.35,1.65,2$, rounded to multiples of 25. At each of 70 points, 12,000 independent mixture draws calibrate a nominal 5\% likelihood-ratio threshold; 6,000 fresh reference draws estimate power and 6,000 fresh mixture draws estimate size. The factorized likelihood ratio is invariant under flipping any group half, so sampling the all-$+$ feasible component gives the same LR distribution as sampling the uniform component mixture. We test this symmetry directly. Calibration is independent of both evaluations. Pointwise Wilson intervals for the evaluations condition on the calibrated threshold and do not include its calibration uncertainty. The empirical size column is retained; the fitted threshold is not asserted to have exact 5\% size. The population NP envelope bounds full-null-valid power, but the estimated curve is descriptive.

For exact computation, subtract the smaller half count in every group from the accepted-count state. The reachable width is then at most $1+\sum_b|n_{b,+}-n_{b,-}|$, rather than $n+1$. A compiled implementation performs the same two-choice recurrence over this interval, with unreachable states at infinity. This compression is exact and is checked against the independent uncompressed dynamic program, including final binomial decisions. Source code and a compiler invocation are supplied; no prebuilt binary is needed.

\begin{figure}[ht]
\centering\includegraphics[width=\linewidth]{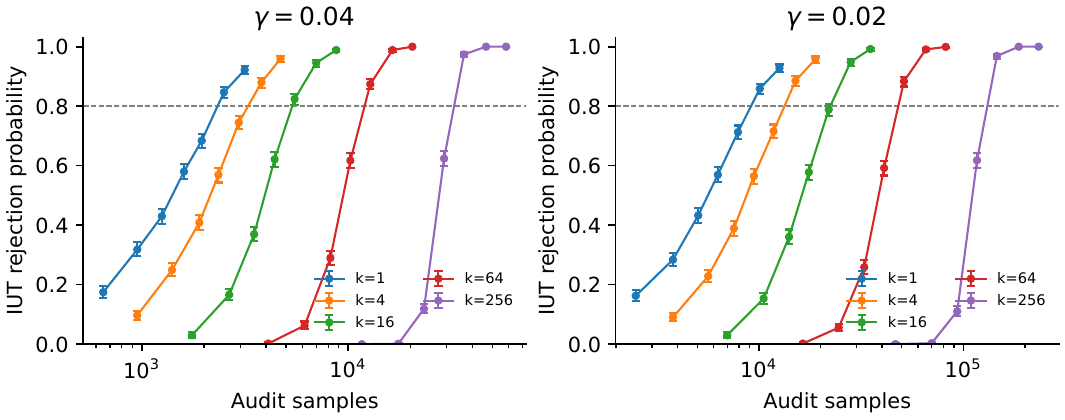}
\caption{Exact IUT power on the two risk-gap grids. Bars are pointwise Wilson 95\% intervals over 1,500 audits. These are Monte Carlo uncertainties, not confidence bands over the sample-size grid.}
\end{figure}
The recorded constructive-witness fraction selects the half with more correct observations in each heavy group and a fixed half in residual groups. Whenever it has accepted count at least $n/2$ and empirical risk at most $\alpha$, IUT rejection is impossible; the simulation checks this deterministic implication. The theorem does not require this particular witness to be optimal.

\subsection{Structured certificate comparison on all frozen outputs}
\paragraph{Semi-synthetic power study.} The protocol is described in Section~\ref{sec:experiments}. Scripts: \texttt{semisynthetic\_power.py}, \texttt{make\_semisyn\_table.py}, \texttt{make\_semisyn\_full.py}. Table~\ref{tab:semisyn-full} lists the IUT for every $K$ and the other certificates at $K=4$.
{\scriptsize\begin{longtable}{llrrrrrrr}
\caption{Semi-synthetic audits, all configurations ($\kappa=.9$; 500 audits per cell for $n\le5{,}000$, 300 at $n=10{,}000$, 200 at $n=20{,}000$). Entries are rejection rates at $\alpha=r^*-\Delta$; the column $\Delta=0$ is the level at the true boundary claim. For CIFAR-10, $\Delta=.10$ gives $\alpha<0$ and is rejected trivially.}\label{tab:semisyn-full}\\
\toprule
Population & Method & $n$ & $K$ & $\Delta=0$ & $.02$ & $.05$ & $.10$ & median frontier\\
\midrule\endfirsthead
\toprule Population & Method & $n$ & $K$ & $\Delta=0$ & $.02$ & $.05$ & $.10$ & median\\\midrule\endhead
CIFAR-10 & IUT & 500 & 1 & 0.006 & 0.13 & 0.92 & 1.00 & 0.044\\
CIFAR-10 & IUT & 500 & 4 & 0.000 & 0.06 & 0.92 & 1.00 & 0.037\\
CIFAR-10 & Bonferroni & 500 & 4 & 0.000 & 0.00 & 0.00 & 1.00 & 0.003\\
CIFAR-10 & DKW & 500 & 4 & 0.000 & 0.00 & 0.00 & 1.00 & 0.000\\
CIFAR-10 & Total loss & 500 & 4 & 0.000 & 0.00 & 0.03 & 1.00 & 0.000\\
CIFAR-10 & IUT & 500 & 16 & 0.000 & 0.05 & 0.92 & 1.00 & 0.037\\
CIFAR-10 & IUT & 1,000 & 1 & 0.004 & 0.24 & 1.00 & 1.00 & 0.052\\
CIFAR-10 & IUT & 1,000 & 4 & 0.000 & 0.20 & 1.00 & 1.00 & 0.048\\
CIFAR-10 & Bonferroni & 1,000 & 4 & 0.000 & 0.00 & 0.00 & 1.00 & 0.013\\
CIFAR-10 & DKW & 1,000 & 4 & 0.000 & 0.00 & 0.00 & 1.00 & 0.000\\
CIFAR-10 & Total loss & 1,000 & 4 & 0.000 & 0.00 & 0.02 & 1.00 & 0.003\\
CIFAR-10 & IUT & 1,000 & 16 & 0.000 & 0.19 & 1.00 & 1.00 & 0.048\\
CIFAR-10 & IUT & 2,500 & 1 & 0.004 & 0.67 & 1.00 & 1.00 & 0.061\\
CIFAR-10 & IUT & 2,500 & 4 & 0.000 & 0.73 & 1.00 & 1.00 & 0.058\\
CIFAR-10 & Bonferroni & 2,500 & 4 & 0.000 & 0.00 & 0.90 & 1.00 & 0.030\\
CIFAR-10 & DKW & 2,500 & 4 & 0.000 & 0.00 & 0.00 & 1.00 & 0.000\\
CIFAR-10 & Total loss & 2,500 & 4 & 0.000 & 0.00 & 0.03 & 1.00 & 0.010\\
CIFAR-10 & IUT & 2,500 & 16 & 0.000 & 0.73 & 1.00 & 1.00 & 0.058\\
CIFAR-10 & IUT & 5,000 & 1 & 0.000 & 0.95 & 1.00 & 1.00 & 0.064\\
CIFAR-10 & IUT & 5,000 & 4 & 0.000 & 0.99 & 1.00 & 1.00 & 0.062\\
CIFAR-10 & Bonferroni & 5,000 & 4 & 0.000 & 0.00 & 1.00 & 1.00 & 0.041\\
CIFAR-10 & DKW & 5,000 & 4 & 0.000 & 0.00 & 0.00 & 1.00 & 0.000\\
CIFAR-10 & Total loss & 5,000 & 4 & 0.000 & 0.00 & 0.01 & 1.00 & 0.013\\
CIFAR-10 & IUT & 5,000 & 16 & 0.000 & 0.99 & 1.00 & 1.00 & 0.062\\
CIFAR-100 & IUT & 500 & 1 & 0.008 & 0.06 & 0.42 & 0.99 & 0.212\\
CIFAR-100 & IUT & 500 & 4 & 0.002 & 0.02 & 0.23 & 0.97 & 0.199\\
CIFAR-100 & Bonferroni & 500 & 4 & 0.000 & 0.00 & 0.00 & 0.00 & 0.090\\
CIFAR-100 & DKW & 500 & 4 & 0.000 & 0.00 & 0.00 & 0.00 & 0.000\\
CIFAR-100 & Total loss & 500 & 4 & 0.004 & 0.03 & 0.25 & 0.94 & 0.195\\
CIFAR-100 & IUT & 500 & 16 & 0.002 & 0.02 & 0.21 & 0.96 & 0.197\\
CIFAR-100 & IUT & 1,000 & 1 & 0.002 & 0.10 & 0.78 & 1.00 & 0.228\\
CIFAR-100 & IUT & 1,000 & 4 & 0.002 & 0.04 & 0.63 & 1.00 & 0.217\\
CIFAR-100 & Bonferroni & 1,000 & 4 & 0.000 & 0.00 & 0.00 & 0.01 & 0.134\\
CIFAR-100 & DKW & 1,000 & 4 & 0.000 & 0.00 & 0.00 & 0.00 & 0.000\\
CIFAR-100 & Total loss & 1,000 & 4 & 0.000 & 0.02 & 0.38 & 1.00 & 0.207\\
CIFAR-100 & IUT & 1,000 & 16 & 0.002 & 0.04 & 0.63 & 1.00 & 0.217\\
CIFAR-100 & IUT & 2,500 & 1 & 0.004 & 0.30 & 1.00 & 1.00 & 0.241\\
CIFAR-100 & IUT & 2,500 & 4 & 0.000 & 0.21 & 0.99 & 1.00 & 0.234\\
CIFAR-100 & Bonferroni & 2,500 & 4 & 0.000 & 0.00 & 0.00 & 0.96 & 0.179\\
CIFAR-100 & DKW & 2,500 & 4 & 0.000 & 0.00 & 0.00 & 0.00 & 0.000\\
CIFAR-100 & Total loss & 2,500 & 4 & 0.000 & 0.01 & 0.68 & 1.00 & 0.217\\
CIFAR-100 & IUT & 2,500 & 16 & 0.000 & 0.20 & 0.99 & 1.00 & 0.234\\
CIFAR-100 & IUT & 5,000 & 1 & 0.014 & 0.62 & 1.00 & 1.00 & 0.249\\
CIFAR-100 & IUT & 5,000 & 4 & 0.008 & 0.55 & 1.00 & 1.00 & 0.243\\
CIFAR-100 & Bonferroni & 5,000 & 4 & 0.000 & 0.00 & 0.09 & 1.00 & 0.203\\
CIFAR-100 & DKW & 5,000 & 4 & 0.000 & 0.00 & 0.00 & 0.00 & 0.000\\
CIFAR-100 & Total loss & 5,000 & 4 & 0.000 & 0.00 & 0.91 & 1.00 & 0.223\\
CIFAR-100 & IUT & 5,000 & 16 & 0.008 & 0.54 & 1.00 & 1.00 & 0.243\\
CIFAR-100 & IUT & 10,000 & 1 & 0.010 & 0.92 & 1.00 & 1.00 & 0.255\\
CIFAR-100 & IUT & 10,000 & 4 & 0.000 & 0.90 & 1.00 & 1.00 & 0.249\\
CIFAR-100 & Bonferroni & 10,000 & 4 & 0.000 & 0.00 & 0.98 & 1.00 & 0.221\\
CIFAR-100 & DKW & 10,000 & 4 & 0.000 & 0.00 & 0.00 & 0.00 & 0.042\\
CIFAR-100 & Total loss & 10,000 & 4 & 0.000 & 0.00 & 1.00 & 1.00 & 0.226\\
CIFAR-100 & IUT & 10,000 & 16 & 0.000 & 0.90 & 1.00 & 1.00 & 0.249\\
CIFAR-100 & IUT & 20,000 & 1 & 0.015 & 0.99 & 1.00 & 1.00 & 0.260\\
CIFAR-100 & IUT & 20,000 & 4 & 0.000 & 0.99 & 1.00 & 1.00 & 0.254\\
CIFAR-100 & Bonferroni & 20,000 & 4 & 0.000 & 0.00 & 1.00 & 1.00 & 0.231\\
CIFAR-100 & DKW & 20,000 & 4 & 0.000 & 0.00 & 0.00 & 0.00 & 0.104\\
CIFAR-100 & Total loss & 20,000 & 4 & 0.000 & 0.00 & 1.00 & 1.00 & 0.228\\
CIFAR-100 & IUT & 20,000 & 16 & 0.000 & 0.99 & 1.00 & 1.00 & 0.254\\
ImageNetV2 & IUT & 500 & 1 & 0.002 & 0.05 & 0.44 & 1.00 & 0.146\\
ImageNetV2 & IUT & 500 & 4 & 0.000 & 0.02 & 0.34 & 1.00 & 0.135\\
ImageNetV2 & Bonferroni & 500 & 4 & 0.000 & 0.00 & 0.00 & 0.00 & 0.047\\
ImageNetV2 & DKW & 500 & 4 & 0.000 & 0.00 & 0.00 & 0.00 & 0.000\\
ImageNetV2 & Total loss & 500 & 4 & 0.000 & 0.02 & 0.19 & 0.95 & 0.123\\
ImageNetV2 & IUT & 500 & 16 & 0.000 & 0.02 & 0.34 & 0.99 & 0.134\\
ImageNetV2 & IUT & 1,000 & 1 & 0.004 & 0.11 & 0.82 & 1.00 & 0.162\\
ImageNetV2 & IUT & 1,000 & 4 & 0.002 & 0.09 & 0.80 & 1.00 & 0.153\\
ImageNetV2 & Bonferroni & 1,000 & 4 & 0.000 & 0.00 & 0.00 & 0.16 & 0.083\\
ImageNetV2 & DKW & 1,000 & 4 & 0.000 & 0.00 & 0.00 & 0.00 & 0.000\\
ImageNetV2 & Total loss & 1,000 & 4 & 0.000 & 0.00 & 0.28 & 1.00 & 0.135\\
ImageNetV2 & IUT & 1,000 & 16 & 0.002 & 0.08 & 0.80 & 1.00 & 0.153\\
ImageNetV2 & IUT & 2,500 & 1 & 0.010 & 0.27 & 1.00 & 1.00 & 0.174\\
ImageNetV2 & IUT & 2,500 & 4 & 0.000 & 0.31 & 1.00 & 1.00 & 0.169\\
ImageNetV2 & Bonferroni & 2,500 & 4 & 0.000 & 0.00 & 0.00 & 1.00 & 0.120\\
ImageNetV2 & DKW & 2,500 & 4 & 0.000 & 0.00 & 0.00 & 0.00 & 0.000\\
ImageNetV2 & Total loss & 2,500 & 4 & 0.000 & 0.00 & 0.52 & 1.00 & 0.144\\
ImageNetV2 & IUT & 2,500 & 16 & 0.000 & 0.31 & 1.00 & 1.00 & 0.169\\
ImageNetV2 & IUT & 5,000 & 1 & 0.004 & 0.52 & 1.00 & 1.00 & 0.180\\
ImageNetV2 & IUT & 5,000 & 4 & 0.000 & 0.66 & 1.00 & 1.00 & 0.176\\
ImageNetV2 & Bonferroni & 5,000 & 4 & 0.000 & 0.00 & 0.28 & 1.00 & 0.140\\
ImageNetV2 & DKW & 5,000 & 4 & 0.000 & 0.00 & 0.00 & 0.00 & 0.000\\
ImageNetV2 & Total loss & 5,000 & 4 & 0.000 & 0.00 & 0.79 & 1.00 & 0.149\\
ImageNetV2 & IUT & 5,000 & 16 & 0.000 & 0.66 & 1.00 & 1.00 & 0.176\\
ImageNetV2 & IUT & 10,000 & 1 & 0.013 & 0.79 & 1.00 & 1.00 & 0.187\\
ImageNetV2 & IUT & 10,000 & 4 & 0.003 & 0.94 & 1.00 & 1.00 & 0.181\\
ImageNetV2 & Bonferroni & 10,000 & 4 & 0.000 & 0.00 & 1.00 & 1.00 & 0.156\\
ImageNetV2 & DKW & 10,000 & 4 & 0.000 & 0.00 & 0.00 & 0.00 & 0.000\\
ImageNetV2 & Total loss & 10,000 & 4 & 0.000 & 0.00 & 0.97 & 1.00 & 0.152\\
ImageNetV2 & IUT & 10,000 & 16 & 0.003 & 0.94 & 1.00 & 1.00 & 0.181\\
ImageNetV2 & IUT & 20,000 & 1 & 0.015 & 0.99 & 1.00 & 1.00 & 0.193\\
ImageNetV2 & IUT & 20,000 & 4 & 0.000 & 1.00 & 1.00 & 1.00 & 0.184\\
ImageNetV2 & Bonferroni & 20,000 & 4 & 0.000 & 0.00 & 1.00 & 1.00 & 0.167\\
ImageNetV2 & DKW & 20,000 & 4 & 0.000 & 0.00 & 0.00 & 0.00 & 0.045\\
ImageNetV2 & Total loss & 20,000 & 4 & 0.000 & 0.00 & 1.00 & 1.00 & 0.155\\
ImageNetV2 & IUT & 20,000 & 16 & 0.000 & 1.00 & 1.00 & 1.00 & 0.184\\
CMU-MOSEI & IUT & 500 & 1 & 0.012 & 0.12 & 0.72 & 1.00 & 0.104\\
CMU-MOSEI & IUT & 500 & 4 & 0.002 & 0.03 & 0.53 & 1.00 & 0.088\\
CMU-MOSEI & Bonferroni & 500 & 4 & 0.000 & 0.00 & 0.00 & 0.09 & 0.025\\
CMU-MOSEI & DKW & 500 & 4 & 0.000 & 0.00 & 0.00 & 0.00 & 0.000\\
CMU-MOSEI & Total loss & 500 & 4 & 0.000 & 0.00 & 0.03 & 0.79 & 0.052\\
CMU-MOSEI & IUT & 500 & 16 & 0.002 & 0.02 & 0.48 & 1.00 & 0.086\\
CMU-MOSEI & IUT & 1,000 & 1 & 0.006 & 0.23 & 0.97 & 1.00 & 0.116\\
CMU-MOSEI & IUT & 1,000 & 4 & 0.000 & 0.10 & 0.97 & 1.00 & 0.103\\
CMU-MOSEI & Bonferroni & 1,000 & 4 & 0.000 & 0.00 & 0.00 & 0.98 & 0.049\\
CMU-MOSEI & DKW & 1,000 & 4 & 0.000 & 0.00 & 0.00 & 0.00 & 0.000\\
CMU-MOSEI & Total loss & 1,000 & 4 & 0.000 & 0.00 & 0.02 & 0.98 & 0.058\\
CMU-MOSEI & IUT & 1,000 & 16 & 0.000 & 0.09 & 0.96 & 1.00 & 0.102\\
CMU-MOSEI & IUT & 2,500 & 1 & 0.004 & 0.60 & 1.00 & 1.00 & 0.127\\
CMU-MOSEI & IUT & 2,500 & 4 & 0.000 & 0.52 & 1.00 & 1.00 & 0.117\\
CMU-MOSEI & Bonferroni & 2,500 & 4 & 0.000 & 0.00 & 0.08 & 1.00 & 0.079\\
CMU-MOSEI & DKW & 2,500 & 4 & 0.000 & 0.00 & 0.00 & 0.00 & 0.000\\
CMU-MOSEI & Total loss & 2,500 & 4 & 0.000 & 0.00 & 0.00 & 1.00 & 0.067\\
CMU-MOSEI & IUT & 2,500 & 16 & 0.000 & 0.50 & 1.00 & 1.00 & 0.116\\
CMU-MOSEI & IUT & 5,000 & 1 & 0.018 & 0.91 & 1.00 & 1.00 & 0.132\\
CMU-MOSEI & IUT & 5,000 & 4 & 0.010 & 0.91 & 1.00 & 1.00 & 0.124\\
CMU-MOSEI & Bonferroni & 5,000 & 4 & 0.000 & 0.00 & 0.96 & 1.00 & 0.095\\
CMU-MOSEI & DKW & 5,000 & 4 & 0.000 & 0.00 & 0.00 & 0.00 & 0.000\\
CMU-MOSEI & Total loss & 5,000 & 4 & 0.000 & 0.00 & 0.01 & 1.00 & 0.072\\
CMU-MOSEI & IUT & 5,000 & 16 & 0.010 & 0.91 & 1.00 & 1.00 & 0.124\\
\bottomrule
\end{longtable}
}
\paragraph{Fixed-split frontiers on 12 outputs.}\label{app:structured-comparison}
We use one predeclared split, seed 3401, for each of the 12 frozen model outputs. Calibration uses half the scores and no audit labels; audit counts use the disjoint half. We report every combination of $K=1,4,16$ at $\kappa=.9$. For each configuration, IUT, policywise Bonferroni, and DKW use the nominal error budget $.05$. The Rademacher calculation uses 200,000 signs for each of the coverage and accepted-error classes; Hoeffding upper confidence corrections cost $.0005$ each, with the remaining $.049$ allocated over four one-sided data events. Guarantees are pointwise in $K$, not simultaneous over all reported configurations. These methods have different inferential targets internally but each resulting frontier is valid for the same existential contract. The global-loss comparator is deliberately described as score-agnostic, and no data-dependent maximum of unadjusted certificates is used.
\small\begin{longtable}{lrrrrr}
\caption{Fixed-split structured comparison at $\kappa=.9$. All frontiers are percentages, pointwise 95\% under the stated sampling assumptions. Numerical uncertainty is included for the empirical-Rademacher comparator.}\\
\toprule
Output / $K$ & IUT & Bonf. & DKW & Rad. & Global\\
\midrule
\endfirsthead
\toprule
Output / $K$ & IUT & Bonf. & DKW & Rad. & Global\\
\midrule
\endhead
CIFAR-10 / 1 & 6.40 & 5.64 & 0.99 & 0.00 & 1.37\\
CIFAR-10 / 4 & 6.22 & 4.16 & 0.00 & 0.00 & 1.37\\
CIFAR-10 / 16 & 6.22 & 3.28 & 0.00 & 0.00 & 1.37\\
CIFAR-100 / 1 & 24.56 & 23.16 & 18.87 & 8.35 & 22.25\\
CIFAR-100 / 4 & 24.29 & 20.37 & 0.00 & 4.29 & 22.25\\
CIFAR-100 / 16 & 24.29 & 18.82 & 0.00 & 1.93 & 22.25\\
CIFAR-100 (2) / 1 & 24.11 & 21.62 & 17.75 & 8.28 & 21.50\\
CIFAR-100 (2) / 4 & 23.70 & 19.74 & 0.00 & 3.93 & 21.50\\
CIFAR-100 (2) / 16 & 23.70 & 17.82 & 0.00 & 1.44 & 21.50\\
CIFAR-100 (3) / 1 & 23.58 & 22.20 & 17.46 & 8.09 & 21.45\\
CIFAR-100 (3) / 4 & 23.58 & 19.77 & 0.00 & 3.77 & 21.45\\
CIFAR-100 (3) / 16 & 23.58 & 17.77 & 0.00 & 1.63 & 21.45\\
CIFAR-10 (2) / 1 & 6.61 & 5.26 & 1.20 & 0.00 & 1.39\\
CIFAR-10 (2) / 4 & 6.57 & 4.39 & 0.00 & 0.00 & 1.39\\
CIFAR-10 (2) / 16 & 6.57 & 3.41 & 0.00 & 0.00 & 1.39\\
CIFAR-10 (3) / 1 & 6.81 & 5.24 & 0.84 & 0.00 & 1.17\\
CIFAR-10 (3) / 4 & 6.17 & 3.91 & 0.00 & 0.00 & 1.17\\
CIFAR-10 (3) / 16 & 6.17 & 3.36 & 0.00 & 0.00 & 1.17\\
ImageNetV2 / 1 & 17.78 & 16.55 & 12.45 & 2.95 & 14.89\\
ImageNetV2 / 4 & 17.71 & 14.13 & 0.00 & 0.00 & 14.89\\
ImageNetV2 / 16 & 17.71 & 12.58 & 0.00 & 0.00 & 14.89\\
ImageNetV2 (2) / 1 & 24.83 & 22.61 & 18.39 & 8.27 & 22.53\\
ImageNetV2 (2) / 4 & 24.12 & 20.22 & 0.00 & 4.24 & 22.53\\
ImageNetV2 (2) / 16 & 24.12 & 18.83 & 0.00 & 2.04 & 22.53\\
ImageNetV2 (3) / 1 & 27.43 & 25.26 & 21.18 & 10.35 & 25.68\\
ImageNetV2 (3) / 4 & 27.38 & 23.04 & 0.00 & 6.24 & 25.68\\
ImageNetV2 (3) / 16 & 27.24 & 21.20 & 0.00 & 3.69 & 25.68\\
MOSEI / 1 & 12.44 & 9.94 & 3.80 & 0.00 & 6.53\\
MOSEI / 4 & 11.58 & 7.20 & 0.00 & 0.00 & 6.53\\
MOSEI / 16 & 11.58 & 5.45 & 0.00 & 0.00 & 6.53\\
MOSEI (2) / 1 & 10.80 & 8.72 & 2.89 & 0.00 & 5.68\\
MOSEI (2) / 4 & 10.80 & 6.56 & 0.00 & 0.00 & 5.68\\
MOSEI (2) / 16 & 10.80 & 4.86 & 0.00 & 0.00 & 5.68\\
MOSEI (3) / 1 & 13.75 & 11.47 & 5.24 & 0.00 & 8.65\\
MOSEI (3) / 4 & 13.51 & 8.56 & 0.00 & 0.00 & 8.65\\
MOSEI (3) / 16 & 13.30 & 6.55 & 0.00 & 0.00 & 8.65\\
\bottomrule
\end{longtable}
\normalsize

\subsection{Held-out ImageNetV2 IUT audit}
\label{app:heldout-iut}
For the reference-free certificate, we use the frozen ImageNetV2 ViT-B/16 output and split it 50/50 with seed 3401. The calibration half defines 57 ordered cells from the stored \texttt{max\_softmax\_score} channel; the audit half contains 5,000 images and supplies all labels/counts. We search unions of at most $K$ contiguous cell intervals and report the 90\%-coverage contract. Alongside the exact IUT we report policywise Bonferroni, the structured DKW comparator of Appendix~\ref{app:dkw-baseline}, and the VC/Rademacher comparator of Appendix~\ref{app:rademacher-baseline}.

\begin{table}[ht]
\caption{Held-out ImageNetV2 audit at $\kappa=.90$. Lower frontiers are 95\% valid under each method's stated probability budget. ``Rad.'' is the Monte-Carlo-certified empirical-Rademacher uniform-convergence comparator.}
\label{tab:heldout-iut}
\centering
\scriptsize
\begin{tabular}{rrrrrrr}
\toprule
$K$ & policy count & empirical & IUT & Bonf. & DKW & Rad.\\
\midrule
1 & $1.65\times10^3$ & 19.996\% & 17.783\% & 16.552\% & 12.451\% & 2.947\%\\
2 & $4.26\times10^5$ & 19.410\% & 17.783\% & 15.242\% & 4.660\% & 1.085\%\\
4 & $1.96\times10^9$ & 19.348\% & 17.712\% & 14.135\% & 0.000\% & 0.000\%\\
8 & $9.10\times10^{13}$ & 19.348\% & 17.712\% & 13.321\% & 0.000\% & 0.000\%\\
16 & $1.23\times10^{17}$ & 19.348\% & 17.712\% & 12.584\% & 0.000\% & 0.000\%\\
\bottomrule
\end{tabular}
\end{table}
The IUT value stabilizes once additional intervals no longer improve the relevant frontier. DKW accumulates endpoint error linearly in $K$, while the empirical-Rademacher baseline explicitly controls the VC-$2K$ interval-union class without a raw policy-count correction but is still conservative at this sample size. Thus the comparison does not rest on Bonferroni as the only alternative. A one-policy split-sample select-then-test procedure is not listed because it tests only the selected member and is not level-valid for the existential union null. We do not claim any displayed comparator is minimax optimal for the existential problem. On the supplied CPU implementation ($M=57,n=5000$), median DP times are 1.15, 1.60, 2.43, 4.34, 8.13, and 14.80 ms for $K=1,2,4,8,16,32$, respectively; absolute timing is environment-specific.

\begin{figure}[ht]
 \centering
 \includegraphics[width=.66\linewidth]{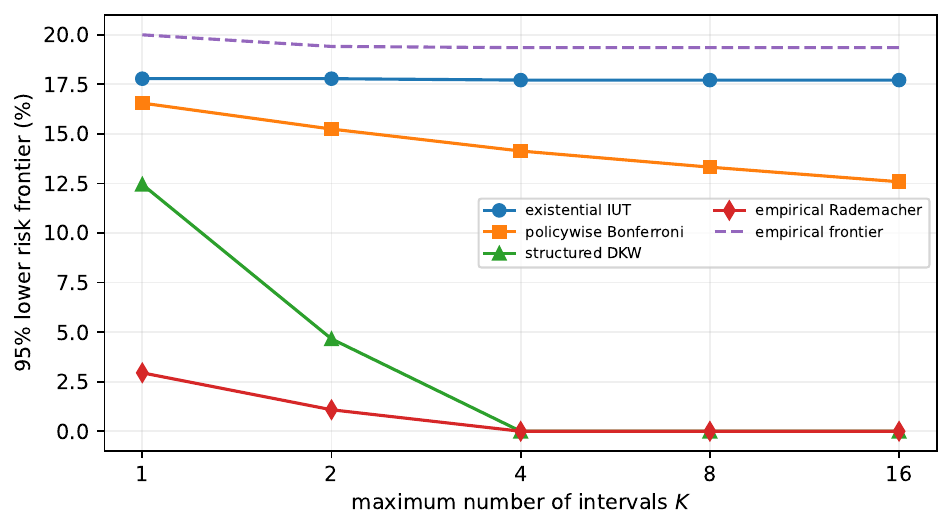}
 \caption{Held-out ImageNetV2 audit at $\kappa=.90$. The exact IUT frontier is shown with policywise Bonferroni, structured DKW, empirical-Rademacher uniform convergence, and the empirical frontier. This illustrates Proposition~\ref{prop:iut}; it is not a minimax-optimality comparison.}
 \label{fig:heldout-iut-app}
\end{figure}

\paragraph{Boundary-null implementation check.}
We additionally simulate 1,000 i.i.d. audits from a finite-cell null containing a known boundary witness: 12 equally likely score cells, $A^*$ equal to the first nine cells, $c(A^*)=.75$, $r(A^*)=.20$, $K=3$, and $n=300$. The IUT+DP implementation rejects the true null in 9/1,000 runs, an empirical rate of 0.9\% with 95\% Wilson interval 0.47--1.70\%. This is intentionally only an implementation regression check; Proposition~\ref{prop:iut} supplies the exact finite-sample level guarantee.

\paragraph{Operating-characteristic check.}
To distinguish conservativeness at the boundary from lack of power, we keep the same $M=12$, $K=3$, $n=300$, $\kappa=.75$, and target $\alpha=.20$ design and vary the best population risk. Over 2,000 simulations per point, rejection rates are:
\begin{center}
\begin{tabular}{rrrrrr}
\toprule
best population risk & .20 & .22 & .25 & .30 & .35\\
\midrule
IUT rejection rate & .0115 & .0555 & .3015 & .8620 & .9955\\
\bottomrule
\end{tabular}
\end{center}
This is a descriptive power curve, not a replacement for Proposition~\ref{prop:iut}'s exact level guarantee. It shows that the 0.9--1.2\% boundary rejection rate is a feature of conservative exact one-sided bounds rather than a procedure that remains inert away from the boundary.

\paragraph{Family-flexibility power cost.}\label{app:flexibility}
The phrase ``no multiplicity correction'' concerns the level proof, not immunity of power to the size of the searched family. To isolate this distinction, we simulate a homogeneous alternative with $M=12$, $n=300$, $\kappa=.75$, target $\alpha=.20$, and population risk $.25$ in every score cell. Hence every policy in every interval-union class has exactly the same population risk; increasing $K$ creates no better population witness. Over 3,000 simulations, IUT rejection falls from 23.1\% at $K=1$ to 7.6\% at $K=2$, 4.7\% at $K=3$, and 4.4\% at $K=4$ and $6$. The median lower frontier simultaneously falls from 18.17\% to 16.03\%. Thus family flexibility can materially reduce power even though the exact type-I argument pays no raw $|\cA|$ factor.

\subsection{Abstention-odds simulation}
\label{app:kappa-law}
\emph{Design.} Scores are uniform over $M=64$ equal-mass cells and $L\sim\Bern(.14)$ independently of the score, so every union of cells has risk $.14=\alpha+\gamma$ with $\alpha=.10$, $\gamma=.04$; the law lies in $\mathcal P_\gamma(\cA_K)$ for every $K$ and $\kappa$, and every rule is exactly as bad as every other. The menus are unions of at most $K\in\{1,2,4,8,16\}$ intervals of cells, and the IUT uses $\zeta_c=\zeta_r=.025$. On a grid of log-spaced sizes from 150 to 15,846 (the first 15 points of a 19-point grid up to 60,000; the run was stopped once every power had reached 1) we draw 250 audits per size, with seeds 6101 and 6102 for the two interleaved halves of the grid. One dynamic program per audit serves all $(K,\kappa)$. Crossings are log-linear interpolations of Monte Carlo power; with 250 audits the power standard error is at most $.032$, which moves a crossing by a few percent.

\emph{Results.} Table~\ref{tab:kappa-law} lists all crossings. The normalization divides by the abstention odds and by $\log(e/\min\{\kappa,1-\kappa\})$, a logarithmic factor suggested by localized VC bounds (Appendix~\ref{app:bracket}); without that logarithm the normalized costs still vary much less than the raw ones (at $K=4$: 3.9 at $\kappa=.5$ and 9.8 at $\kappa=.95$, against a raw variation of 7.7-fold in the opposite direction). At $\kappa=.3$ the normalized costs are 0.56--0.64 times those at $\kappa=.5$, within the constant factor that separates $(1-\kappa)/\kappa$ from $\min\{\kappa,1-\kappa\}/\kappa^2$. At $\kappa\ge.9$ and $K\ge8$ the crossings stop growing: the abstained region contains only $M(1-\kappa)\le6.4$ cells, so no new rule competes. The constant-risk law is not the least favorable law of Theorem~\ref{cor:intervals}; the experiment checks the $\kappa$-dependence of the IUT's cost, which is governed by the upper bound. Script: \texttt{kappa\_law.py}; figure and table: \texttt{make\_kappa\_figure.py}, \texttt{make\_kappa\_table.py}.
\begin{table}[ht]
\caption{Abstention-odds simulation: 80\%-power crossings $n_{80}$ of the exact IUT ($M=64$ equal cells, constant risk $.14$, $\alpha=.10$, $\gamma=.04$, 250 audits per grid point, log-linear interpolation), and normalized menu costs $\gamma^2\{n_{80}(K)-n_{80}(1)\}\big/\{\frac{1-\kappa}{\kappa}\log\frac{e}{\min(\kappa,1-\kappa)}\}$.}
\label{tab:kappa-law}
\centering\scriptsize
\begin{tabular}{@{}lrrrrrr@{}}
\toprule
$K$ \textbackslash\ $\kappa$ & .3 & .5 & .7 & .8 & .9 & .95\\
\midrule
1 & 3{,}832 & 1{,}958 & 1{,}070 & 881 & 755 & 656\\
2 & 5{,}764 & 3{,}092 & 1{,}847 & 1{,}402 & 973 & 839\\
4 & 8{,}252 & 4{,}414 & 2{,}737 & 1{,}887 & 1{,}316 & 977\\
8 & 11{,}961 & 6{,}151 & 3{,}458 & 2{,}584 & 1{,}568 & 1{,}024\\
16 & 14{,}050 & 7{,}389 & 3{,}921 & 2{,}732 & 1{,}580 & 1{,}024\\
\midrule
norm., $K=2$ & 0.60 & 1.07 & 1.32 & 1.28 & 0.95 & 1.39\\
norm., $K=4$ & 1.38 & 2.32 & 2.82 & 2.47 & 2.45 & 2.45\\
norm., $K=8$ & 2.53 & 3.96 & 4.04 & 4.18 & 3.54 & 2.80\\
norm., $K=16$ & 3.18 & 5.13 & 4.83 & 4.54 & 3.60 & 2.80\\
\bottomrule
\end{tabular}
\end{table}

\subsection{A valid non-EFC audit: cross-fit universal inference}
\label{app:nonefc}
Identify each law with its pushforward to the $2M$ cell-by-loss categories; as in Appendix~\ref{app:constant}, this loses nothing for the null because every rule is a union of cells. For $A\in\cA$ let $\cH_A$ be the laws under which $A$ meets the contract, so $\cH=\bigcup_A\cH_A$.
\begin{lemma}[Profile likelihood of the existential null]
\label{lem:profile}
Let $x$ be category counts with total $n_0$, and let $\hat c$ be the empirical coverage of $A$. When $\hat c>0$, let $\hat r$ be its empirical risk; when $\hat c=0$, define the product $\hat c\,\1\{\hat r>\alpha\}\,\mathrm{kl}(\hat r,\alpha)$ below to be zero. Then
\[
 \sup_{Q\in\cH_A}\sum_ix_i\log Q_i=\sum_ix_i\log\frac{x_i}{n_0}-n_0\Big[\1\{\hat c<\kappa\}\,\mathrm{kl}(\hat c,\kappa)+\hat c\,\1\{\hat r>\alpha\}\,\mathrm{kl}(\hat r,\alpha)\Big].
\]
\end{lemma}
\begin{proof}
The constraints of $\cH_A$ involve $Q$ only through the four masses $Q(A,0),Q(A,1),Q(A^c,0),Q(A^c,1)$. For fixed masses, the likelihood is maximized by splitting each mass in proportion to the counts, which gives $\sum_ix_i\log(x_i/n_0)-n_0\,\KL(\hat p\Vert Q_4)$, where $\hat p$ and $Q_4$ are the empirical and model four-cell laws. Parametrize $Q_4$ by $c=Q(A)$, $r=Q(L=1\mid A)$ and $s=Q(L=1\mid A^c)$. The chain rule gives $\KL(\hat p\Vert Q_4)=\mathrm{kl}(\hat c,c)+\hat c\,\mathrm{kl}(\hat r,r)+(1-\hat c)\mathrm{kl}(\hat s,s)$, with zero-weight conditional terms defined as zero (so the empirical conditional proportion on an empty block may be chosen arbitrarily). The constraints are $c\ge\kappa$ and $r\le\alpha$, and the three terms are minimized separately by $c=\max\{\hat c,\kappa\}$, by $r=\min\{\hat r,\alpha\}$ when $\hat c>0$, and by $s=\hat s$ when $\hat c<1$; the corresponding zero-weight choices are immaterial.
\end{proof}
For $t>0$, the bracket is increasing in the error count for a fixed accepted count, so the minimal error counts $e_K(t)$ of the interval dynamic program (Appendix~\ref{app:iut}) suffice. If a nonempty frozen score cell receives zero observations in the evaluated split, a nonempty rule with $t=0$ is also available and contributes the coverage-only penalty $\mathrm{kl}(0,\kappa)$; we include this candidate explicitly. The structurally empty mask is not a witness for the existential null. On small instances the formula agrees with numerical constrained optimization to $10^{-9}$, and the dynamic-program minimum, including the zero-count edge case, agrees with brute-force enumeration.

\emph{The audit.} Split the audit sample into halves $D_0,D_1$. Let $\hat P_1$ be the add-$1/2$ smoothed empirical law of $D_1$ over the $2M$ categories, $\Lambda_0=\hat P_1^{n_0}(D_0)/\sup_{Q\in\cH}Q^{n_0}(D_0)$ the split likelihood ratio, $\Lambda_1$ its mirror image, and reject when $(\Lambda_0+\Lambda_1)/2\ge1/\delta$. For every $Q\in\cH$, $\E_Q[\Lambda_0\mid D_1]\le\E_Q[\hat P_1^{n_0}(D_0)/Q^{n_0}(D_0)\mid D_1]\le1$: explicitly, the latter expectation sums $\hat P_1^{n_0}$ only over sample sequences with positive $Q^{n_0}$ mass, and equals one when $\hat P_1\ll Q$. The same holds for $\Lambda_1$, so the average is an e-value and Markov's inequality gives level $\delta$ over the unrestricted null \citep{wasserman2020universal}. The audit is not EFC: it may reject although a rule is empirically feasible on the full sample. It is a natural reference-free competitor, and it is not the testing-scale audit of Theorem~\ref{thm:constant}, which requires constant risk.

\emph{Results.} Table~\ref{tab:nonefc} compares it with the IUT on the same draws. No boundary rejection was observed for the universal-inference audit. The IUT's largest boundary rejection rate in these runs is 4.0\% (ImageNetV2, $K=1$, $n=50{,}000$, 200 audits), within Monte Carlo error of its exact level; up to $n=20{,}000$ it matches the at most 1.8\% of Table~\ref{tab:semisyn-full} within that error. The $\chi^2$ independence test rejected in all 200 audits at $n=500$ on every population. The numerator must fit the full cell-by-loss law, which has up to $2M-1=113$ free parameters here (105 for CIFAR-10 after ties are merged); this is a substantial finite-sample burden that the IUT does not incur. Script: \texttt{nonefc\_semisyn.py}; table: \texttt{make\_nonefc\_table.py}.
\begin{table}[ht]
\caption{A valid non-EFC audit on the frozen populations ($\kappa=.9$, $K=4$, 200 audits per point, $n\in\{500,1000,2500,5000,10^4,2\times10^4,5\times10^4\}$). Entries give the smallest grid size at which the rejection rate of the false claim $\alpha=r^*-\Delta$ reaches 80\%. ``Level'' is the largest rejection rate of a true boundary claim over all $n$ and $K\in\{1,4,16\}$. The last column is the median $p$-value of a $\chi^2$ test of cell--loss independence at $n=500$.}
\label{tab:nonefc}
\centering\scriptsize
\setlength{\tabcolsep}{4pt}
\begin{tabular}{@{}lr|rr|rr|rr|r@{}}
\toprule
& & \multicolumn{2}{c|}{$\Delta=.05$} & \multicolumn{2}{c|}{$\Delta=.02$} & \multicolumn{2}{c|}{level} & independence\\
Population & $r^*$ & IUT & split LR & IUT & split LR & IUT & split LR & median $p$\\
\midrule
CIFAR-10 & 0.074 & 500 & 10{,}000 & 5{,}000 & $>50{,}000$ & 2.0\% & 0.0\% & 2e-17\\
CIFAR-100 & 0.262 & 2{,}500 & 50{,}000 & 10{,}000 & $>50{,}000$ & 3.0\% & 0.0\% & 3e-19\\
ImageNetV2 & 0.193 & 2{,}500 & 50{,}000 & 10{,}000 & $>50{,}000$ & 4.0\% & 0.0\% & 3e-17\\
CMU-MOSEI & 0.136 & 1{,}000 & 50{,}000 & 5{,}000 & $>50{,}000$ & 3.5\% & 0.0\% & 5e-10\\
\bottomrule
\end{tabular}
\end{table}

\section{Artifact provenance and automated checks}

The release includes the frozen dependency artifact plus an analysis directory containing the operational benchmark, sign-family simulation, full-null window reconstruction, exact parity certificates, validation tests, and generated manuscript tables. All computational configurations have recorded deterministic seeds. Figures and numerical controls computed from frozen outputs are identified as such; their source scripts and input arrays are included. The parity certificates are verified with rational arithmetic; the sign dynamic program and binomial rejection lookup are checked against independent exhaustive calculations; singleton radii are checked against constrained four-cell numerical optimization. The release README supplies commands, software versions, and claim-to-file mappings.

\end{document}